\documentclass[letterpaper]{article} % DO NOT CHANGE THIS
\usepackage{aaai2026}  % DO NOT CHANGE THIS
\usepackage{times}  % DO NOT CHANGE THIS
\usepackage{helvet}  % DO NOT CHANGE THIS
\usepackage{courier}  % DO NOT CHANGE THIS
\usepackage[hyphens]{url}  % DO NOT CHANGE THIS
\usepackage{graphicx} % DO NOT CHANGE THIS
\usepackage{natbib}  % DO NOT CHANGE THIS AND DO NOT ADD ANY OPTIONS TO IT
\usepackage{caption} % DO NOT CHANGE THIS AND DO NOT ADD ANY OPTIONS TO IT
\usepackage{amsmath}
\usepackage{amssymb}
\usepackage[capitalize]{cleveref}
\crefname{section}{Sec.}{Secs.}
\Crefname{section}{Section}{Sections}
\Crefname{table}{Table}{Tables}
\crefname{table}{Tab.}{Tabs.}
\crefname{algocf}{Alg.}{Algs.}
\Crefname{algocf}{Algorithm}{Algorithms}
\usepackage{dsfont}
\usepackage{booktabs}
\usepackage{multirow}
\usepackage{booktabs}

\usepackage{newfloat}
\usepackage{listings}
\DeclareCaptionStyle{ruled}{labelfont=normalfont,labelsep=colon,strut=off} % DO NOT CHANGE THIS
\usepackage[table,dvipsnames]{xcolor}
\usepackage{colortbl}
\usepackage{pifont}
\usepackage{array}
\usepackage[ruled,vlined,linesnumbered]{algorithm2e}
\usepackage{arydshln}

\title{Open-World Deepfake Attribution via Confidence-Aware Asymmetric Learning}
\author{
    Haiyang Zheng\textsuperscript{\rm 1}, Nan Pu\textsuperscript{\rm 1,2*}, Wenjing Li\textsuperscript{\rm 2}\thanks{Corresponding author.}, Teng Long\textsuperscript{\rm 1}, Nicu Sebe\textsuperscript{\rm 1}, Zhun Zhong\textsuperscript{\rm 2} \\
}
\affiliations{
    \textsuperscript{\rm 1}University of Trento, \\
    \textsuperscript{\rm 2}Hefei University of Technology\\
    
}

\usepackage{bibentry}
\begin{document}

\maketitle

\begin{abstract}
  The proliferation of synthetic facial imagery has intensified the need for robust Open-World DeepFake Attribution (OW-DFA), which aims to attribute both known and unknown forgeries using labeled data for known types and unlabeled data containing a mixture of known and novel types. However, existing OW-DFA methods face two critical limitations: 1) A \textbf{confidence skew} that leads to unreliable pseudo-labels for novel forgeries, resulting in biased training. 2) An \textbf{unrealistic assumption} that the number of unknown forgery types is known \textit{a priori}.
  To address these challenges, we propose a Confidence-Aware Asymmetric Learning (\textbf{CAL}) framework, which adaptively balances model confidence across known and novel forgery types. CAL mainly consists of two components: Confidence-Aware Consistency Regularization (\textbf{CCR}) and Asymmetric Confidence Reinforcement (\textbf{ACR}). \textit{CCR} mitigates pseudo-label bias by dynamically scaling sample losses based on normalized confidence, gradually shifting the training focus from high- to low-confidence samples. \textit{ACR} complements this by separately calibrating confidence for known and novel classes through selective learning on high-confidence samples, guided by their confidence gap. Together, \textit{CCR} and \textit{ACR} form a mutually reinforcing loop that significantly improves the model's OW-DFA performance.
  Moreover, we introduce a Dynamic Prototype Pruning (\textbf{DPP}) strategy that automatically estimates the number of novel forgery types in a coarse-to-fine manner, removing the need for unrealistic prior assumptions and enhancing the scalability of our methods to real-world OW-DFA scenarios.
  Extensive experiments on the standard OW-DFA benchmark and a newly extended benchmark incorporating advanced manipulations demonstrate that \textit{CAL} consistently outperforms previous methods, achieving new state-of-the-art performance on both known and novel forgery attribution.
\end{abstract}

\begin{links}
    \link{Code}{https://haiyangzheng.github.io/OWDFA-CAL}
    % \link{Datasets}{https://aaai.org/example/datasets}
    % \link{Extended version}{https://aaai.org/example/extended-version}
\end{links}

\section{Introduction}

The rapid advancement of image generation techniques, particularly diffusion models~\cite{rombach2022high,midjourney2024} and autoregressive modeling~\cite{tian2024visual}, has led to a surge in synthetic facial imagery on social media platforms. While such content often serves entertainment purposes, it raises serious concerns regarding identity misuse and the spread of misinformation. To mitigate these risks, DeepFake Attribution (DFA)~\cite{yang2022deepfake,yu2021artificial,guarnera2022exploitation} has been proposed not only to determine whether the content is fake, but also to identify the specific model architectures responsible for forgery generation. However, existing methods mainly focus on GAN-generated images, neglecting more advanced and realistic manipulations such as identity swaps~\cite{mobileswap, uniface}. Moreover, they often assume a closed-world setting, in which training and testing categories are shared—an assumption that rarely holds in open-world scenarios.
\begin{figure}[!t]
  \centering
  \includegraphics[width=\linewidth]{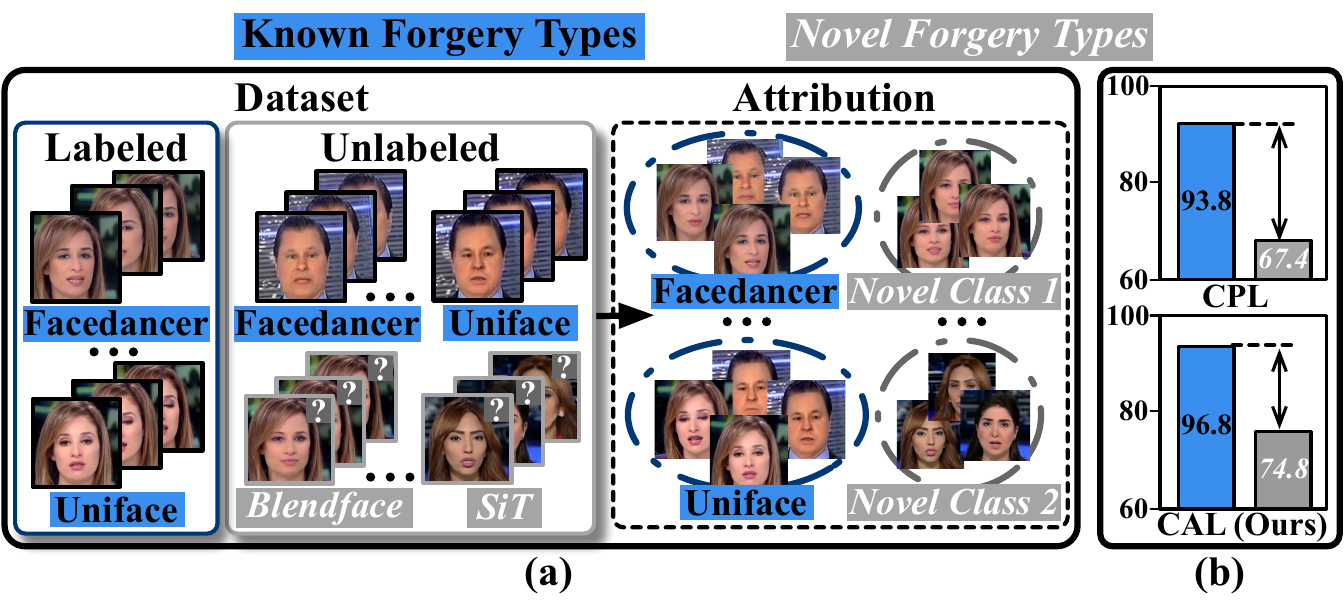}

    \caption{(a) Schema of the Open-World Deepfake Attribution. (b) Our CAL reduces the performance gap compared to the state-of-the-art method CPL~\cite{cpl}, showing average results across all evaluation protocols.\label{fig:highlight}}

\end{figure}

\par

To address these limitations, Sun \textit{et al.}~\cite{cpl} introduce a new Open-World DFA (OW-DFA) task, as shown in \cref{fig:highlight}(a), which requires models to identify both known and novel forgery types in an unlabeled set of manipulated face images by transferring knowledge from a labeled set containing only known forgeries.  
Existing OW-DFA methods~\cite{cpl,sun2025rethinking,cdal} typically adopt pseudo-labeling strategies to exploit unlabeled data and have achieved promising attribution performance. However, these methods exhibit a significant performance gap between known and novel forgery types. Through an in-depth analysis (detailed in~\cref{sec:skew}) of this observation, we argue that this performance gap results from a \textbf{\textit{confidence skew}}: the model assigns lower confidence scores to predictions on novel forgery types, resulting in unreliable pseudo-labels. These inaccurate labels misguide the training objective, reinforcing the bias through a negative feedback loop and further amplifying the skew. Additionally, existing approaches assume that the number of forgery types is known \textit{a priori}. However, such an assumption is unrealistic in real-world open-world scenarios, where generative forgery models evolve continuously. The number of forgery types in newly collected unlabeled data is inherently unknown and potentially unbounded, rendering these methods ill-suited for deployment in dynamic real-world environments.

\par
To address these drawbacks, we propose a novel \textbf{Confidence-Aware Asymmetric Learning (CAL)} framework that adaptively mitigates the model’s confidence imbalance between known and novel forgery types throughout training, consistently improving attribution accuracy on both seen and unseen forgery types. Our CAL framework consists of two key components: \textit{Confidence-Aware Consistency Regularization (CCR)} and \textit{Asymmetric Confidence Reinforcement (ACR)}. CCR effectively rectifies biased pseudo-label learning on unlabeled data by adaptively adjusting the regularization strength in a threshold-free manner. Specifically, during the early training stages, CCR mitigates the effect of noisy supervision by down-weighting low-confidence samples. This is achieved by scaling their loss contributions with normalized confidence scores. As training progresses, the model gradually shifts focus: it reduces the weight assigned to high-confidence samples and emphasizes learning from low-confidence ones, which predominantly correspond to novel forgery types. This adaptive strategy enables more stable and discriminative representation learning throughout the training process. Complementarily, ACR explicitly encourages the model to generate high-confidence predictions separately for known and novel categories via an asymmetric learning strategy. Specifically, we introduce a coefficient based on the model’s confidence gap between known and novel categories to personalize the selection of high-confidence samples for known and novel forgery types. By enforcing learning on these high-confidence samples, ACR facilitates more effective consistency regularization in CCR, creating a mutually beneficial feedback loop. As shown in~\cref{fig:highlight}(b) and \cref{fig:conf_hist}, our CAL largely improves the model's prediction confidence and accuracy on all classes. In addition, we introduce a \textit{Dynamic Prototype Pruning (DPP)} strategy to estimate the number of novel forgery types with negligible computational overhead. We design a coarse-to-fine pruning mechanism to dynamically merge low-usage and redundant prototypes. This enables CAL to scale effectively to real-world OW-DFA deployments, where the number of attack types is unknown.

\par
Overall, our contributions are summarized as follows:

\begin{itemize}
    \item We identify the importance of the confidence skew issue, which significantly degrades the overall performance of existing OW-DFA methods.
    \item We propose a new CAL framework that effectively reduces the confidence skew issue and promotes OW-DFA models toward unbiased learning.
    \item We design a novel DPP strategy that can automatically estimate the number of novel forgery types during training with negligible overhead.
    \item We extend the existing OW-DFA benchmark by incorporating advanced diffusion-based forgeries to build a challenging yet practical OW-DFA-40 benchmark. Extensive experiments show that our CAL achieves state-of-the-art performance on both benchmarks.
\end{itemize}

\section{Related Work}
\label{sec_related}

\noindent\textbf{Open-World DeepFake Attribution (OW-DFA)}. Motivated by growing concerns over privacy protection, DeepFake Attribution (DFA)~\cite{yang2022deepfake,yu2021artificial,guarnera2022exploitation} aims to detect manipulated content while simultaneously identifying the specific generative model architecture responsible for its creation. However, most existing GAN attribution methods~\cite{yang2022deepfake,yu2021artificial,guarnera2022exploitation} rely on model-specific fingerprints and operate under a \textit{closed-world assumption}, where the training and testing distributions are aligned. This assumption often fails in real-world applications, where novel manipulation techniques continue to emerge. To overcome this limitation, the OW-DFA task was introduced in CPL~\cite{cpl} as a more realistic extension of DFA. The goal is to attribute each manipulated face in the unlabeled set, regardless of whether it originates from a known or novel generative model. Building upon CPL, CDAL~\cite{cdal} leverages causal inference and counterfactual contrast to eliminate confounding biases in attention, thus improving the model's ability to identify discriminative generation patterns for robust attribution. \textit{In this work, we follow the OW-DFA setting and further extend the existing benchmark, inspired by \cite{yan2023deepfakebench}, by incorporating diffusion-based forgeries in response to the rapid evolution of generative techniques. Moreover, we propose a more challenging and practical evaluation protocol to comprehensively assess OW-DFA methods under realistic conditions.}

\noindent\textbf{Category Discovery.}
Novel Category Discovery (NCD) was originally proposed to cluster unlabeled data containing only unknown categories by transferring knowledge from labeled categories~\cite{ncd, QING202124,openmix2020,ncl,uno,dualrs,9747827,PSSCNCD,li2023closerlooknovelclass,ResTune,li2023supervised, liu2024novel, cai2023broaden,Li2023ncdiic,sun2023nscl,peiyan2023class,wei2023ncdSkin,10328468,hasan2023novelcategoriesdiscoveryconstraints}. To better reflect real-world scenarios, Generalized Category Discovery (GCD) extends NCD by allowing unlabeled data to contain both known and unknown categories, which has since emerged as a widely adopted and practical setting for category discovery~\cite{he2025category, zhu2024open, gcd, pu2023dynamic, simgcd, legogcd, protogcd, palgcd, zheng2025generalized, rastegar2024selex, rastegar2023learn, peng2025mos, wang2025learning, fan2025open, aplgcd,congcd,hypcd}. Beyond the standard setting, recent studies further extend GCD to more practical and realistic scenarios, including federated GCD~\cite{fgcd,zhang2023unbiasedtrainingfederatedopenworld,wang2023federatedcontinualnovelclass, wang2025federated}, on-the-fly discovery~\cite{ocd,phe,liu2025generate, liudaa}, multi-modal category discovery~\cite{clipgcd,textgcd,mgcd,get, fan2025learning, jing2025video}, domain-shifted environments~\cite{Yu_Ikami_Irie_Aizawa_2022,zhuang2022opensetdomainadaptation,zang2023boostingnovelcategorydiscovery,rongali2024cdadnetbridgingdomaingaps,wang2024exclusivestyleremovalcross,wang2024hilo}, and other open-world recognition settings. 

At the algorithmic level, existing GCD methods explore diverse strategies to leverage labeled and unlabeled data, 
including contrastive representation learning on pre-trained features~\cite{gcd,dino}, 
parametric classifiers with self-distillation~\cite{simgcd}, 
view-consistency regularization~\cite{legogcd}, 
prototype-based modeling for category discovery and outlier detection~\cite{protogcd}, 
and hybrid parametric--non-parametric association learning frameworks~\cite{palgcd}. Closely related to GCD, \textbf{Open-World Semi-Supervised Learning (OWSSL)} shares an essentially identical problem setting but was independently introduced in the context of open-set recognition~\cite{cao2021open}. OWSSL methods follow similar principles, such as prototype-based alignment~\cite{sun2022opencon}, pairwise similarity learning~\cite{rizve2022openldn}, hierarchical semantic decomposition~\cite{wang2023discover}, contrastive clustering~\cite{lps}, and conditional self-labeling~\cite{owmatch}. \textit{Despite the shared assumptions, GCD and OWSSL tasks focus on understanding the global semantics of natural images, while OW-DFA requires attention to fine-grained facial details and forgery traces introduced by different deepfake methods. Moreover, existing methods and category estimation strategies are designed based on strong pre-trained models, making them less effective for the OW-DFA task.}

\section{Confidence Skew in OW-DFA}
\label{sec:skew}
\noindent \textbf{Problem Setup.}
We consider open-world deepfake attribution with a labeled set generated by \emph{known} methods,
$\mathcal{D}_{L}=\{(\mathbf{x}^{l}_{i},y^{l}_{i})\in\mathcal{X}\times\mathcal{Y}_{L}\}_{i=1}^{N}$,
and an unlabeled set generated by both known and \emph{novel} methods,
$\mathcal{D}_{U}=\{\mathbf{x}^{u}_{i}\in\mathcal{X}\}_{i=1}^{M}$.
Here $N$ and $M$ are the numbers of samples in $\mathcal{D}_{L}$ and $\mathcal{D}_{U}$, respectively.
Let $\mathcal{Y}_{U}$ be the label space covered by $\mathcal{D}_{U}$, with $\mathcal{Y}_{L}\subset\mathcal{Y}_{U}$.
We define $K_{L} = |\mathcal{Y}_{L}|$ as the number of known classes and $K_{U} = |\mathcal{Y}_{U}|$ as the total number of classes. If $K_{U}$ is known in advance, it can be directly utilized during training; otherwise, it must be estimated in training.

\begin{figure}[h]

  \centering
  \includegraphics[width=\linewidth]{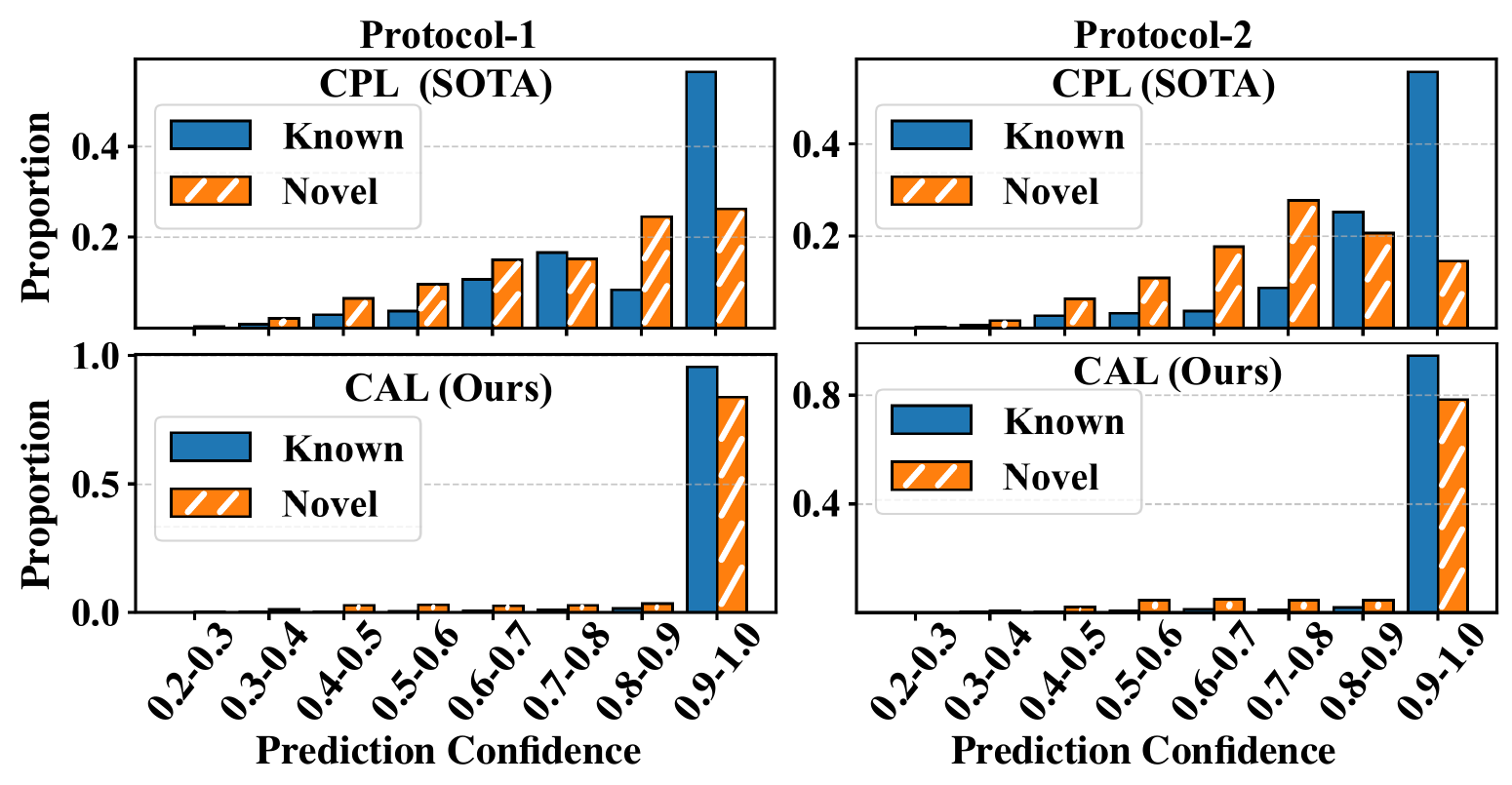}
\caption{Distribution of sample confidence for known vs. novel classes. \label{fig:conf_hist}}
\end{figure}

\begin{figure}[h]
  \centering
  \includegraphics[width=\linewidth]{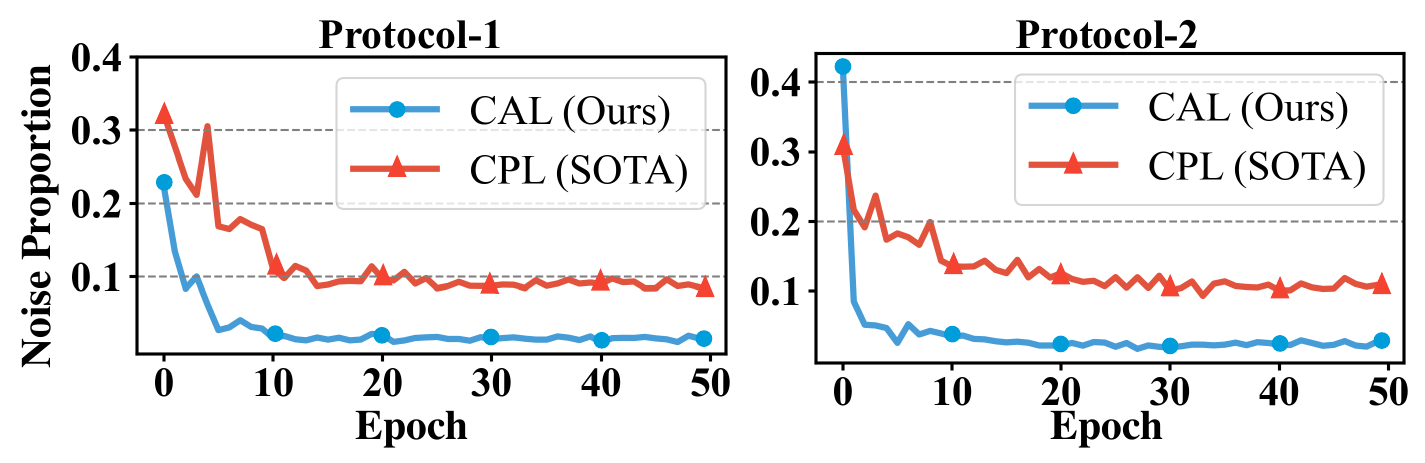}
\caption{Proportion of pseudo-label noise in the low-confidence range $[0, 0.5)$ among all novel class samples.
\label{fig:conf_line}}
\end{figure}

\noindent \textbf{Definition and Observations.}
We define \emph{confidence skew} as a consistent confidence gap between known and novel classes, leading to unreliable pseudo-labels for novel samples and reinforcing the skew.
We make two key observations:
\textbf{(O1)} Novel class samples under CPL show lower average confidence and a flatter distribution than known class samples (\cref{fig:conf_hist}).
\textbf{(O2)} CPL yields a higher ratio of pseudo-label noise in the low-confidence range $[0, 0.5)$ for novel-class samples compared to ours (\cref{fig:conf_line}).

\noindent\textbf{Mechanism.}
\textit{(1) Supervision imbalance $\Rightarrow$ initial confidence skew.}  
Since the labeled set $\mathcal{D}_{L}$ only covers the known classes $\mathcal{Y}_{L}$, early optimization predominantly updates representations toward known classes, thereby biasing the feature space. \textit{(2) Gumbel-Softmax on low-confidence samples $\Rightarrow$ noisy pseudo-labels.}  
CPL generates pseudo-labels using Gumbel-Softmax sampling~\cite{gumbel}. When the predicted logits are flat, the sampling becomes highly sensitive to Gumbel noise, leading to unstable and noisy pseudo-labels.  
For novel forgery samples, which typically exhibit low confidence and flat logit distributions, this results in weak and unreliable supervision.
\textit{(3) Error-reinforcement loop $\Rightarrow$ amplified skew.}  
Low confidence in novel samples leads to unreliable pseudo-labels, which in turn cause incorrect updates that hinder the learning of novel classes, thereby reinforcing and amplifying the skew. A formal analysis of these mechanisms is provided in Appendix~A.1.

\section{Method}
\label{sec_method}

\begin{figure*}[h]
  \centering
  \includegraphics[width=0.98\linewidth]{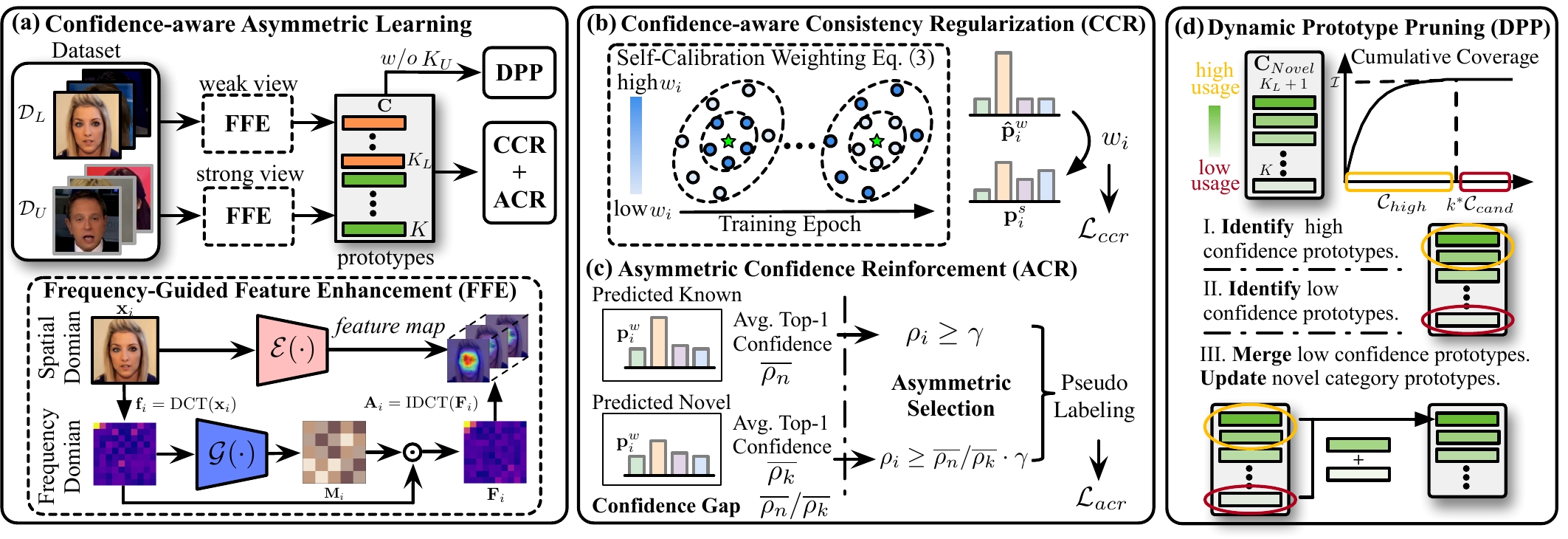}
  \caption{Framework of the proposed Confidence-Aware Asymmetric Learning (CAL).\label{fig:framework}}
\end{figure*}

\subsection{Framework Overview}
As illustrated in~\cref{fig:framework}, we propose a novel Confidence-Aware Asymmetric Learning (CAL) framework to accurately attribute deepfake generation methods. \textit{First}, we introduce a Frequency-guided Feature Enhancement (FFE) module that emphasizes informative regions of manipulated content in the frequency domain, thereby enhancing the discriminative power of learned features. \textit{Second}, we propose a Confidence-Aware Consistency Regularization (CCR) mechanism, which adaptively adjusts the regularization strength for unlabeled samples in a threshold-free manner based on their confidence scores (\cref{fig:framework}(b)). This design mitigates the impact of noisy pseudo-labels and stabilizes the learning process. \textit{Third}, we incorporate an Asymmetric Confidence Reinforcement (ACR) strategy that explicitly guides the model to learn high-confidence prototypical classifiers for both known and novel categories through asymmetric optimization (\cref{fig:framework}(c)). \textit{Finally}, we design a Dynamic Prototype Pruning (DPP) strategy to estimate the number of unseen categories by analyzing the confidence associated with each learned prototype (\cref{fig:framework}(d)).

\noindent \textbf{Frequency-Guided Feature Enhancement.}
Given an input face image $\mathbf{x}_i$, we compute its frequency-domain representation using the Discrete Cosine Transform (DCT), defined as
$\mathbf{f}_i = \mathrm{DCT}(\mathbf{x}_i)$. We then use a randomly initialized convolutional network $\mathcal{G}(\cdot)$ to generate a frequency-domain attention mask:
$\mathbf{M}_i = \mathcal{G}(\mathbf{f}_i)$. We apply this mask to the frequency features via element-wise multiplication to obtain the weighted frequency representation, denoted as
$\mathbf{F}_i = \mathbf{M}_i \odot \mathbf{f}_i$, which is expected to highlight discriminative frequency patterns indicative of different deepfake methods. We next transform the weighted frequency representation back to the spatial domain using the inverse Discrete Cosine Transform (IDCT), denoted as
$\mathbf{A}_i = \mathrm{IDCT}(\mathbf{F}_i)$. The resulting map $\mathbf{A}_i$ captures the spatial localization of frequency-domain responses and serves as a \textit{frequency-aware attention map}.
 To integrate these responses into the attribution process, we define the frequency-aware image representation as the multiplication of $\mathbf{A}_i$ and the spatial feature:
\begin{equation}
    \mathbf{h}_i = Pooling(\mathbf{A}_i \odot \mathcal{E}(\mathbf{x}_i); 1 \times 1),
\end{equation}
where $\mathcal{E}(\cdot)$ denotes the image encoder and $Pooling(\cdot;1\times1)$ denotes global $1\times1$ pooling. The resulting feature $\mathbf{h}_i$ fuses both spatial and frequency-domain cues, enabling more robust and accurate deepfake attribution.

\par\noindent
\textbf{Discussion.}
Prior studies~\cite{zhang2019detecting,ricker2022towards,li2024masksim} in deepfake detection have demonstrated that different types of deepfake generation methods often leave distinct artifacts in the frequency domain of synthesized images. To capture the unique frequency fingerprints of different deepfake methods, we introduce the FFE module to provide complementary frequency-domain insights to enhance spatial-domain features. Although frequency-domain features have been explored in related deepfake detection tasks~\cite{tan2024frequency,li2024masksim,zhou2024freqblender,cao2024towards}, how to inject frequency-domain knowledge into OW-DFA remains under-explored.

\subsection{Confidence-Aware Asymmetric Learning}
\noindent\textbf{Motivation.} To comprehensively address the issue of \textit{confidence skew} in OW-DFA methods, we design two components targeting complementary aspects: \textit{1) Progressive emphasis on low-confidence samples ($\to$CCR).}
In the early training stages, low-confidence predictions—primarily from novel forgeries—are highly prone to label noise. Thus, we initially \emph{down-weight} their gradient contributions. As the model evolves, these weights are progressively increased, enabling the network to effectively learn from previously disregarded samples. \textit{2) Confidence-skew-aware pseudo-label selection ($\to$ACR).}
Although reliable pseudo-labels typically originate from high-confidence predictions, the confidence distributions of known and novel classes are inherently misaligned. To account for this, we select pseudo-labels \emph{independently} within each partition using asymmetric, dynamically updated thresholds that reflect the evolving confidence gap.
This strategy ensures clean supervision across both partitions while mitigating further bias.

\par
\noindent \textbf{Confidence-Aware Consistency Regularization (CCR).}
%\noindent \textbf{Self-Calibration Weighting.} 
Given the lack of ground-truth labels for the unlabeled dataset, we employ a consistency regularization-based training strategy~\cite{sohn2020fixmatch} to encourage similar predictions for perturbed versions of the same image. Specifically, we build a prototypical classifier with $K$ learnable prototypes $\mathbf{C} = \{\mathbf{c}_1, \ldots, \mathbf{c}_K\}$, where each prototype represents a ``fingerprint'' characteristic of a deepfake generation method. Here, $K$ can be set to the ground-truth number of forgery types or estimated by our method in~\cref{sec_dpp}. For the feature representation $\mathbf{h}_i$ of the $i$-th face image, its similarity to the $k$-th prototype is defined as $\mathbf{s}_{i \rightarrow k} = \mathbf{h}_i \cdot \mathbf{c}_k$. The similarities between image $\mathbf{x}_i$ and all $K$ prototypes form the similarity vector $\mathbf{s}_i = [\mathbf{s}_{i \rightarrow 1}, \ldots, \mathbf{s}_{i \rightarrow K}]$. We define the predicted probability as $\mathbf{p}_i = \sigma(\mathbf{s}_i)$, with $\sigma$ denoting the softmax function. For each input image $\mathbf{x}_i$, we generate two augmented views: a weakly augmented version $\mathbf x_i^w$ and a strongly augmented version $\mathbf x_i^s$, and extract the corresponding feature embeddings $\mathbf{h}_i^w$ and $\mathbf{h}_i^s$. Following FixMatch~\cite{sohn2020fixmatch}, we define the consistency regularization loss as:

{\small
\begin{equation}
\mathcal{L}_{ws} = \frac{1}{|\mathcal{B}|}\sum_i \mathds{1}(\max(\mathbf{\hat{p}}_i^w) > \delta) \ell \bigl(\mathbf{\hat{p}}_i^w, \mathbf{p}_i^s\bigr),
\label{loss:ws}
\end{equation}
}
\normalsize

\noindent where $\mathcal{B}$ denotes a mini-batch, $\delta$ is a confidence threshold used to filter high-confidence samples, $\ell(\cdot)$ denotes the cross-entropy loss, and $\mathbf{\hat{p}}_i^w = \sigma(\mathbf{s}_i^w / \tau)$ represents the sharpened predicted probability for the weakly augmented view, with the temperature parameter $\tau$ set to 0.1.
\par\noindent
\textit{\textbf{Limitation.}} The regularization encourages the model to learn meaningful representation structures by enforcing prediction consistency under different augmentations for unlabeled data. However, directly applying regularization to the OW-DFA task introduces two key limitations. \textit{First}, in the absence of ground-truth labels for novel categories, the model's predictions tend to be biased toward known categories, and such regularization may further aggravate this bias. \textit{Second}, the threshold-based filtering used in $\mathcal{L}_{ws}$ is sensitive to the choice of the hyperparameter $\delta$, making it inflexible to adapt to varying prediction confidence throughout the training process. 
\par\noindent
To address these limitations, we propose a \textbf{self-calibration weighting} strategy that dynamically rectifies learning strength based on prediction confidence, in a \textbf{threshold-free} fashion. Initially, the model assigns higher weights to high-confidence samples primarily from known forgery types, facilitating the learning of reliable attribution cues from known types. As training progresses, the model gradually shifts emphasis from high-confidence to low-confidence samples, thereby enhancing the learning of novel categories that generally exhibit lower confidence. Specifically, given the sharpened predicted probability $\mathbf{\hat{p}}_i^w$ of the weakly augmented view for the $i$-th face image, we compute its confidence with respect to the most similar category as $\hat{\rho_i} = \max(\mathbf{\hat{p}}_i^w)$. We then assign a weight to each sample using a scheduling parameter that increases linearly with training epochs, defined as $\frac{e}{E}$, where $e$ denotes the current training epoch and $E$ represents the maximum number of training epochs. The final sample weight is computed as:
\begin{equation}
w_i = (1 - \frac{e}{E}) \cdot \hat{\rho_i} + \frac{e}{E} \cdot (1 - \hat{\rho_i}).
\end{equation}
This dynamic weighting scheme enables the model to focus on high-confidence samples from known categories in early training, and gradually shift toward low-confidence samples from novel categories as training progresses. Formally, the CCR loss is:
{\small
\begin{equation}
\mathcal{L}_{ccr} = \frac{1}{|\mathcal{B}|} \sum_{i \in \mathcal{B}} w_i \cdot \ell \bigl(\mathbf{\hat{p}}_i^w, \mathbf{p}_i^s\bigr).
\label{loss:cal}
\end{equation}
}
\normalsize

\par\noindent
\textbf{Asymmetric Confidence Reinforcement (ACR).} 
To further address the confidence skew issue, we propose ACR, which selects and debiases hard pseudo-labels to reinforce the model’s predictions based on the current confidence gap between known and novel classes. Specifically, after a warm-up period of $e_{0}$ epochs, we collect the model’s top-1 predictions and corresponding confidence scores for all unlabeled samples, and compute the average top-$1$ confidence for samples predicted as known and novel classes, respectively. Formally, the model’s average top-$1$ prediction confidence for samples predicted as known classes, denoted as $\overline{\rho_{k}}$, is defined as the mean over the set $\{\rho_{i} = \max(\mathbf{p}_{i}^{w}) \mid \hat{y}_{i} = \arg\max(\mathbf{p}_{i}^{w}) \leq K_L\}$. Similarly, the average top-$1$ confidence for samples predicted as novel classes is denoted as $\overline{\rho_{n}}$. We define the model's confidence gap between known and novel classes as $\overline{\rho_{n}} / \overline{\rho_{k}}$. Given this confidence gap, we apply asymmetric thresholds for selecting hard pseudo-labels for known and novel categories. For samples predicted as known categories, we apply a high confidence threshold $\gamma$. For samples predicted as novel categories, we adopt a relaxed threshold $\overline{\rho_{n}} / \overline{\rho_{k}} \cdot \gamma$. This rule is encoded by the indicator: 
{\small
\begin{equation}
\eta_i =
\mathbb{I}\!\left[
(\hat{y}_i \le K_L \land \rho_i \ge \gamma)\ \lor\
(\hat{y}_i > K_L \land \rho_i \ge \tfrac{\overline{\rho_n}}{\overline{\rho_k}}\cdot\gamma)
\right].
\end{equation}
}
\normalsize

\noindent This adaptive scheme ensures that, even when the confidence for novel classes is relatively low, suitably reliable novel samples are still incorporated during training. As the model gradually improves its understanding of novel categories, the ratio $\overline{\rho_n} / \overline{\rho_k}$ increases adaptively, causing the selection threshold to tighten automatically. Consequently, progressively higher-quality novel samples are fed back into training, steadily closing the confidence gap between known and novel categories. The pseudo-label loss based on the selected pseudo-labels is defined as:
{\small
\begin{equation}
\mathcal{L}_{acr} = \frac{1}{|\mathcal{B}_U|} \sum_{i \in \mathcal{B}_U} \eta_i \cdot \ell\bigl(\hat{y}_i, \mathbf{p}_i\bigr),
\label{loss_cls_sup}
\end{equation}
}
\normalsize

\noindent where $\mathcal{B}_U$ denotes the unlabeled subset in the sampled mini-batch.

\noindent \textbf{Total Loss.}
Following CPL~\cite{cpl}, we adopt a supervised classification loss $\mathcal L_{ce}$ on the labeled dataset $\mathcal{D}_L$ for both weak and strong views,  along with a regularization term $\mathcal R$ on all training data to avoid the trivial solution of assigning all instances to the same class. During the model training process, the total loss is formulated as follows:
{\small
\begin{equation}
\mathcal L = \mathcal L_{ce}+ \mathcal R+  \mathcal L_{acr} +\alpha \cdot \mathcal L_{ccr},
\label{loss_total}
\end{equation}
}
\normalsize	

\noindent where $\alpha$ is the weight of $\mathcal L_{ccr}$.

\subsection{Dynamic Prototype Pruning for Unknown Category Number Estimation}
\label{sec_dpp}

To address scenarios where the number of forgery methods is unknown, we propose a Dynamic Prototype Pruning (DPP) strategy in a multi-stage fashion, as illustrated in~\cref{fig:framework}~(d). DPP dynamically prunes prototypes of novel classes at each training epoch, ultimately enabling automatic estimation of the number of novel forgery types. To accommodate potentially unknown classes, we allocate a large prototype budget $K$, which is much larger than the number of known classes $K_L$. We provide a theoretical analysis of the impact of this setting on the prototypes for novel classes. Let $u_j^{(t)} = \sum_{i=1}^{M} \mathds{1}\bigl[\,\arg\max {\mathbf{p}_{i}^w}^{(t)} = j\bigr]$ denote the usage count of the $j$-$th$ prototype after the $t$-th update. Under conditions such as bounded intra-class noise of novel forgery types, it can be shown that there exist $\varepsilon \in (0,1)$ and an iteration upper bound $T_{0}$ such that, for all $t \le T_{0}$, at least $\bigl(1 - \tfrac{K_U}{K}\bigr)K$ prototypes satisfy $u_j^{(t)} \le \varepsilon M/K$. The formal statement is given in Lemma~4.1, and the proof is deferred to Appendix~A.2.

\noindent\textbf{Lemma 4.1.}
\textit{Assume bounded step sizes, angular cluster separability, and bounded feature noise.  
If the prototype budget satisfies $K\gg K_U$, then there exist $\varepsilon\ll1$ and $T_{0}>0$ such that, for every $t\le T_{0}$,}
{\small
\begin{equation}
\mathbb{E}\!\Bigl[
  \bigl|\{\,j \mid u_j^{(t)} \le \varepsilon M / K \}\bigr|
\Bigr]
\;\ge\;
\Bigl(1 - \tfrac{K_U}{K}\Bigr)K.
\end{equation}
}
\normalsize	

\noindent Based on this analysis, we further propose a coarse-to-fine strategy that dynamically merges redundant low-usage prototypes during training. 

\noindent\textbf{Stage-I: High-confidence Prototype Identification.} 
We perform a coarse partition by selecting high-confidence prototypes and treating the remaining ones as candidate low-confidence prototypes. The index set of novel-class prototypes is initialized as $\mathcal{C}_{novel} = \{K_L+1, \dots, K\}$. The usage counts of novel-class prototypes are sorted in descending order as $\mathcal{U} = \{u_k \mid k \in \mathcal{C}_{novel}\}_{\downarrow}
$. We leverage the cumulative coverage of prototype usage to identify high-confidence prototypes, considering the overall distribution of samples in the unlabeled dataset. The cumulative coverage for the k-$th$ prototype in $\mathcal C_{novel}$ is defined as $r_{k}= \bigl( \sum_{j=1}^{k} u_{j} \bigr)/\bigl(\sum_{u_j \in \mathcal{U}} u_j \bigr).$ Given a target coverage ratio $\mathcal{I}$, the condition $r_k \ge \mathcal{I}$ implies that the top-$k$ prototypes collectively account for at least $\mathcal{I}$ proportion of the samples predicted as novel forgery types. Inspired by the \textit{``$2\sigma$ rule under a normality assumption''}, we set the coverage threshold to $\mathcal{I} = 95.44\%$ to identify a core set of frequently used prototypes, analogous to selecting statistically significant regions in a normal distribution. 
We define the index set of high-confidence prototypes as $\mathcal{C}_{high} = \{j \mid j \le k^{\star}\}$, where $k^{\star} = \min\{k \mid r_k \ge \mathcal{I}\}$. The remaining prototypes are regarded as candidate low-confidence prototypes, denoted as $\mathcal{C}_{cand} = \mathcal{C}_{novel} \setminus \mathcal{C}_{high}$.

\noindent\textbf{Stage-II: Low-confidence Prototype Filtering.} 
We adopt a strict strategy to further filter out potential noisy entries. For all candidate prototypes, we compute the average usage count $\bar{u}$ and the first-order difference of the cumulative coverage, defined as $\Delta r_k = r_k - r_{k-1}$. We then calculate the average first-order difference $\overline{\Delta r}$ across all candidate prototypes. Low-confidence prototypes are defined as those with small usage counts and minimal changes in cumulative coverage, with the corresponding index set defined as $\mathcal{C}_{\text{low}} = \Bigl\{k \in \mathcal{C}_{\text{cand}} \,\Big|\, u_k < \bar{u} \;\land\; \Delta r_k < \overline{\Delta r} \Bigr\}$.

\noindent\textbf{Stage-III: Similarity-driven Prototype Merge.}
To avoid information loss caused by hard pruning, we merge low-confidence prototypes into their most similar high-confidence prototypes. The updated index set of novel-class prototypes $\mathcal{C}_{novel}$ is defined as $\mathcal{C}_{novel} \gets \mathcal{C}_{novel} \setminus  \mathcal{C}_{low}$.

\section{Experiment}
\label{sec_exp}

\begin{table*}[h]
\small
\setlength{\tabcolsep}{0.28mm}

\label{tab:comparison_ourdataset}
\begin{center}

\begin{tabular}{lccccccccccccccccccccc}
\toprule
\multicolumn{1}{c}{\multirow{3}{*}{\textbf{Method}}} &
\multicolumn{7}{c}{\textbf{Protocol-1}} & 
\multicolumn{7}{c}{\textbf{Protocol-2}} & \multicolumn{7}{c}{\textbf{Protocol-3}} \\

\cmidrule(lr){2-8} \cmidrule(lr){9-15} \cmidrule(lr){16-22}
& \multicolumn{3}{c}{\textbf{All}} & \multicolumn{3}{c}{\textbf{New}} & \textbf{Known} 
  & \multicolumn{3}{c}{\textbf{All}} & \multicolumn{3}{c}{\textbf{New}} & \textbf{Known} 
  & \multicolumn{3}{c}{\textbf{All}} & \multicolumn{3}{c}{\textbf{New}} & \textbf{Known} \\
  
\cmidrule(lr){2-4} \cmidrule(lr){5-7} \cmidrule(lr){8-8} \cmidrule(lr){9-11} \cmidrule(lr){12-14} \cmidrule(lr){15-15} \cmidrule(lr){16-18} \cmidrule(lr){19-21} \cmidrule(lr){22-22}

& \textbf{ACC} & \textbf{NMI} & \textbf{ARI} & \textbf{ACC} & \textbf{NMI} & \textbf{ARI} & \textbf{ACC} 
  & \textbf{ACC} & \textbf{NMI} & \textbf{ARI} & \textbf{ACC} & \textbf{NMI} & \textbf{ARI} & \textbf{ACC} 
  & \textbf{ACC} & \textbf{NMI} & \textbf{ARI} & \textbf{ACC} & \textbf{NMI} & \textbf{ARI} & \textbf{ACC}\\
  
\midrule

SimGCD     & 73.3 & 82.5 & 54.5 & 65.0 & 75.0 & 57.6 & 81.7 & 74.3 & 81.3 & 68.4 & 62.0 & 76.1 & 55.8 & 86.4 & 85.6 & 86.3 & 82.9 & 78.5 & 79.7 & 72.3 & 90.4\\

OwMatch  & 74.1 & 82.1 & 81.7 & 61.2 & 74.3 & 53.6 & 96.2 & 72.6 & 80.1 & 64.2 & 61.2 & 74.7 & 54.4 & 84.1 & 84.3 & 88.4 & 87.9 & 66 & 77.6 & 63.3 & 94.8\\

LPS    & 76.1 & 79.0 & 75.4 & 61.2 & 69.5 & 53.1 & 90.9 & 75.1& 79.2 & 78.5 & 59.9 & 73.7 & 53.2 & 92.8 & 80.0 & 86.9 & 85.9 & 57.8 & 70.5 & 52.5 & 92.5\\

LegoGCD    & 70.9 & 76.3 & 57.6 & 61.0 & 69.2 & 52.9 & 82.6 & 72.2 & 81.0 & 58.3 & 64.5 & 77.1 & 57.5 & 80.8 & 82.0 & 81.9 & 75.3 & 68.6 & 74.8 & 64.8 & 86.8\\

ProtoGCD   & 62.9 & 79.7 & 37.7 & 59.8 & 71.1 & 51.3 & 66.8 & 61.8 & 70.0 & 41.5 & 58.1 & 68.1 & 48.7 & 69.3 & 78.2 & 81.0 & 63.4 & 73.1 & 75.9 & 67.3 & 82.1\\

PALGCD    & 76.5 & 80.0 &79.0 &54.2 &66.5 &45.1 &95.9 &77.6 &80.6 &79.2 &65.0 &75.4 &56.9 &92.3 &83.3 &82.5 &79.3 &63.0 &71.4 &61.7 &89.5 \\

\midrule

CPL & 82.6 & 85.5 & 85.1 & 64.3 & 73.2 & 56.3 & 96.2 & 79.6 & 83.7 & 86.3 & 63.6 & 75.3 & 55.3 & 95.0 & 86.3 & 88.1 & 79.2 & 74.4 & 79.4 & 70.4 & 90.3\\

CDAL & 84.3 & 87.4 & 86.0 & 70.0 & 77.7 & 62.5 & 95.3 & \underline{80.2} & 83.9 & 85.7 & \underline{65.7} & 76.2 & 57.3 & 94.1 & 87.3 & 88.7 & 83.4 & 74.4 & 78.3 & 68.7 & 91.6\\
\midrule
Ours (w/o $K_U$)         & \underline{87.5} & \underline{89.7} & \underline{90.2} & \underline{74.5} & \textbf{82.3} & \underline{70.9} & \textbf{98.2} & 80.0 & \underline{86.1} & \underline{88.7} & 63.0 & \underline{78.2} & \underline{59.1} & \underline{97.5} & \textbf{91.5} & \textbf{92.1} & \textbf{91.7} & \textbf{83.2} & \textbf{85.4} & \textbf{79.6} & \textbf{95.0}\\
Ours                     & \textbf{88.3} & \textbf{90.3} & \textbf{91.5} & \textbf{76.5} & \underline{82.2} & \textbf{72.4} & \underline{98.0} & \textbf{82.8} & \textbf{87.2} & \textbf{90.0} & \textbf{67.5} & \textbf{80.2} & \textbf{63.5} & \textbf{97.6} & \underline{91.0} & \underline{91.8} & \underline{91.6} & \underline{80.3} & \underline{83.9} & \underline{77.3} & \underline{94.8}\\

\bottomrule
\end{tabular}
\end{center}
\caption{Results on the OW-DFA-40 benchmark. Best results are in \textbf{bold}, second-best are \underline{underlined}.}
\end{table*}

\subsection{Experiment Setup}

\noindent \textbf{OW-DFA-40 Benchmark.} 
We construct a new OW-DFA-40 benchmark comprising 40 deepfake generation methods covering five mainstream facial forgery categories. In addition to the 20 methods from the original OW-DFA benchmark~\cite{cpl}, we include 20 newly added state-of-the-art techniques spanning Face Swapping~\shortcite{mobileswap, uniface, blendface, e4s, facedancer}, Face Reenactment~\shortcite{oneshot, tpsmm, lia, dagan, sadtalker, mcnet, hyperreenact}, Face Editing~\shortcite{e4e}, Entire Face Synthesis~\shortcite{stylegan3, styleganxl, sd2.1}, and Diffusion-based Generation~\shortcite{ditxl, rddm, pixart, sit}. These fake face images are generated using real facial data from two widely-used datasets: FaceForensics++~\cite{ff++} and Celeb-DF~\cite{celebdf}. Further construction details are provided in Appendix~B.1.

\noindent\textbf{Evaluation Protocol.} We define three evaluation protocols. \textbf{Protocol-1} includes 40 forgery methods—19 known and 22 unknown—along with real face data, yielding a total of 41 attribution classes. \textbf{Protocol-2} builds on Protocol-1 by treating all methods under \textit{Entire Face Synthesis} and \textit{Diffusion-based Generation} as unknown, resulting in 13 known and 28 unknown classes. This setup simulates the emergence of new attack paradigms. \textbf{Protocol-3} also extends Protocol-1, incorporating more known methods across forgery categories, leading to 29 known and 12 unknown classes. Detailed specifications of each protocol are provided in Appendix~B.1. For all protocols, we follow the data split strategy of the original OW-DFA benchmark: 80\% of the data is used for training and 20\% for testing, forming the labeled dataset $\mathcal{D}_{L}$ and the unlabeled dataset $\mathcal{D}_{U}$. Unlike the original OW-DFA benchmark, we unify real face data from FaceForensics++ and Celeb-DF into a single category, denoted as \textit{real}, which is included as an attribution class in each protocol. Experiments are conducted on both the OW-DFA-40 benchmark and the original OW-DFA benchmark.

\noindent \textbf{Evaluation Metrics.} Following the OW-DFA benchmark, we use three standard metrics to evaluate the performance of all evaluated methods: Accuracy (ACC), Normalized Mutual Information (NMI), and Adjusted Rand Index (ARI). Predicted labels are aligned with the ground-truth class labels via an optimal permutation computed using the Hungarian algorithm~\cite{Hungarian}. Further details on metric computation are provided in Appendix~B.2.

\noindent \textbf{Implementation Details.}
For a fair comparison, we use a ResNet-50 pre-trained on ImageNet~\cite{imagenet} as the image encoder, following CPL~\cite{cpl}. For face preprocessing, we resize all input images to $256 \times 256$ and detect faces using dlib. The optimizer is Adam with a learning rate of $2 \times 10^{-4}$. All models are trained for 50 epochs with a batch size of 128. The module $\mathcal G(\cdot)$ is implemented as a lightweight convolutional network with two layers of $3\times3$ kernels. The warm-up epoch for $\mathcal{L}_{acr}$ is set to $e_0 = 5$. The weak augmentation includes only \texttt{RandomHorizontalFlip} with a probability of 0.5, while the strong augmentation combines \texttt{RandomHorizontalFlip}, \texttt{RandomResizedCrop}, and brightness adjustment, each applied with a probability of 0.2. The initial number of attribution prototypes is set to $K = 10 \times K_L$. We set the hyperparameters to $\alpha = 0.2$ and $\gamma = 0.9$, determined based on the labeled dataset under Protocol-1. To avoid complex fine-tuning, we use this hyperparameter setting for all experiments. All models are trained on a single NVIDIA A100 GPU, and results are averaged over three runs using random seeds $\{0, 1, 2\}$. Additional details and pseudo-code of the training process are provided in Appendix~C.1.

\noindent \textbf{Compared Methods.} 
We compare our method with CPL~\cite{cpl} and CDAL~\cite{cdal}, along with strong baselines from the Generalized Category Discovery (GCD) and Open-World Semi-Supervised Learning (OWSSL) settings. Specifically, we include SimGCD~\cite{simgcd}, LegoGCD~\cite{legogcd}, and ProtoGCD~\cite{protogcd} from GCD, as well as Owmatch~\cite{owmatch}, LPS~\cite{lps}, and PALGCD~\cite{palgcd} from OWSSL. All compared methods are configured according to the settings of CPL~\cite{cpl}. Additionally, these baseline methods require prior knowledge of the true number of forgery types, which are set to the ground-truth number of classes. Detailed implementation settings are provided in Appendix~C.2.

\subsection{Comparison with State of the Art}
\noindent \textbf{Results on OW-DFA-40.}
We compare our method with the aforementioned baselines on the OW-DFA-40 benchmark, as shown in~\cref{tab:comparison_ourdataset}. The proposed CAL consistently outperforms all state-of-the-art competitors across nearly all metrics. Compared to CPL~\cite{cpl} and CDAL~\cite{cdal}, CAL achieves average All ACC improvements of 4.5\% and 3.4\%, respectively. More importantly, CAL surpasses CPL by an average of 7.3\% in Novel ACC across all protocols, indicating its superior ability to discover novel deepfake methods and to mitigate model bias toward known classes. \textit{Even when the number of novel forgery types \textbf{$K_U$ is unknown}, our CAL still outperforms CPL by an average of 3.5\% in All ACC, demonstrating its applicability in real open-world scenarios.} OWSSL and GCD methods are primarily designed for capturing global visual similarity in natural images, making them less effective for deepfake attribution. Specifically, compared to two state-of-the-art OWSSL methods, CAL improves All ACC by 10.4\% over OwMatch and 10.3\% over LPS, on average across all protocols. Furthermore, CAL surpasses four GCD methods in average All ACC: by 9.6\% over SimGCD, 19.7\% over ProtoGCD, 12.3\% over LegoGCD, and 8.2\% over PALGCD.

\noindent \textbf{Results on OW-DFA~\cite{cpl}.}
We further evaluate our method on the original OW-DFA benchmark (see Appendix~D.1), where CAL achieves an average All ACC improvement of 3.4\% and a notable 7.1\% gain in Novel ACC over CPL across the two evaluation protocols.

\subsection{Ablation Study}

\noindent \textbf{Ablation on Components.}
We conduct an ablation study of the proposed components in our CAL on Protocol-1 of the OW-DFA-40 benchmark, as shown in~\cref{tab:ablation1}. The components under evaluation include the base loss $\mathcal{L}_{ce} + \mathcal{R}$, our proposed Confidence-Aware Consistency Regularization loss $\mathcal{L}_{ccr}$, our Asymmetric Confidence Reinforcement loss $\mathcal{L}_{acr}$, and Frequency-Guided Feature Enhancement (FFE) module. Models I, III, and V illustrate the step-by-step addition of training components. Using only the base loss enables the model to effectively learn known classes, but yields only 33.2\% in Novel ACC. Adding $\mathcal{L}_{ccr}$ results in an 18.4\% improvement in All ACC and a substantial 32.8\% improvement in Novel ACC, highlighting the critical role of self-calibration weighting in facilitating novel deepfake discovery. Building on Model III, incorporating $\mathcal{L}_{acr}$ further enhances performance by 3.7\% in All ACC and 10.5\% in Novel ACC. 
The comparisons between Models II and III, as well as between Models IV and V, demonstrate the effect of FFE. The inclusion of FFE yields an average improvement of 2.7\% in All ACC in both comparisons, underscoring the benefit of incorporating frequency-domain information and suggesting that different types of deepfake methods generate images with distinctive frequency characteristics. 
\begin{table}[t]
  \centering
    \setlength{\tabcolsep}{0.5mm}
    \small
    \begin{tabular}{cccccccccccc}
      \toprule
      \multirow{2}{*}{}
      &\multirow{1}{*}{$\mathcal{L}_{ce}$} & 
      \multirow{2}{*}{$\mathcal{L}_{ccr}$} & 
      \multirow{2}{*}{$\mathcal{L}_{acr}$} & 
      \multirow{2}{*}{FFE} & 
      \multicolumn{3}{c}{All} & 
      \multicolumn{3}{c}{New} & 
      Known \\
      \cmidrule(lr){6-8} \cmidrule(lr){9-11} \cmidrule(lr){12-12}
      & $+\mathcal R$&  & & & ACC & NMI & ARI & ACC & NMI & ARI & ACC \\
      
      \midrule
      I &\checkmark & & & \checkmark & 66.2 & 71.3 & 74.5 & 33.2 & 43.0 & 18.8 & 95.6 \\
      II &\checkmark & \checkmark & &  & 80.1 & 83.3 & 85.6 & 54.8 & 68.4 & 47.3 & 98.1 \\
      III &\checkmark & \checkmark & & \checkmark & 84.6 & 87.9 & 89.6 & 66.0 & 76.5 & 61.5 & 98.3 \\
      IV &\checkmark & \checkmark & \checkmark &  & 87.4 & 88.9 & 90.5 & 73.4 & 79.3 & 69.9 & \textbf{98.6}  \\
      V &\checkmark & \checkmark & \checkmark & \checkmark & \textbf{88.3} & \textbf{90.3} & \textbf{91.5} & \textbf{76.5} & \textbf{82.2} & \textbf{72.4} & 98.0 \\
      \bottomrule
    \end{tabular}
    \caption{Ablation study on training components. The best results are marked in \textbf{bold}.}
    \label{tab:ablation1}
\end{table}

\noindent \textbf{Ablation on Strategies.}
In~\cref{tab:ablation2}, we evaluate the impact of different training strategies used in our method. 
``FixMatch'' refers to a variant where the threshold-based loss $\mathcal{L}_{ws}$ in~\cref{loss:ws} replaces our self-calibration weighting–based loss $\mathcal{L}_{ccr}$. In this variant, the threshold $\delta$ is set to 0.5. 
``Fixed $\gamma$'' refers to using the same confidence threshold $\gamma$ as that for known classes to perform hard pseudo-label selection for novel classes. For the first variant, our method achieves 1.5\% higher All ACC and, notably, 4.5\% higher Novel ACC, demonstrating that the self-calibration weighting strategy is more effective than the fixed-threshold approach of $\mathcal{L}_{ws}$ in enhancing the learning of novel-class samples. 
For the second variant, we observe a 2.4\% gain in All ACC and a 3.6\% improvement in Novel ACC, confirming that our asymmetric pseudo-labeling strategy outperforms the symmetric counterpart and is more beneficial for learning novel forgery types.

\subsection{Evaluation}
\label{sec:evaluation}

\begin{figure}[h]
  \centering
  \includegraphics[width=\linewidth]{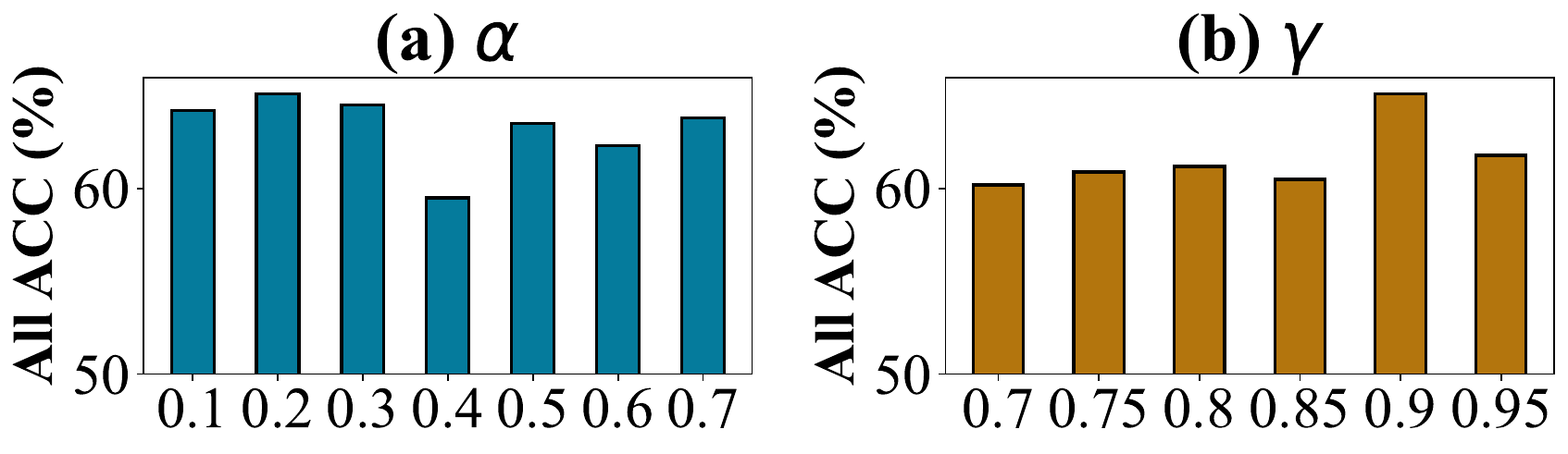}
\caption{Selection of hyper-parameters. \label{fig:hyper_main}}
\end{figure}

\noindent \textbf{Hyper-parameter Selection.} 
Our CAL framework involves two hyper-parameters: $\alpha$, the weight of the $\mathcal{L}_{ccr}$ loss, and $\gamma$, the confidence threshold for pseudo-label selection in ACR. We select these hyper-parameters using the labeled dataset from Protocol-1. Specifically, we split the labeled dataset into a new labeled sub-dataset and a new unlabeled sub-dataset, following the same partitioning strategy. The 19 classes are divided into 9 known and 10 unknown classes, and we use All ACC on the unlabeled sub-dataset as the evaluation metric. As shown in~\cref{fig:hyper_main}, the best performance is achieved when $\alpha=0.2$ and $\gamma=0.9$. 
We also evaluate these hyper-parameters across all three protocols, demonstrating that CAL is relatively robust to variations in $\alpha$ and $\gamma$ (see Appendix~D.2).

\noindent \textbf{Evaluation of Class Number Estimation.} 
We compare our DPP strategy with the class number estimation method proposed in GCD~\cite{gcd}, as shown in~\cref{tab:esk_gcd}. 
GCD treats clustering accuracy on the labeled dataset as a black-box scoring function and selects the class number that yields the highest accuracy as the final estimate—a strategy widely adopted in both GCD and OWSSL tasks. 
For a fair comparison, we evaluate both methods using DINO-ViT-B and an ImageNet-pretrained ResNet-50 as backbones. 
Our DPP method consistently achieves lower estimation errors across all protocols. 
In contrast, GCD’s approach heavily depends on the pretrained backbone’s ability to extract meaningful features from unknown classes, which proves inadequate in the OW-DFA setting. 
We further illustrate how the estimated class number evolves over training epochs in~\cref{fig:estk_main}, and provide a more detailed discussion on the robustness of DPP in Appendix~D.3.
\begin{figure}[h]
  \centering
  \includegraphics[width=0.95\linewidth]{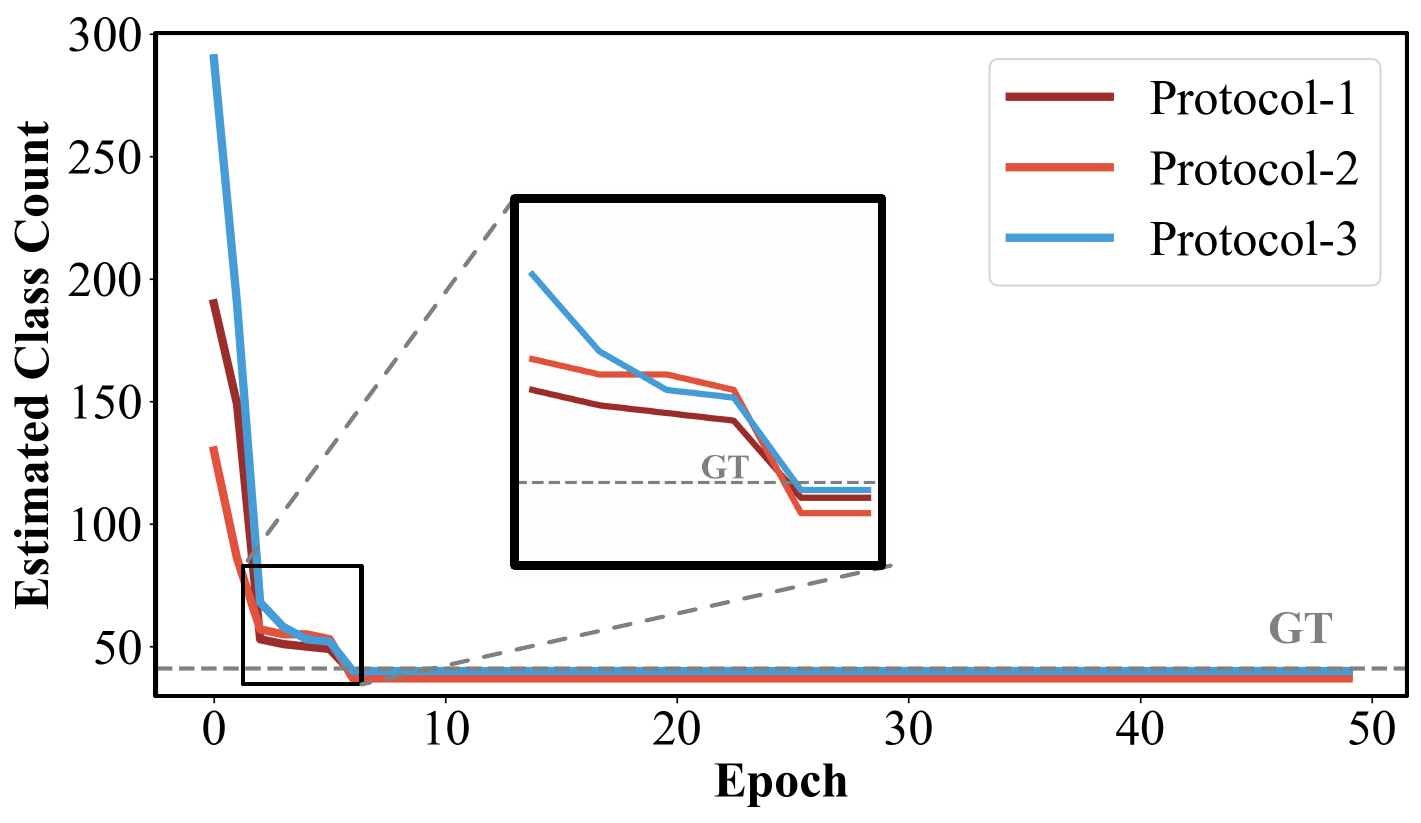}
\caption{Estimated class count over training epochs. \label{fig:estk_main}}
\end{figure}

\begin{table}[!t]
  \centering
    \setlength{\tabcolsep}{1mm}

\label{tab:ablation2}
\setlength{\tabcolsep}{2pt}
\renewcommand{\arraystretch}{1.21}
\small
  \begin{tabular}{lccccccc}
    \toprule
    \multirow{2}{*}{Method} & 
    \multicolumn{3}{c}{All} & 
    \multicolumn{3}{c}{New} & 
    Known \\
    \cmidrule(lr){2-4} \cmidrule(lr){5-7} \cmidrule(lr){8-8}
    & ACC & NMI & ARI & ACC & NMI & ARI & ACC \\
    \midrule
    FixMatch   & 86.8 & 89.4 & 90.0 & 72.0 & 81.6 & 68.9 & 97.9   \\
    Fixed $\gamma$  & 85.9 & 89.0 & 89.8 & 72.9 & 80.3 & 69.8 & 97.7   \\
    Ours & \textbf{88.3} & \textbf{90.3} & \textbf{91.5} & \textbf{76.5} & \textbf{82.2} & \textbf{72.4} & \textbf{98.0}       \\
    \bottomrule
  \end{tabular}
\caption{Ablation study on training strategies. The best results are marked in \textbf{bold}.}
\end{table}

\begin{table}[!t]
  \centering

  \setlength{\tabcolsep}{1mm}
    \centering
    \begin{tabular}{lccccccc}
      \toprule
      \multirow{2}{*}{Method} & 
      \multirow{2}{*}{backbone} &
      \multicolumn{2}{c}{\textbf{Protocol-1}} & 
      \multicolumn{2}{c}{\textbf{Protocol-2}} &
      \multicolumn{2}{c}{\textbf{Protocol-3}}  \\
      \cmidrule(lr){3-4} \cmidrule(lr){5-6} \cmidrule(lr){7-8} 
      & & K & Err(\%)  & K & Err(\%) & K & Err(\%) \\
      \midrule
      GCD & DINO-ViT-B  &19 & 53.7 &13 & 68.3 & 33&  19.5  \\
      GCD & ResNet-50  & 20 & 51.2 &13 & 68.3 &35 & 14.6   \\
      Ours & ResNet-50 & 39 & \textbf{4.8} & 37 & \textbf{9.7} &40 & \textbf{2.4} \\
      \bottomrule
    \end{tabular}
    \label{tab:esk_gcd}
\caption{Comparison with class number estimation method from GCD.}
\end{table}

\section{Conclusion}

We propose Confidence-Aware Asymmetric Learning (CAL), a novel framework for Open-World DeepFake Attribution (OW-DFA), which addresses key challenges of confidence skew and unknown category estimation. By combining Confidence-Aware Consistency Regularization (CCR) for adaptive pseudo-label regularization and Asymmetric Confidence Reinforcement (ACR) for asymmetric confidence calibration, CAL effectively mitigates bias toward known forgery types. Furthermore, the proposed Dynamic Prototype Pruning (DPP) strategy enables dynamic estimation of novel categories without knowing the number of novel forgery types, improving the scalability of CAL. Experiments on standard and extended OW-DFA benchmarks show that CAL achieves state-of-the-art performance on both known and novel forgeries.

\noindent\textbf{Acknowledgment}.  This work was funded by the National Natural Science Foundation of China ( No. 62572166 \& No. 62402157) and the Fundamental Research Funds for the Central Universities (No. JZ2025HGTB0219). This work was supported by the EU Horizon project “ELIAS - European Lighthouse of AI for Sustainability” (No. 101120237) and the FIS project GUIDANCE (Debugging Computer Vision Models via Controlled Cross-modal
Generation) (No. FIS2023-03251). We also acknowledge CINECA and the ISCRA initiative for providing high-performance computing resources.

\bibliography{aaai2026}

@String(PAMI  = {IEEE TPAMI})

@String(IJCV  = {IJCV})

@String(CVPR  = {CVPR})

@String(ICCV  = {ICCV})

@String(ECCV  = {ECCV})

@String(NIPS  = {NeurIPS})

@String(ICLR  = {ICLR})

@String(IJCAI = {IJCAI})

@String(AAAI = {AAAI})

@string(icml = {ICML})

@string(miccai = {MICCAI})

@article{cao2021open,
  title={Open-world semi-supervised learning},
  author={Cao, Kaidi and Brbic, Maria and Leskovec, Jure},
  journal={arXiv preprint arXiv:2102.03526},
  year={2021}
}

@inproceedings{openldn,
  title={Openldn: Learning to discover novel classes for open-world semi-supervised learning},
  author={Rizve, Mamshad Nayeem and Kardan, Navid and Khan, Salman and Shahbaz Khan, Fahad and Shah, Mubarak},
  booktitle={ECCV},
  @@pages={382--401},
  year={2022},
  @organization={Springer}
}

@article{sun2022opencon,
  title={Opencon: Open-world contrastive learning},
  author={Sun, Yiyou and Li, Yixuan},
  journal={arXiv preprint arXiv:2208.02764},
  year={2022}
}

@article{wang2023discover,
  title={Discover and align taxonomic context priors for open-world semi-supervised learning},
  author={Wang, Yu and Zhong, Zhun and Qiao, Pengchong and Cheng, Xuxin and Zheng, Xiawu and Liu, Chang and Sebe, Nicu and Ji, Rongrong and Chen, Jie},
  journal={NeurIPS},
  @volume={36},
  @pages={19015--19028},
  year={2023}
}

@inproceedings{rombach2022high,
  title={High-resolution image synthesis with latent diffusion models},
  author={Rombach, Robin and Blattmann, Andreas and Lorenz, Dominik and Esser, Patrick and Ommer, Bj{\"o}rn},
  booktitle=CVPR,
  @@pages={10684--10695},
  year={2022}
}

@inproceedings{yang2022deepfake,
  title={Deepfake network architecture attribution},
  author={Yang, Tianyun and Huang, Ziyao and Cao, Juan and Li, Lei and Li, Xirong},
  booktitle=AAAI,
  @@volume={36},
  @@number={4},
  @@pages={4662--4670},
  year={2022}
}

@inproceedings{yu2021artificial,
  title={Artificial fingerprinting for generative models: Rooting deepfake attribution in training data},
  author={Yu, Ning and Skripniuk, Vladislav and Abdelnabi, Sahar and Fritz, Mario},
  booktitle=ICCV,
  @@pages={14448--14457},
  year={2021}
}

@inproceedings{guarnera2022exploitation,
  title={On the exploitation of deepfake model recognition},
  author={Guarnera, Luca and Giudice, Oliver and Nie{\ss}ner, Matthias and Battiato, Sebastiano},
  booktitle=CVPR,
  @@pages={61--70},
  year={2022}
}

@misc{Deepfacelab,
  title        = {Deepfacelab},
  @author       = {CiteDrive, Inc},
  year         = 2024,
  @note         = {访问日期: (使用访问日期)},
  howpublished = {\url{https://github.com/iperov/DeepFaceLab}}
}

@misc{Deepfakes,
  title        = {Deepfakes},
  @author       = {CiteDrive, Inc},
  year         = 2024,
  @note         = {访问日期: (使用访问日期)},
  howpublished = {\url{https://github.com/deepfakes/faceswap}}
}

@misc{faceswap,
  author       = {{Marek Kowalski}},
  title        = {{FaceSwap}},
  howpublished = {\url{https://github.com/MarekKowalski/FaceSwap/}},
  note         = {Accessed: July 6, 2024},
  year         = {2024}
}

@misc{midjourney2024,
  title        = {Midjourney},
  author       = {CiteDrive, Inc},
  year         = 2024,
  @note         = {访问日期: (使用访问日期)},
  howpublished = {\url{https://www.midjourney.com/home/}}
}

@article{sun2025rethinking,
  title={Rethinking open-world deepfake attribution with multi-perspective sensory learning},
  author={Sun, Zhimin and Chen, Shen and Yao, Taiping and Yi, Ran and Ding, Shouhong and Ma, Lizhuang},
  journal=IJCV,
  @@volume={133},
  @@number={2},
  @@pages={628--651},
  year={2025},
  @publisher={Springer}
}

@article{tian2024visual,
  title={Visual autoregressive modeling: Scalable image generation via next-scale prediction},
  author={Tian, Keyu and Jiang, Yi and Yuan, Zehuan and Peng, Bingyue and Wang, Liwei},
  journal=NIPS,
  @@volume={37},
  @@pages={84839--84865},
  year={2024}
}

@inproceedings{fsgan,
  title={Fsgan: Subject agnostic face swapping and reenactment},
  author={Nirkin, Yuval and Keller, Yosi and Hassner, Tal},
  booktitle={ICCV},
  @pages={7184--7193},
  year={2019}
}

@article{faceshifter,
  title={Faceshifter: Towards high fidelity and occlusion aware face swapping},
  author={Li, Lingzhi and Bao, Jianmin and Yang, Hao and Chen, Dong and Wen, Fang},
  journal={CVPR},
  year={2020}
}

@inproceedings{ff++,
  title={Faceforensics++: Learning to detect manipulated facial images},
  author={Rossler, Andreas and Cozzolino, Davide and Verdoliva, Luisa and Riess, Christian and Thies, Justus and Nie{\ss}ner, Matthias},
  booktitle={CVPR},
  @@pages={1--11},
  year={2019}
}

@inproceedings{forgerynet,
  title={Forgerynet: A versatile benchmark for comprehensive forgery analysis},
  author={He, Yinan and Gan, Bei and Chen, Siyu and Zhou, Yichun and Yin, Guojun and Song, Luchuan and Sheng, Lu and Shao, Jing and Liu, Ziwei},
  booktitle={CVPR},
  @@pages={4360--4369},
  year={2021}
}

@inproceedings{face2face,
  title={Face2face: Real-time face capture and reenactment of rgb videos},
  author={Thies, Justus and Zollhofer, Michael and Stamminger, Marc and Theobalt, Christian and Nie{\ss}ner, Matthias},
  booktitle={CVPR},
  @@pages={2387--2395},
  year={2016}
}

@article{fomm,
  title={First order motion model for image animation},
  author={Siarohin, Aliaksandr and Lathuili{\`e}re, St{\'e}phane and Tulyakov, Sergey and Ricci, Elisa and Sebe, Nicu},
  journal={NeurIPS},
  @@volume={32},
  year={2019}
}

@article{Neuraltextures,
  title={Deferred neural rendering: Image synthesis using neural textures},
  author={Thies, Justus and Zollh{\"o}fer, Michael and Nie{\ss}ner, Matthias},
  journal={SIGGRAPH},
  @@volume={38},
  @@number={4},
  @@pages={1--12},
  year={2019},
  @publisher={ACM New York, NY, USA}
}

@inproceedings{ATVGNet,
  title={Hierarchical cross-modal talking face generation with dynamic pixel-wise loss},
  author={Chen, Lele and Maddox, Ross K and Duan, Zhiyao and Xu, Chenliang},
  booktitle={CVPR},
  @@pages={7832--7841},
  year={2019}
}

@inproceedings{maskgan,
  title={Maskgan: Towards diverse and interactive facial image manipulation},
  author={Lee, Cheng-Han and Liu, Ziwei and Wu, Lingyun and Luo, Ping},
  booktitle={CVPR},
  @@pages={5549--5558},
  year={2020}
}

@misc{faceapp,
  author       = {{FaceApp Inc.}},
  title        = {{FaceApp}},
  howpublished = {\url{https://faceapp.com/app}},
  note         = {Accessed: February 28, 2023},
  year         = {2023}
}

@inproceedings{stargan,
  title={Stargan: Unified generative adversarial networks for multi-domain image-to-image translation},
  author={Choi, Yunjey and Choi, Minje and Kim, Munyoung and Ha, Jung-Woo and Kim, Sunghun and Choo, Jaegul},
  booktitle={CVPR},
  @@pages={8789--8797},
  year={2018}
}

@inproceedings{starganv2,
  title={Stargan v2: Diverse image synthesis for multiple domains},
  author={Choi, Yunjey and Uh, Youngjung and Yoo, Jaejun and Ha, Jung-Woo},
  booktitle={CVPR},
  @@pages={8188--8197},
  year={2020}
}

@inproceedings{scfegan,
  title={SC-FEGAN: Face editing generative adversarial network with user's sketch and color},
  author={Jo, Youngjoo and Park, Jongyoul},
  booktitle={ICCV},
  @@pages={1745--1753},
  year={2019}
}

@inproceedings{dffd,
  title={On the detection of digital face manipulation},
  author={Dang, Hao and Liu, Feng and Stehouwer, Joel and Liu, Xiaoming and Jain, Anil K},
  booktitle={CVPR},
  @@pages={5781--5790},
  year={2020}
}

@inproceedings{stylegan,
  title={A style-based generator architecture for generative adversarial networks},
  author={Karras, Tero and Laine, Samuli and Aila, Timo},
  booktitle={CVPR},
  @@pages={4401--4410},
  year={2019}
}

@inproceedings{styleganv2,
  title={Analyzing and improving the image quality of stylegan},
  author={Karras, Tero and Laine, Samuli and Aittala, Miika and Hellsten, Janne and Lehtinen, Jaakko and Aila, Timo},
  booktitle={CVPR},
  @@pages={8110--8119},
  year={2020}
}

@inproceedings{cyclegan,
  title={Unpaired image-to-image translation using cycle-consistent adversarial networks},
  author={Zhu, Jun-Yan and Park, Taesung and Isola, Phillip and Efros, Alexei A},
  booktitle={ICCV},
  @@pages={2223--2232},
  year={2017}
}

@article{pggan,
  title={Progressive growing of gans for improved quality, stability, and variation},
  author={Karras, Tero and Aila, Timo and Laine, Samuli and Lehtinen, Jaakko},
  journal={ICLR},
  year={2018}
}

@inproceedings{celebdf,
  title={Celeb-df: A large-scale challenging dataset for deepfake forensics},
  author={Li, Yuezun and Yang, Xin and Sun, Pu and Qi, Honggang and Lyu, Siwei},
  booktitle={CVPR},
  @@pages={3207--3216},
  year={2020}
}

@inproceedings{mobileswap,
  title={Mobilefaceswap: A lightweight framework for video face swapping},
  author={Xu, Zhiliang and Hong, Zhibin and Ding, Changxing and Zhu, Zhen and Han, Junyu and Liu, Jingtuo and Ding, Errui},
  booktitle={AAAI},
  @@volume={36},
  @@number={3},
  @@pages={2973--2981},
  year={2022}
}

@inproceedings{uniface,
  title={Designing one unified framework for high-fidelity face reenactment and swapping},
  author={Xu, Chao and Zhang, Jiangning and Han, Yue and Tian, Guanzhong and Zeng, Xianfang and Tai, Ying and Wang, Yabiao and Wang, Chengjie and Liu, Yong},
  booktitle={ECCV},
  @@pages={54--71},
  year={2022},
  @organization={Springer}
}

@inproceedings{blendface,
  title={Blendface: Re-designing identity encoders for face-swapping},
  author={Shiohara, Kaede and Yang, Xingchao and Taketomi, Takafumi},
  booktitle={ICCV},
  @@pages={7634--7644},
  year={2023}
}

@inproceedings{e4s,
  title={Fine-grained face swapping via regional gan inversion},
  author={Liu, Zhian and Li, Maomao and Zhang, Yong and Wang, Cairong and Zhang, Qi and Wang, Jue and Nie, Yongwei},
  booktitle={CVPR},
  @@pages={8578--8587},
  year={2023}
}

@inproceedings{facedancer,
  title={Facedancer: Pose-and occlusion-aware high fidelity face swapping},
  author={Rosberg, Felix and Aksoy, Eren Erdal and Alonso-Fernandez, Fernando and Englund, Cristofer},
  booktitle={WACV},
  @@pages={3454--3463},
  year={2023}
}

@inproceedings{oneshot,
  title={One-shot free-view neural talking-head synthesis for video conferencing},
  author={Wang, Ting-Chun and Mallya, Arun and Liu, Ming-Yu},
  booktitle={CVPR},
  @@pages={10039--10049},
  year={2021}
}

@inproceedings{tpsmm,
  title={Thin-plate spline motion model for image animation},
  author={Zhao, Jian and Zhang, Hui},
  booktitle={CVPR},
  @@pages={3657--3666},
  year={2022}
}

@article{lia,
  title={Latent image animator: Learning to animate images via latent space navigation},
  author={Wang, Yaohui and Yang, Di and Bremond, Francois and Dantcheva, Antitza},
  journal={ICLR},
  year={2022}
}

@inproceedings{dagan,
  title={Depth-aware generative adversarial network for talking head video generation},
  author={Hong, Fa-Ting and Zhang, Longhao and Shen, Li and Xu, Dan},
  booktitle={CVPR},
  @@pages={3397--3406},
  year={2022}
}

@inproceedings{sadtalker,
  title={Sadtalker: Learning realistic 3d motion coefficients for stylized audio-driven single image talking face animation},
  author={Zhang, Wenxuan and Cun, Xiaodong and Wang, Xuan and Zhang, Yong and Shen, Xi and Guo, Yu and Shan, Ying and Wang, Fei},
  booktitle={CVPR},
  @@pages={8652--8661},
  year={2023}
}

@inproceedings{mcnet,
  title={Implicit identity representation conditioned memory compensation network for talking head video generation},
  author={Hong, Fa-Ting and Xu, Dan},
  booktitle={ICCV},
  @@pages={23062--23072},
  year={2023}
}

@inproceedings{hyperreenact,
  title={Hyperreenact: one-shot reenactment via jointly learning to refine and retarget faces},
  author={Bounareli, Stella and Tzelepis, Christos and Argyriou, Vasileios and Patras, Ioannis and Tzimiropoulos, Georgios},
  booktitle={ICCV},
  @@pages={7149--7159},
  year={2023}
}

@article{e4e,
  title={Designing an encoder for stylegan image manipulation},
  author={Tov, Omer and Alaluf, Yuval and Nitzan, Yotam and Patashnik, Or and Cohen-Or, Daniel},
  journal={SIGGRAPH},
  @@volume={40},
  @@number={4},
  @@pages={1--14},
  year={2021},
  @publisher={ACM New York, NY, USA}
}

@article{stylegan3,
  title={Alias-free generative adversarial networks},
  author={Karras, Tero and Aittala, Miika and Laine, Samuli and H{\"a}rk{\"o}nen, Erik and Hellsten, Janne and Lehtinen, Jaakko and Aila, Timo},
  journal={NeurIPS},
  @@volume={34},
  @@pages={852--863},
  year={2021}
}

@inproceedings{styleganxl,
  title={Stylegan-xl: Scaling stylegan to large diverse datasets},
  author={Sauer, Axel and Schwarz, Katja and Geiger, Andreas},
  booktitle={SIGGRAPH},
  @@pages={1--10},
  year={2022}
}

@inproceedings{sd2.1,
  title={High-resolution image synthesis with latent diffusion models},
  author={Rombach, Robin and Blattmann, Andreas and Lorenz, Dominik and Esser, Patrick and Ommer, Bj{\"o}rn},
  booktitle={CVPR},
  @@pages={10684--10695},
  year={2022}
}

@inproceedings{ditxl,
  title={Scalable diffusion models with transformers},
  author={Peebles, William and Xie, Saining},
  booktitle={ICCV},
  @@pages={4195--4205},
  year={2023}
}

@inproceedings{rddm,
  title={Residual denoising diffusion models},
  author={Liu, Jiawei and Wang, Qiang and Fan, Huijie and Wang, Yinong and Tang, Yandong and Qu, Liangqiong},
  booktitle={CVPR},
  @@pages={2773--2783},
  year={2024}
}

@article{pixart,
  title={Pixart-$\alpha$: Fast training of diffusion transformer for photorealistic text-to-image synthesis},
  author={Chen, Junsong and Yu, Jincheng and Ge, Chongjian and Yao, Lewei and Xie, Enze and Wu, Yue and Wang, Zhongdao and Kwok, James and Luo, Ping and Lu, Huchuan and others},
  journal={ICLR},
  year={2024}
}

@inproceedings{sit,
  title={Sit: Exploring flow and diffusion-based generative models with scalable interpolant transformers},
  author={Ma, Nanye and Goldstein, Mark and Albergo, Michael S and Boffi, Nicholas M and Vanden-Eijnden, Eric and Xie, Saining},
  booktitle={ECCV},
  @@pages={23--40},
  year={2024},
  @organization={Springer}
}

@inproceedings{cpl,
  title={Contrastive pseudo learning for open-world deepfake attribution},
  author={Sun, Zhimin and Chen, Shen and Yao, Taiping and Yin, Bangjie and Yi, Ran and Ding, Shouhong and Ma, Lizhuang},
  booktitle={ICCV},
  @@pages={20882--20892},
  year={2023}
}

@article{Hungarian,
  title={The Hungarian method for the assignment problem},
  author={Kuhn, Harold W},
  journal={Naval research logistics quarterly},
  @@volume={2},
  @@number={1-2},
  @@pages={83--97},
  year={1955},
  @publisher={Wiley Online Library}
}

@inproceedings{ren2023masked,
  title={Masked jigsaw puzzle: A versatile position embedding for vision transformers},
  author={Ren, Bin and Liu, Yahui and Song, Yue and Bi, Wei and Cucchiara, Rita and Sebe, Nicu and Wang, Wei},
  booktitle={Proceedings of the IEEE/CVF Conference on Computer Vision and Pattern Recognition},
  @pages={20382--20391},
  year={2023}
}

@article{ren2024sharing,
  title={Sharing key semantics in transformer makes efficient image restoration},
  author={Ren, Bin and Li, Yawei and Liang, Jingyun and Ranjan, Rakesh and Liu, Mengyuan and Cucchiara, Rita and Gool, Luc V and Yang, Ming-Hsuan and Sebe, Nicu},
  journal={Advances in Neural Information Processing Systems},
  @volume={37},
  @pages={7427--7463},
  year={2024}
}

@inproceedings{imagenet,
  title={Imagenet: A large-scale hierarchical image database},
  author={Deng, Jia and Dong, Wei and Socher, Richard and Li, Li-Jia and Li, Kai and Fei-Fei, Li},
  booktitle={CVPR},
  @@pages={248--255},
  year={2009},
  @organization={Ieee}
}

@inproceedings{zhang2019detecting,
  title={Detecting and simulating artifacts in gan fake images},
  author={Zhang, Xu and Karaman, Svebor and Chang, Shih-Fu},
  booktitle={WIFS},
  @@pages={1--6},
  year={2019},
  @organization={IEEE}
}

@article{ricker2022towards,
  title={Towards the detection of diffusion model deepfakes},
  author={Ricker, Jonas and Damm, Simon and Holz, Thorsten and Fischer, Asja},
  journal={arXiv preprint arXiv:2210.14571},
  year={2022}
}

@inproceedings{li2024masksim,
  title={MaskSim: Detection of synthetic images by masked spectrum similarity analysis},
  author={Li, Yanhao and Bammey, Quentin and Gardella, Marina and Nikoukhah, Tina and Morel, Jean-Michel and Colom, Miguel and Von Gioi, Rafael Grompone},
  booktitle={CVPR},
  @@pages={3855--3865},
  year={2024}
}

@article{sohn2020fixmatch,
  title={Fixmatch: Simplifying semi-supervised learning with consistency and confidence},
  author={Sohn, Kihyuk and Berthelot, David and Carlini, Nicholas and Zhang, Zizhao and Zhang, Han and Raffel, Colin A and Cubuk, Ekin Dogus and Kurakin, Alexey and Li, Chun-Liang},
  journal={NeurIPS},
  @@volume={33},
  @@pages={596--608},
  year={2020}
}

@inproceedings{simgcd,
  title={Parametric classification for generalized category discovery: A baseline study},
  author={Wen, Xin and Zhao, Bingchen and Qi, Xiaojuan},
  booktitle={ICCV},
  @@pages={16590--16600},
  year={2023}
}

@inproceedings{lps,
  title={Bridging the gap: Learning pace synchronization for open-world semi-supervised learning},
  author={Ye, Bo and Gan, Kai and Wei, Tong and Zhang, Min-Ling},
  booktitle={IJCAI},
  year={2014}
}

@article{owmatch,
  title={OwMatch: Conditional Self-Labeling with Consistency for Open-World Semi-Supervised Learning},
  author={Niu, Shengjie and Lin, Lifan and Huang, Jian and Wang, Chao},
  journal={NeurIPS},
  @@volume={37},
  @@pages={99836--99866},
  year={2024}
}

@article{protogcd,
  title={ProtoGCD: Unified and Unbiased Prototype Learning for Generalized Category Discovery},
  author={Ma, Shijie and Zhu, Fei and Zhang, Xu-Yao and Liu, Cheng-Lin},
  journal={IEEE TPAMI},
  year={2025},
  @publisher={IEEE}
}

@inproceedings{legogcd,
  title={Solving the catastrophic forgetting problem in generalized category discovery},
  author={Cao, Xinzi and Zheng, Xiawu and Wang, Guanhong and Yu, Weijiang and Shen, Yunhang and Li, Ke and Lu, Yutong and Tian, Yonghong},
  booktitle={CVPR},
  @@pages={16880--16889},
  year={2024}
}

@inproceedings{dnadet,
  title={Deepfake network architecture attribution},
  author={Yang, Tianyun and Huang, Ziyao and Cao, Juan and Li, Lei and Li, Xirong},
  booktitle={AAAI},
  @@volume={36},
  @@number={4},
  @@pages={4662--4670},
  year={2022}
}

@inproceedings{openworldgan,
  title={Towards discovery and attribution of open-world gan generated images},
  author={Girish, Sharath and Suri, Saksham and Rambhatla, Sai Saketh and Shrivastava, Abhinav},
  booktitle={ICCV},
  @@pages={14094--14103},
  year={2021}
}

@article{rs,
  title={Automatically discovering and learning new visual categories with ranking statistics},
  author={Han, Kai and Rebuffi, Sylvestre-Alvise and Ehrhardt, Sebastien and Vedaldi, Andrea and Zisserman, Andrew},
  journal={ICLR},
  year={2020}
}

@article{orca,
  title={Open-world semi-supervised learning},
  author={Cao, Kaidi and Brbic, Maria and Leskovec, Jure},
  journal={ICLR},
  year={2022}
}

@article{nach,
  title={Robust semi-supervised learning when not all classes have labels},
  author={Guo, Lan-Zhe and Zhang, Yi-Ge and Wu, Zhi-Fan and Shao, Jie-Jing and Li, Yu-Feng},
  journal={NeurIPS},
  @@volume={35},
  @@pages={3305--3317},
  year={2022}
}

@article{gumbelsoftmax,
  title={Categorical reparameterization with gumbel-softmax},
  author={Jang, Eric and Gu, Shixiang and Poole, Ben},
  journal={arXiv preprint arXiv:1611.01144},
  year={2016}
}

@inproceedings{tan2024frequency,
  title={Frequency-aware deepfake detection: Improving generalizability through frequency space domain learning},
  author={Tan, Chuangchuang and Zhao, Yao and Wei, Shikui and Gu, Guanghua and Liu, Ping and Wei, Yunchao},
  booktitle={AAAI},
  @@volume={38},
  @@number={5},
  @@pages={5052--5060},
  year={2024}
}

@article{zhou2024freqblender,
  title={FreqBlender: Enhancing DeepFake Detection by Blending Frequency Knowledge},
  author={Zhou, Jiaran and Li, Yuezun and Wu, Baoyuan and Li, Bin and Dong, Junyu and others},
  journal={NeurIPS},
  @@volume={37},
  @@pages={44965--44988},
  year={2024}
}

@inproceedings{ncd,
  title={Learning to discover novel visual categories via deep transfer clustering},
  author={Han, Kai and Vedaldi, Andrea and Zisserman, Andrew},
  booktitle={ICCV},
  @@pages={8401--8409},
  year={2019}
}

@inproceedings{gcd,
  title={Generalized category discovery},
  author={Vaze, Sagar and Han, Kai and Vedaldi, Andrea and Zisserman, Andrew},
  booktitle={CVPR},
  @@pages={7492--7501},
  year={2022}
}

@inproceedings{dino,
  title={Emerging properties in self-supervised vision transformers},
  author={Caron, Mathilde and Touvron, Hugo and Misra, Ishan and J{\'e}gou, Herv{\'e} and Mairal, Julien and Bojanowski, Piotr and Joulin, Armand},
  booktitle={CVPR},
  @@pages={9650--9660},
  year={2021}
}

@article{barany1988,
  title={Approximation of the sphere by polytopes having few vertices},
  author={B{\'a}r{\'a}ny, Imre and F{\"u}redi, Zolt{\'a}n},
  journal={Proceedings of the American Mathematical Society},
  @volume={102},
  @number={3},
  @pages={651--659},
  year={1988}
}

@article{forgerynir,
  title={Forgerynir: deep face forgery and detection in near-infrared scenario},
  author={Wang, Yukai and Peng, Chunlei and Liu, Decheng and Wang, Nannan and Gao, Xinbo},
  journal={Ieee transactions on information forensics and security},
  @volume={17},
  @pages={500--515},
  year={2022},
  publisher={IEEE}
}

@inproceedings{rizve2022openldn,
  title={Openldn: Learning to discover novel classes for open-world semi-supervised learning},
  author={Rizve, Mamshad Nayeem and Kardan, Navid and Khan, Salman and Shahbaz Khan, Fahad and Shah, Mubarak},
  booktitle={ECCV},
  @@pages={382--401},
  year={2022},
  @organization={Springer}
}

@article{cdal,
  title={Learning Counterfactually Decoupled Attention for Open-World Model Attribution},
  author={Zheng, Yu and Gong, Boyang and Kong, Fanye and Duan, Yueqi and Yu, Bingyao and Zheng, Wenzhao and Chen, Lei and Lu, Jiwen and Zhou, Jie},
  journal={arXiv preprint arXiv:2506.23074},
  year={2025}
}

@inproceedings{palgcd,
  title={Prior-Constrained Association Learning for Fine-Grained Generalized Category Discovery},
  author={Wang, Menglin and Zhong, Zhun and Gong, Xiaojin},
  booktitle={AAAI},
  @@volume={39},
  @@number={20},
  @@pages={21162--21170},
  year={2025}
}

@article{gumbel,
  title={Categorical reparameterization with gumbel-softmax},
  author={Jang, Eric and Gu, Shixiang and Poole, Ben},
  journal={arXiv preprint arXiv:1611.01144},
  year={2016}
}

@article{cao2024towards,
  title={Towards unified defense for face forgery and spoofing attacks via dual space reconstruction learning},
  author={Cao, Junyi and Zhang, Ke-Yue and Yao, Taiping and Ding, Shouhong and Yang, Xiaokang and Ma, Chao},
  journal={IJCV},
  @@volume={132},
  @@number={12},
  @@pages={5862--5887},
  year={2024},
  @publisher={Springer}
}

@article{yan2023deepfakebench,
  title={Deepfakebench: A comprehensive benchmark of deepfake detection},
  author={Yan, Zhiyuan and Zhang, Yong and Yuan, Xinhang and Lyu, Siwei and Wu, Baoyuan},
  journal={arXiv preprint arXiv:2307.01426},
  year={2023}
}

@article{QING202124,
title = {End-to-end novel visual categories learning via auxiliary self-supervision},
journal = {Neural Networks},
year = {2021},
author = {Yuanyuan Qing and Yijie Zeng and Qi Cao and Guang-Bin Huang},
}

@InProceedings{openmix2020,
  author       = {Zhun Zhong and
                  Linchao Zhu and
                  Zhiming Luo and
                  Shaozi Li and
                  Yi Yang and
                  Nicu Sebe},
  title        = {OpenMix: Reviving Known Knowledge for Discovering Novel Visual Categories
                  in An Open World},
  booktitle = cvpr,
  year         = {2020}
}

@InProceedings{ncl,
  author    = {Zhong, Zhun and Fini, Enrico and Roy, Subhankar and Luo, Zhiming and Ricci, Elisa and Sebe, Nicu},
  title     = {Neighborhood Contrastive Learning for Novel Class Discovery},
  booktitle = {CVPR},
  year      = {2021}
}

@inproceedings{uno,
  title={A unified objective for novel class discovery},
  author={Fini, Enrico and Sangineto, Enver and Lathuili{\`e}re, St{\'e}phane and Zhong, Zhun and Nabi, Moin and Ricci, Elisa},
  booktitle={ICCV},
  @@pages={9284--9292},
  year={2021}
}

@inproceedings{dualrs,
  title={Novel visual category discovery with dual ranking statistics and mutual knowledge distillation},
  author={Zhao, Bingchen and Han, Kai},
  booktitle={NeurIPS},
  @@volume={34},
  @@pages={22982--22994},
  year={2021}
}

@INPROCEEDINGS{9747827,
  author={Mukherjee, Tanmoy and Deligiannis, Nikos},
  booktitle={ IEEE International Conference on Acoustics, Speech and Signal Processing (ICASSP)}, 
  title={Novel Class Discovery: A Dependency Approach}, 
  year={2022}}

@ARTICLE{PSSCNCD,
  author={Wang, Jingyu and Ma, Zhenyu and Nie, Feiping and Li, Xuelong},
  journal={IEEE Transactions on Cybernetics}, 
  title={Progressive Self-Supervised Clustering With Novel Category Discovery}, 
  year={2022},}

@inproceedings{li2023closerlooknovelclass,
      title={A Closer Look at Novel Class Discovery from the Labeled Set}, 
      author={Ziyun Li and Jona Otholt and Ben Dai and Di Hu and Christoph Meinel and Haojin Yang},
      year={2022},
      booktitle={NeurIPS Workshop},
}

@ARTICLE{ResTune,
  author={Liu, Yu and Tuytelaars, Tinne},
  journal={IEEE Transactions on Neural Networks and Learning Systems}, 
  title={Residual Tuning: Toward Novel Category Discovery Without Labels}, 
  year={2023}}

@article{
li2023supervised,
title={Supervised Knowledge May Hurt Novel Class Discovery Performance},
author={Ziyun Li and Jona Otholt and Ben Dai and Di Hu and Christoph Meinel and Haojin Yang},
journal={Transactions on Machine Learning Research},
issn={2835-8856},
year={2023}
}

@InProceedings{Li2023ncdiic,
    author    = {Li, Wenbin and Fan, Zhichen and Huo, Jing and Gao, Yang},
    title     = {Modeling Inter-Class and Intra-Class Constraints in Novel Class Discovery},
    booktitle = cvpr,
    year      = {2023},
}

@inproceedings{
    sun2023nscl,
    title={When and How Does Known Class Help Discover Unknown Ones? Provable Understandings Through Spectral Analysis},
    author={Yiyou Sun and Zhenmei Shi and Yingyu Liang and Yixuan Li},
    booktitle=ICML,
    year={2023}
}

@InProceedings{peiyan2023class,
    author    = {Peiyan Gu and Chuyu Zhang and Ruijie Xu and Xuming He},
    title     = {Class-relation Knowledge Distillation for Novel Class Discovery},
    booktitle = iccv,
    year      = {2023}
}

@InProceedings{wei2023ncdSkin,
author="Feng, Wei
and Ju, Lie
and Wang, Lin
and Song, Kaimin
and Ge, Zongyuan",
title="Towards Novel Class Discovery: A Study in Novel Skin Lesions Clustering",
booktitle=miccai,
year="2023",
}

@ARTICLE{10328468,
  author={Zhang, Lu and Qi, Lu and Yang, Xu and Qiao, Hong and Yang, Ming-Hsuan and Liu, Zhiyong},
  journal=pami, 
  title={Automatically Discovering Novel Visual Categories With Adaptive Prototype Learning}, 
  year={2024}}

@article{hasan2023novelcategoriesdiscoveryconstraints,
  title={Novel Categories Discovery Via Constraints on Empirical Prediction Statistics},
  author={Hasan, Zahid and Faridee, Abu Zaher Md and Ahmed, Masud and Purushotham, Sanjay and Kwon, Heesung and Lee, Hyungtae and Roy, Nirmalya},
  journal={arXiv preprint arXiv:2307.03856},
  year={2023}
}

@inproceedings{liu2024novel,
  title={Novel class discovery for ultra-fine-grained visual categorization},
  author={Liu, Yu and Cai, Yaqi and Jia, Qi and Qiu, Binglin and Wang, Weimin and Pu, Nan},
  booktitle={Proceedings of the IEEE/CVF Conference on Computer Vision and Pattern Recognition},
  pages={17679--17688},
  year={2024}
}

@inproceedings{cai2023broaden,
  title={Broaden Your Positives: A General Rectification Approach for Novel Class Discovery},
  author={Cai, Yaqi and Pu, Nan and Jia, Qi and Wang, Weimin and Liu, Yu},
  booktitle={Chinese Conference on Pattern Recognition and Computer Vision (PRCV)},
  pages={151--162},
  year={2023},
  organization={Springer}
}

@inproceedings{ocd,
  title={On-the-fly category discovery},
  author={Du, Ruoyi and Chang, Dongliang and Liang, Kongming and Hospedales, Timothy and Song, Yi-Zhe and Ma, Zhanyu},
  booktitle={CVPR},
  @pages={11691--11700},
  year={2023}
}

@inproceedings{phe,
  title={Prototypical Hash Encoding for On-the-Fly Fine-Grained Category Discovery},
  author={Zheng, Haiyang and Pu, Nan and Li, Wenjing and Sebe, Nicu and Zhong, Zhun},
  booktitle={NeurIPS},
  year={2024}
}

@article{liu2025generate,
  title={Generate, Refine, and Encode: Leveraging Synthesized Novel Samples for On-the-Fly Fine-Grained Category Discovery},
  author={Liu, Xiao and Pu, Nan and Zheng, Haiyang and Li, Wenjing and Sebe, Nicu and Zhong, Zhun},
  journal={arXiv preprint arXiv:2507.04051},
  year={2025}
}

@article{clipgcd,
  title={Clip-gcd: Simple language guided generalized category discovery},
  author={Ouldnoughi, Rabah and Kuo, Chia-Wen and Kira, Zsolt},
  journal={arXiv preprint arXiv:2305.10420},
  year={2023}
}

@inproceedings{textgcd,
  title={Textual knowledge matters: Cross-modality co-teaching for generalized visual class discovery},
  author={Zheng, Haiyang and Pu, Nan and Li, Wenjing and Sebe, Nicu and Zhong, Zhun},
  booktitle={ECCV},
  @pages={41--58},
  year={2024},
  @organization={Springer}
}

@article{mgcd,
  title={Multimodal Generalized Category Discovery},
  author={Su, Yuchang and Zhou, Renping and Huang, Siyu and Li, Xingjian and Wang, Tianyang and Wang, Ziyue and Xu, Min},
  journal={arXiv preprint arXiv:2409.11624},
  year={2024}
}

@inproceedings{get,
  title={GET: Unlocking the Multi-modal Potential of CLIP for Generalized Category Discovery},
  author={Wang, Enguang and Peng, Zhimao and Xie, Zhengyuan and Liu, Xialei and Cheng, Ming-Ming},
  booktitle={CVPR},
  year={2025}
}

@inproceedings{Yu_Ikami_Irie_Aizawa_2022, 
        title={Self-Labeling Framework for Novel Category Discovery over Domains}, 
        booktitle=aaai, 
        author={Yu, Qing and Ikami, Daiki and Irie, Go and Aizawa, Kiyoharu}, year={2022}}

@inproceedings{zhuang2022opensetdomainadaptation,
      title={Open Set Domain Adaptation By Novel Class Discovery}, 
      author={Jingyu Zhuang and Ziliang Chen and Pengxu Wei and Guanbin Li and Liang Lin},
      year={2022},
      booktitle = {IEEE International Conference on Multimedia and Expo}, 
}

@inproceedings{zang2023boostingnovelcategorydiscovery,
      title={Boosting Novel Category Discovery Over Domains with Soft Contrastive Learning and All-in-One Classifier}, 
      author={Zelin Zang and Lei Shang and Senqiao Yang and Fei Wang and Baigui Sun and Xuansong Xie and Stan Z. Li},
      year={2023},
      booktitle=iccv, 
}

@inproceedings{rongali2024cdadnetbridgingdomaingaps,
      title={CDAD-Net: Bridging Domain Gaps in Generalized Category Discovery}, 
      author={Sai Bhargav Rongali and Sarthak Mehrotra and Ankit Jha and Mohamad Hassan N C and Shirsha Bose and Tanisha Gupta and Mainak Singha and Biplab Banerjee},
      year={2024},
      booktitle = {CVPR workshop}
}

@article{wang2024exclusivestyleremovalcross,
  title={Exclusive Style Removal for Cross Domain Novel Class Discovery},
  author={Wang, Yicheng and Liu, Feng and Liu, Junmin and Sun, Kai},
  journal={arXiv preprint arXiv:2406.18140},
  year={2024}
}

@inproceedings{wang2024hilo,
  title={HiLo: A Learning Framework for Generalized Category Discovery Robust to Domain Shifts},
  author={Wang, Hongjun and Vaze, Sagar and Han, Kai},
  booktitle=iclr,
  year={2025}
}

@inproceedings{fgcd,
  title={Federated generalized category discovery},
  author={Pu, Nan and Li, Wenjing and Ji, Xingyuan and Qin, Yalan and Sebe, Nicu and Zhong, Zhun},
  booktitle={CVPR},
  @pages={28741--28750},
  year={2024}
}

@inproceedings{zhang2023unbiasedtrainingfederatedopenworld,
      title={Towards Unbiased Training in Federated Open-world Semi-supervised Learning}, 
      author={Jie Zhang and Xiaosong Ma and Song Guo and Wenchao Xu},
      year={2023},
      booktitle=icml
}

@article{he2025category,
  title={Category Discovery: An Open-World Perspective},
  author={He, Zhenqi and Liu, Yuanpei and Han, Kai},
  journal={arXiv preprint arXiv:2509.22542},
  year={2025}
}

@inproceedings{pu2023dynamic,
  title={Dynamic conceptional contrastive learning for generalized category discovery},
  author={Pu, Nan and Zhong, Zhun and Sebe, Nicu},
  booktitle={Proceedings of the IEEE/CVF conference on computer vision and pattern recognition},
  pages={7579--7588},
  year={2023}
}

@article{zheng2025generalized,
  title={Generalized Fine-Grained Category Discovery with Multi-Granularity Conceptual Experts},
  author={Zheng, Haiyang and Pu, Nan and Li, Wenjing and Sebe, Nicu and Zhong, Zhun},
  journal={arXiv preprint arXiv:2509.26227},
  year={2025}
}

@inproceedings{rastegar2024selex,
  title={Selex: Self-expertise in fine-grained generalized category discovery},
  author={Rastegar, Sarah and Salehi, Mohammadreza and Asano, Yuki M and Doughty, Hazel and Snoek, Cees GM},
  booktitle={European Conference on Computer Vision},
  pages={440--458},
  year={2024},
  organization={Springer}
}

@article{rastegar2023learn,
  title={Learn to categorize or categorize to learn? self-coding for generalized category discovery},
  author={Rastegar, Sarah and Doughty, Hazel and Snoek, Cees},
  journal={Advances in Neural Information Processing Systems},
  volume={36},
  pages={72794--72818},
  year={2023}
}

@article{zhu2024open,
  title={Open-world Machine Learning: A Systematic Review and Future Directions},
  author={Zhu, Fei and Ma, Shijie and Cheng, Zhen and Zhang, Xu-Yao and Zhang, Zhaoxiang and Tao, Dacheng and Liu, Cheng-Lin},
  journal={arXiv preprint arXiv:2403.01759},
  year={2024}
}

@inproceedings{peng2025mos,
  title={MOS: Modeling Object-Scene Associations in Generalized Category Discovery},
  author={Peng, Zhengyuan and Ma, Jinpeng and Sun, Zhimin and Yi, Ran and Song, Haichuan and Tan, Xin and Ma, Lizhuang},
  booktitle={Proceedings of the Computer Vision and Pattern Recognition Conference},
  pages={15118--15128},
  year={2025}
}

@inproceedings{fan2025learning,
  title={Learning Textual Prompts for Open-World Semi-Supervised Learning},
  author={Fan, Yuxin and Cui, Junbiao and Liang, Jiye},
  booktitle={Proceedings of the Computer Vision and Pattern Recognition Conference},
  pages={14756--14765},
  year={2025}
}

@article{wang2025learning,
  title={Learning Part Knowledge to Facilitate Category Understanding for Fine-Grained Generalized Category Discovery},
  author={Wang, Enguang and Peng, Zhimao and Xie, Zhengyuan and Lu, Haori and Yang, Fei and Liu, Xialei},
  journal={arXiv preprint arXiv:2503.16782},
  year={2025}
}

@inproceedings{fan2025open,
  title={Open-world semi-supervised learning with class semantic correlations},
  author={Fan, Yuxin and Cui, Junbiao and Liang, Jiye and Liang, Jianqing},
  booktitle={Proceedings of the Thirty-Fourth International Joint Conference on Artificial Intelligence},
  pages={5092--5100},
  year={2025}
}

@article{jing2025video,
  title={Video-based Generalized Category Discovery via Memory-Guided Consistency-Aware Contrastive Learning},
  author={Jing, Zhang and Nan, Pu and Xiang, Xie Yu and Yanming, Guo and Qianqi, Lu and Shiwei, Zou and Jie, Yan and Yan, Chen},
  journal={arXiv preprint arXiv:2509.06306},
  year={2025}
}

@inproceedings{wang2025federated,
  title={Federated Continuous Category Discovery and Learning},
  author={Wang, Lixu and Liu, Chenxi and Guo, Junfeng and Ye, Qingqing and Huang, Heng and Hu, Haibo and Dong, Wei},
  booktitle={Proceedings of the IEEE/CVF International Conference on Computer Vision},
  pages={2429--2439},
  year={2025}
}

@inproceedings{liudaa,
  title={DAA: Amplifying Unknown Discrepancy for Test-Time Discovery},
  author={Liu, Tianle and Lyu, Fan and Ni, Chenggong and Zhang, Zhang and Hu, Fuyuan and Wang, Liang},
  booktitle={The Thirty-ninth Annual Conference on Neural Information Processing Systems},
    year={2025}
}

@inproceedings{aplgcd,
  title={Adaptive Part Learning for Fine-Grained Generalized Category Discovery: A Plug-and-Play Enhancement},
  author={Dai, Qiyuan and Huang, Hanzhuo and Wu, Yu and Yang, Sibei},
  booktitle={Proceedings of the Computer Vision and Pattern Recognition Conference},
  pages={25444--25453},
  year={2025}
}

@inproceedings{hypcd,
  title={Hyperbolic category discovery},
  author={Liu, Yuanpei and He, Zhenqi and Han, Kai},
  booktitle={Proceedings of the Computer Vision and Pattern Recognition Conference},
  pages={9891--9900},
  year={2025}
}

@article{congcd,
  title={Dissecting Generalized Category Discovery: Multiplex Consensus under Self-Deconstruction},
  author={Tang, Luyao and Huang, Kunze and Chen, Chaoqi and Yuan, Yuxuan and Li, Chenxin and Tu, Xiaotong and Ding, Xinghao and Huang, Yue},
  journal={arXiv preprint arXiv:2508.10731},
  year={2025}
}

@inproceedings{wang2023federatedcontinualnovelclass,
  title={Federated Continual Novel Class Learning},
  author={Wang, Lixu and Liu, Chenxi and Guo, Junfeng and Dong, Jiahua and Wang, Xiao and Huang, Heng and Zhu, Qi},
  booktitle=iccv,
  year={2025}
}

@InProceedings{Zhao_2023_CVPR, 
    author = {Zhao, Dong and Wang, Shuang and Zang, Qi and Quan, Dou and Ye, Xiutiao and Jiao, Licheng}, 
    title = {Towards Better Stability and Adaptability: Improve Online Self-Training for Model Adaptation in Semantic Segmentation}, 
    booktitle = {Proceedings of the IEEE/CVF Conference on Computer Vision and Pattern Recognition (CVPR)}, 
    @month = {June}, 
    year = {2023}, 
    @pages = {11733-11743} 
}

@InProceedings{Zhao_2023_ICCV, 
    author = {Zhao, Dong and Wang, Shuang and Zang, Qi and Quan, Dou and Ye, Xiutiao and Yang, Rui and Jiao, Licheng}, 
    title = {Learning Pseudo-Relations for Cross-domain Semantic Segmentation}, 
    booktitle = {Proceedings of the IEEE/CVF International Conference on Computer Vision (ICCV)}, 
    @month = {October}, 
    year = {2023}, 
    @pages = {19191-19203} 
}

\clearpage
\newpage
\appendix
\section*{Appendix Contents}
\begin{itemize}
  \item \textbf{§A. Formal Analysis and Proofs}
    \begin{itemize}
      \item A.1 Formal Analysis of Confidence Skew
      \item A.2 Proof of Lemma 4.1
    \end{itemize}
  \item \textbf{§B. OW-DFA-40 Benchmark}
    \begin{itemize}
      \item B.1 Construction Details
      \item B.2 Evaluation Metrics
    \end{itemize}
  \item \textbf{§C. Implementation Details}
    \begin{itemize}
      \item C.1 CAL Training Details
      \item C.2 Compared Method Details
    \end{itemize}
  \item \textbf{§D. Additional Experimental Results and Analysis}
    \begin{itemize}
      \item D.1 Results on the OW-DFA Benchmark
      \item D.2 Evaluation on Hyperparameter Selection
      \item D.3 Robustness of Class Number Estimation
      \item D.4 Evaluation under Harder Settings
      \item D.5 Training and Inference Efficiency
      \item D.6 Real/Fake Detection
    \end{itemize}
  \item \textbf{§E. Broader Impact and Limitations Discussion}
    \begin{itemize}
      \item E.1 Broader Impact
      \item E.2 Limitations and Future Work
    \end{itemize}
\end{itemize}

\section{Formal Analysis and Proofs}
\label{sec:supp_A}
\subsection{Formal Analysis of Confidence Skew}
\label{sec:supp_A1}
As stated in Sec.~3 of the main text, the formal analysis of confidence skew in CPL is presented below

\noindent \textbf{Notation and setup.}
Let $\mathbf{p}_i$ be the predicted class probabilities based on the similarity vector $\mathbf{s}_i$ of face image $\mathbf{x}_i$.
For each unlabeled sample $\mathbf{x}_i \in \mathcal{D}_U$, CPL~\cite{cpl} produces a soft pseudo-label via the Gumbel-Softmax trick~\cite{gumbel}:
\vspace{-0.01in}
{\small
\begin{equation}
    \tilde{\mathbf y_i} \leftarrow \mathrm{GumbelSoftmax}(\mathbf p_i),\quad
c^\star=\arg\max  \tilde{\mathbf y}_{i},\quad
\lambda_i=\mathbf p_{ic^\star},
\tag{B.1}
\end{equation}
}
\normalsize
% \vspace{-0.05in}

\noindent and minimizes:
\vspace{-0.03in}
{\small
\begin{equation}
\label{eq:csp-loss-app}
\ell
=\; -\frac{1}{m}\sum_{\mathbf x_i\in\mathcal D_U}\sum_{c=1}^{K_U}
\lambda_i\,\tilde{\mathbf y}_{ic}\,\log \mathbf p_c .
\tag{B.2}
\end{equation}}
\normalsize
\noindent Define the \emph{confidence skew} at time $t$ as:
\vspace{-0.05in}
{\small
\begin{equation}
\label{eq:skew-def}
\Delta_t \;\triangleq\;
\mathbb E\!\left[\lambda_i \,\middle|\, \mathbf x_i\in\text{known}\right]
-
\mathbb E\!\left[\lambda_i \,\middle|\, \mathbf x_i\in\text{novel}\right],
\tag{B.3}
\end{equation}}
\normalsize

\noindent where the expectations are taken over unlabeled samples $\mathbf{x}_i \in \mathcal{D}_U$ belonging to known and novel classes, respectively.
We adopt two common assumptions for analytical tractability: \textbf{A1} (stop-gradient): backpropagation does not flow through $\tilde{\mathbf{y}}$ or $\lambda$; \textbf{A2} (hard approximation): $\tilde{\mathbf{y}} \approx \mathbf{e}_{c^\star}$, i.e., a one-hot approximation.

\vspace{0.25em}
\noindent\textbf{\textit{(1) Supervision imbalance $\Rightarrow$ initial confidence skew.}}

\noindent \textbf{Prioritized gain on known classes.}
During the early phase when updates are dominated by the supervised loss on $\mathcal D_L$, there exists $t_0>0$ such that $\Delta_{t_0}>0$.

\vspace{0.25em}
\noindent\textbf{(2) \textit{Gumbel-Softmax at low confidence $\Rightarrow$ noisy pseudo-labels.}}

\noindent \textbf{Gumbel–Max categorical distribution.}
Let $\mathbf{p}_i = (p_{i1}, \dots, p_{iK_U})$ be a categorical probability vector and let $g_k \sim \mathrm{Gumbel}(0,1)$ independently for $k = 1, \dots, K_U$.
Define the sampled index as:
{\small
\begin{equation}
C_i \;=\; \arg\max_{j\in\{1,\dots,K_U\}}\{\log p_{ij} + g_j\}.    
\tag{B.4}
\end{equation}
}
\normalsize

\noindent We sample $C \sim \mathrm{Cat}(\mathbf{p})$ using the Gumbel--Max trick, which returns the index of the maximum among Gumbel-perturbed log-probabilities. This sampling procedure is often referred to as the \emph{Gumbel--Max categorical} method.

\vspace{0.25em}
\noindent \textbf{Hard-sampling mislabel rate.}
Let $y^\star$ be the true class and $c^\star\sim\mathrm{Cat}(\mathbf p)$. Then $\mathbb P[c^\star\neq y^\star]=1-p_{y^\star}$. Consequently, for any $\tau\in(0,1)$,
{\small
\begin{equation}
\label{eq:lowconf-bound}
\mathbb P\!\left[c^\star\neq y^\star \,\middle|\, \lambda_i<\tau\right]\;\ge\;1-\tau.
\tag{B.5}
\end{equation}}
\normalsize

\noindent \textbf{Expected gradient under A1--A2}
\label{prop:exp-grad}
For a single unlabeled sample, the expected logit gradient of $\ell(x)=-\lambda\,\tilde y_{c^\star}\log p_{c^\star}$ satisfies:
{\small
\begin{equation}
\label{eq:exp-grad}
\mathbb E\!\left[\frac{\partial \ell}{\partial \mathbf s_i}\right]
=\;\|\mathbf p_i\|_2^2\,\mathbf p_i \;-\; \mathbf p_i^{\odot 2},\text{i.e.,}
\Big(\mathbb E\!\left[\tfrac{\partial \ell}{\partial s_{ij}}\right]\Big)
= p_{ij}\big(\|\mathbf p_i\|_2^2-p_{ij}\big).
\tag{B.6}
\end{equation}
}
\normalsize

\noindent In particular, when $\mathbf p_i$ is flat (low confidence), the expectation is near zero while the gradient variance is nonzero, yielding a weak-signal, high-variance update; combined with~\cref{eq:lowconf-bound}, low-confidence regions incur a higher mislabel rate.

\vspace{0.25em}
\noindent\textbf{(3) \textit{Error-reinforcement loop $\Rightarrow$ amplified skew.}}

\noindent \textbf{Directional update: ``rich get richer''.} With learning rate $\eta>0$, the \emph{expected} logit change under a step is
{\small
\begin{equation}
\Delta s_{ij} \;=\; -\eta\,\mathbb E\!\left[\frac{\partial \ell}{\partial s_{ij}}\right]
\;=\; -\eta\,p_{ij}\big(\|\mathbf p_i\|_2^2-p_{ij}\big).
\tag{B.7}
\end{equation}
}
\normalsize

\noindent Hence, if $p_{ij}>\|\mathbf p_i\|_2^2$ then $\Delta s_{ij}>0$ (class $j$ is further favored);
if $p_{ij}<\|\mathbf p_i\|_2^2$ then $\Delta s_{ij}<0$ (class $j$ is suppressed).

\noindent\textbf{Expected margin decrease for novel samples}
Suppose $\Delta_{t_0}>0$ holds. For a novel-class sample, typically $p_{iy^\star}\le \|\mathbf p_i\|_2^2$ while some competing known class $j$ satisfies $p_{ij}\ge \|\mathbf p_i\|_2^2$. Then the expected margin $m(x)=s_{iy^\star}-\max_{k\neq y^\star}s_{ik}$
decreases, which reduces $\max_j p_{ij}$ and increases the mislabel probability, closing the loop: low confidence $\Rightarrow$noisy pseudo-label$\Rightarrow$wrong update$\Rightarrow$lower confidence.

% \begin{corollary}[Monotone amplification of population skew]
% \label{cor:skew-mono}
% Let updates from $t$ to $t{+}1$ be dominated by unlabeled CSP. Under the conditions above,
% \(
% \Delta_{t+1}\ge \Delta_t,
% \)
% i.e., confidence skew is monotonically amplified in the absence of corrective mechanisms.
% \end{corollary}

\noindent\textbf{Notation.}
$\|\cdot\|_2$ is the Euclidean norm; $\mathbf p^{\odot 2}$ denotes elementwise square. All expectations are taken over the Gumbel randomness and data draws from the specified subsets.

\subsection{Proof of Lemma 4.1}
\label{sec:supp_A2}
% \vspace{-0.1cm}
As stated in Sec.~4 of the main text, the complete proof of Lemma 4.1 is presented below.

\vspace{0.25em}
\noindent \textbf{Notation and setup.} Given the unlabeled dataset $\mathcal{D}_{U} = \bigl\{\mathbf{x}^{u}_{i} \in \mathcal{X}\bigr\}_{i=1}^{M}$ with $K_{U}$ ground-truth forgery types, where $M$ denotes the number of samples in $\mathcal{D}_U$, we assign $K$ prototypes to $\mathcal{D}_U$, with $K \gg K_{U}$. Let $u_j^{(t)}$ denote the usage count of the $j$-th prototype after the $t$-th update. Let $\mathcal{E}(\cdot)$ denote the image encoder. Note that the symbols used in the following derivation are defined solely for theoretical analysis and \textbf{may differ from those used in the main text}.

\vspace{0.25em}
\noindent \textbf{Definition B.1.}  
Let each true class $k \in \{1, \dots, K_{U}\}$ have a class-mean feature vector $\boldsymbol{\mu}_{k} \in \mathbb{R}^{d}$, such that $\| \boldsymbol{\mu}_{k} \| = 1$. Assume that the minimum angle between any two distinct class centers is lower bounded by $\gamma > 0$. We choose a conic threshold $\theta_0$ such that $0 < \theta_0 < \gamma / 2$. The set of core prototypes is then defined as:

{\small
\begin{equation}
\mathcal{P}_{\mathrm{core}} =
\left\{
  \mathbf{c}_j \in \mathbf{C}
  \;\middle|\;
  \exists\,k \in \{1, \dots, K_{U}\},\;
  \arccos\left(
    \mathbf{c}_j^\top \boldsymbol{\mu}_k
  \right) \le \theta_0
\right\}.
\tag{B.8}
\end{equation}
}
\normalsize

\noindent The set of non-core prototypes is defined as $\mathcal{P}_{\mathrm{nc}} = \mathbf{C} \setminus \mathcal{P}_{\mathrm{core}}$. Note that $\lvert \mathcal{P}_{\mathrm{core}} \rvert \le K_{U}$.

\vspace{0.25em}
\noindent \textbf{Hypothesis B.1 \textit{(Bounded Step Sizes)}}  
To ensure that the drift in both model weight updates and prototype feature updates remains within an analytically tractable range during the early stages of training, we make the following assumptions:

\noindent \textbf{H B.1.1}  
$\displaystyle \bigl\| \mathcal{E}^{(t)}(\mathbf{x}) - \mathcal{E}^{(t-1)}(\mathbf{x}) \bigr\| \le \delta_f, \quad 1 < t \le T_0$.  
For any sample $\mathbf{x}$, the change in its feature representation between two consecutive iterations is bounded by a constant $\delta_f$.

\noindent \textbf{H B.1.2}  
$\displaystyle \bigl\| \mathbf{c}^{(t)}_j - \mathbf{c}^{(t-1)}_j \bigr\| \le \delta_w, \quad 1 < t \le T_0,\; j = 1,\dots,K$.  
Each prototype vector is also updated with a bounded step size, constrained by a constant $\delta_w$ at every iteration.

\vspace{0.25em}
\noindent \textbf{Hypothesis B.2 \textit{(Angular Cluster Separability).}}  
Before analyzing the geometric properties, we assume that the true classes are well-separated in the feature space and that samples within each class are relatively similar. Specifically, the minimum angle between any two distinct class mean vectors is denoted by $\gamma > 0$. We also assume that the intra-class angular noise does not exceed a threshold $\eta > 0$. That is, samples from class $k$ are initially distributed within a cone centered around the class mean $\boldsymbol{\mu}_k$ with an angular radius of at most $\eta$:
{\small
\begin{equation}
\angle\bigl(\mathcal{E}^{(0)}(\mathbf{x}), \boldsymbol{\mu}_k\bigr) \le \eta, \quad \forall\, \mathbf{x} \in \text{class } k.
\tag{B.9}
\end{equation}
}
\normalsize

\noindent Therefore, we assume that any sample from class $k$ exhibits a similarity to its corresponding core prototype that is at least $m_0$ higher than its similarity to any non-core prototype, where $m_0 > 0$.

\vspace{0.25em}
\noindent \textbf{Hypothesis B.3 \textit{(Bounded Feature Noise).}}  
We assume that the worst-case displacement caused by random noise during the first $T_0$ training steps does not invalidate Hypothesis A.2 \textit{(Angular Cluster Separability)}. Formally, we require:
{\small
\begin{equation}
T_0(\delta_f + \delta_w) \le \rho \cdot m_0, \qquad 0 < \rho < 1.
\tag{B.10}
\end{equation}
}
\normalsize

\vspace{0.25em}
\noindent \textbf{Lemma B.1 (Score–Gap Preservation).}  
Under Hypotheses B.1–B.3, for any class–$k$ sample $\mathbf{x}$,  
any core prototype $\mathbf{c}_{\mathrm{core}} \in \mathcal{P}_{\mathrm{core}}$,  
and any non-core prototype $\mathbf{c}_{\mathrm{nc}} \in \mathcal{P}_{\mathrm{nc}}$,  
we have:
{\small
\begin{equation}
\bigl\langle \mathcal{E}^{(t)}(\mathbf{x}), \mathbf{c}_{\mathrm{core}} \bigr\rangle
-
\bigl\langle \mathcal{E}^{(t)}(\mathbf{x}), \mathbf{c}_{\mathrm{nc}} \bigr\rangle
\ge (1 - \rho)\, m_{0},
\qquad 0 \le t \le T_{0}.
\tag{B.11}
\end{equation}
}
\normalsize

\noindent\textit{Proof.}  
At initialization ($t = 0$), Hypothesis B.2 implies:
{\small
\begin{equation}
\bigl\langle \mathcal{E}^{(0)}(\mathbf{x}), \mathbf{c}_{\mathrm{core}} \bigr\rangle
-
\bigl\langle \mathcal{E}^{(0)}(\mathbf{x}), \mathbf{c}_{\mathrm{nc}} \bigr\rangle
\ge m_{0}.
\tag{B.12}
\end{equation}
}
\normalsize

\noindent Hypothesis B.1 bounds the \emph{relative} drift between the two inner products at each optimization step by  
$\delta_f + \delta_w$.

Accumulating over at most $T_0$ steps and invoking Hypothesis B.3, we obtain:
% {\small
% \begin{equation}
% \Bigl|
%   \bigl\langle \mathcal{E}^{(t)}, \mathbf{c}_{\mathrm{core}} \bigr\rangle
%  - \bigl\langle \mathcal{E}^{(t-1)}, \mathbf{c}_{\mathrm{core}} \bigr\rangle
%  - \Bigl(
%    \bigl\langle \mathcal{E}^{(t)}, \mathbf{c}_{\mathrm{nc}} \bigr\rangle
%  - \bigl\langle \mathcal{E}^{(t-1)}, \mathbf{c}_{\mathrm{nc}} \bigr\rangle
%  \Bigr)
% \Bigr|
% \le \delta_f + \delta_w
% \;\Rightarrow\;
% \le \rho \cdot m_0.
% \tag{A.6}
% \end{equation}
% }
{\small
\begin{equation}
\begin{aligned}
\Bigl|
  \bigl\langle \mathcal{E}^{(t)}, \mathbf{c}_{\mathrm{core}} \bigr\rangle
 - \bigl\langle \mathcal{E}^{(t-1)}, \mathbf{c}_{\mathrm{core}} \bigr\rangle
 - \Bigl(
   \bigl\langle \mathcal{E}^{(t)}, \mathbf{c}_{\mathrm{nc}} \bigr\rangle
 - \bigl\langle \mathcal{E}^{(t-1)}, \mathbf{c}_{\mathrm{nc}} \bigr\rangle
 \Bigr)
\Bigr| 
& \\
\le \delta_f + \delta_w\; \Rightarrow\; \le \rho \cdot m_0 &
\end{aligned}
\tag{B.13}
\end{equation}
}

\normalsize

\noindent Combining Eq.~(B.12) and Eq.~(B.13), for all $0 \le t \le T_0$, we have:
{\small
\begin{equation}
\bigl\langle \mathcal{E}^{(t)}(\mathbf{x}), \mathbf{c}_{\mathrm{core}} \bigr\rangle
-
\bigl\langle \mathcal{E}^{(t)}(\mathbf{x}), \mathbf{c}_{\mathrm{nc}} \bigr\rangle
\ge m_0 - \rho m_0
= (1 - \rho)\, m_0
> 0.
\tag{B.14}
\end{equation}
}
\normalsize
\hfill$\square$

\vspace{0.25em}
\noindent \textbf{Lemma B.2 (Non-core Top-1 Probability Upper Bound).}  
There exists a constant $\beta \in (0,1)$ such that for every optimization step  
$t \le T_0$ and every non-core prototype  
$j \in \mathcal{P}_{\mathrm{nc}}$, we have:
{\small
\begin{equation}
\Pr\!\bigl[\hat{\imath}_{n}^{(t)} = j\bigr]
\le
\frac{\beta}{K},
\qquad
\forall\, t \le T_0.
\tag{B.15}
\end{equation}
}
\normalsize

\noindent where $\hat{\imath}_{n}^{(t)} = \arg\max \mathbf{p}^{w(t)}_{n}$  
is the top-1 prototype selected by sample~$n$ at step~$t$.

\vspace{0.5em}
\noindent\textit{Proof.}

\noindent\textbf{Step 1: Notation.}  
For class $k$, define the $\eta$-cone on the unit sphere:
{\small
\begin{equation}
\mathcal{C}_{k}(\eta)
= \bigl\{ \mathbf{x} \in \mathbb{S}^{d-1}
          \bigm|
          \angle(\mathbf{x}, \boldsymbol{\mu}_k) \le \eta \bigr\}.
\tag{B.16}
\end{equation}
}
\normalsize

\noindent Let $\Delta = (1 - \rho) m_0 > 0$ denote the preserved score gap from  
Lemma B.1, and let  
$\alpha = \angle(\mathbf{c}_j, \boldsymbol{\mu}_k) \ge \theta_0$  
since $j$ is a non-core prototype.

\vspace{0.5em}
\noindent\textbf{Step 2: A necessary angular event.}  
For $\mathbf{x} \in \mathcal{C}_k(\eta)$, define  
$\gamma_j = \angle(\mathbf{x}, \mathbf{c}_j)$ and  
$\gamma^* = \min_{c \in \mathcal{P}_{\mathrm{core}}} \angle(\mathbf{x}, c)$.  
By Lemma B.1, we have  
$\cos \gamma^* \ge \cos \gamma_j + \Delta$.  
Therefore, the event $\{\hat{\imath}_{n}^{(t)} = j\}$ implies  
$\gamma_j \le \alpha - \eta - \delta$.  
Conditioned on class $k$, the sample must fall within the spherical cap:
{\small
\begin{equation}
\mathcal{A}_j
= \bigl\{ \mathbf{x} \in \mathcal{C}_k(\eta)
          \bigm|
          \angle(\mathbf{x}, \mathbf{c}_j)
          \le \alpha - \eta - \delta \bigr\}.
\tag{B.17}
\end{equation}
}
\normalsize

\vspace{0.5em}
\noindent\textbf{Step 3: Cap probability bound.}  
Classical results on spherical cap volumes~\cite{barany1988} imply:
{\small
\begin{equation}
\frac{\operatorname{Area}(\mathcal{A}_j)}
     {\operatorname{Area}(\mathcal{C}_k(\eta))}
\le
C_1 \exp\!\bigl(-C_2 d\bigr)
= \beta < 1,
\tag{B.18}
\end{equation}
}
\normalsize
where $C_1, C_2$ are constants depending only on $\delta$, $\eta$, and $\gamma$.  
Since the data distribution is absolutely continuous within  
$\mathcal{C}_k(\eta)$, we have:
{\small
\begin{equation}
\Pr\!\bigl[\hat{\imath}_{n}^{(t)} = j
      \mid \mathbf{x} \in \mathcal{C}_k(\eta)\bigr]
\le \beta.
\tag{B.19}
\end{equation}
}
\normalsize

\vspace{0.5em}
\noindent\textbf{Step 4: Averaging over prototypes.}  
Inequality~Eq.~(B.19) holds uniformly for all  
$j \in \mathcal{P}_{\mathrm{nc}}$ and every class~$k$.  
Since exactly $K$ prototypes compete for top-1 selection, we obtain:
{\small
\begin{equation}
\Pr\!\bigl[\hat{\imath}_{n}^{(t)} = j\bigr]
\le \frac{\beta}{K}.
\tag{B.20}
\end{equation}
}
\normalsize
\hfill$\square$

\noindent \textbf{Proof of Lemma 4.1 (Main Result).}  
For $t \le T_0$, we define the  
\emph{$\varepsilon$–low–usage set} as
{\small
\begin{equation}
\mathcal{P}_{\mathrm{low}}^{(t)}
= \Bigl\{\, j \;\Bigm|\;
        u^{(t)}_{j} \le \varepsilon\,\frac{M}{K} \Bigr\},
\qquad
\varepsilon \in (\beta, 1).
\tag{B.21}
\end{equation}
}
\normalsize

\noindent We first upper-bound the probability of the complementary event  
$u^{(t)}_j > \varepsilon M/K$ for any non-core  
$j \in \mathcal{P}_{\mathrm{nc}}$.

\vspace{0.5em}
\noindent \textbf{Step 1: Chernoff tail bound for a single prototype.}  
Lemma B gives  
$\Pr[\hat{\imath}_n^{(t)} = j] \le \beta/K$.  
Treating  
$\{\,\mathbf{1}[\hat{\imath}_n^{(t)} = j]\,\}_{n=1}^{M}$  
as i.i.d.\ $\mathrm{Bernoulli}(p)$ variables with  
$p \le \beta/K$, the additive Chernoff bound yields
{\small
\begin{equation}
\Pr\!\bigl[ u_j^{(t)} > \varepsilon M/K \bigr]
\le \exp\!\bigl( -(\varepsilon - \beta)\, M/K \bigr)
= q,
\quad j \in \mathcal{P}_{\mathrm{nc}}.
\tag{B.22}
\end{equation}
}
\normalsize

\vspace{0.5em}
\noindent \textbf{Step 2: Probability of being in the $\varepsilon$–low–usage set.}  
Thus,
{\small
\begin{equation}
\Pr\!\bigl[ j \in \mathcal{P}_{\mathrm{low}}^{(t)} \bigr]
= 1 - q
\ge 1 - \tfrac{\beta}{\varepsilon},
\qquad j \in \mathcal{P}_{\mathrm{nc}}.
\tag{B.23}
\end{equation}
}
\normalsize

\noindent When $M$ is sufficiently large so that  
$q \le \beta/\varepsilon \le 1/K$, we have  
$\Pr[ j \in \mathcal{P}_{\mathrm{low}}^{(t)} ] \ge 1 - 1/K$.

\vspace{0.5em}
\noindent \textbf{Step 3: Expectation over all prototypes.}  
Taking the expectation,
{\small
\begin{equation}
\begin{aligned}
\mathbb{E}\bigl[ |\mathcal{P}_{\mathrm{low}}^{(t)}| \bigr]
&= \sum_{j \in \mathcal{P}_{\mathrm{core}}}
     \Pr[ j \in \mathcal{P}_{\mathrm{low}}^{(t)} ]
  + \sum_{j \in \mathcal{P}_{\mathrm{nc}}}
     \Pr[ j \in \mathcal{P}_{\mathrm{low}}^{(t)} ] \\[0.2em]
&\ge 0
  + \bigl(1 - \tfrac{1}{K} \bigr)\, |\mathcal{P}_{\mathrm{nc}}| 
= \bigl(1 - \tfrac{1}{K} \bigr)\, \bigl( K - |\mathcal{P}_{\mathrm{core}}| \bigr).
\end{aligned}
\tag{B.24}
\end{equation}
}
\normalsize

\noindent Since $|\mathcal{P}_{\mathrm{core}}| \le K_{U}$, it follows that
{\small
\begin{equation}
\mathbb{E}\bigl[ |\mathcal{P}_{\mathrm{low}}^{(t)}| \bigr]
\ge \bigl(1 - \tfrac{1}{K} \bigr) \bigl( K - K_{U} \bigr).
\tag{B.25}
\end{equation}
}
\normalsize

\vspace{0.5em}
\noindent \textbf{Step 4: Asymptotics for $K \gg K_{U}$.}  
When $K \gg K_{U}$, we observe that  
$(1 - \tfrac{1}{K}) (K - K_{U})  
\approx (1 - \tfrac{K_{U}}{K}) K - 1$.  
Hence, for all $t \le T_0$,
{\small
\begin{equation}
\boxed{\,
\mathbb{E}\!\Bigl[
  \bigl| \{\, j \mid u_j^{(t)} \le \varepsilon M/K \} \bigr|
\Bigr]
\ge \Bigl(1 - \tfrac{K_{U}}{K} \Bigr) K - 1
\approx \bigl(1 - \tfrac{K_{U}}{K} \bigr) K
\,}
\tag{B.26}
\end{equation}
}
\normalsize

\hfill$\square$

\section{OW-DFA-40 Benchmark}
\label{sec:supp_B}
As stated in the ``OW-DFA-40 Benchmark'' and ``Evaluation Metrics'' parts of Sec.~5 in the main text, the details of the benchmark construction and metric calculations are presented below.

\subsection{Construction Details}
\label{sec:supp_B1}
\subsubsection{Overview.}
As shown in \cref{tab:dataset}, we incorporate 20 additional state-of-the-art deepfake techniques covering five major forgery types:
Face Swapping~\cite{mobileswap, uniface, blendface, e4s, facedancer},
Face Reenactment~\cite{oneshot, tpsmm, lia, dagan, sadtalker, mcnet, hyperreenact},
Face Editing~\cite{e4e},
Entire Face Synthesis~\cite{stylegan3, styleganxl, sd2.1},
and advanced Diffusion-based Generation~\cite{ditxl, rddm, pixart, sit}.
The partitioning of labeled and unlabeled data follows the original OW-DFA benchmark~\cite{cpl}, as shown in \cref{tab:owdfa-split}.
We split all data into training and testing sets with an 80\%–20\% ratio, also consistent with the OW-DFA benchmark.

\noindent \textbf{Protocol-1} constructs the dataset using 40 deepfake methods along with \textit{real} faces.
These methods are highlighted in gray in \cref{tab:dataset}, and the known/novel split follows the original OW-DFA protocol for methods inherited from OW-DFA.
For newly introduced methods, we randomly assign them to either known or novel categories and fix the partition.
In total, Protocol-1 includes 19 known and 22 novel classes.

\noindent \textbf{Protocol-2} builds upon Protocol-1 and considers a more challenging setting, where new deepfake paradigms (i.e., Entire Face Synthesis and Diffusion-based Generation) emerge.
To simulate this, all classes under Entire Face Synthesis and Diffusion-based Generation in Protocol-1 are reclassified as novel.
As a result, Protocol-2 contains 13 known and 28 novel classes.

\noindent \textbf{Protocol-3} also builds upon Protocol-1 but introduces more known classes to simulate a setting with more supervision.
In total, Protocol-3 includes 29 known and 12 novel classes.

\begin{table*}[h!]
\small
\setlength{\tabcolsep}{3.5pt}
\renewcommand{\arraystretch}{0.85}
\begin{center}
\caption{Overview of Deepfake Methods in the OW-DFA-40 Benchmark. Methods marked with \ding{115} are newly introduced in OW-DFA-40. Classes with gray backgrounds indicate the Protocol-1 partition. Protocols 2 and 3 are derived from Protocol-1, with modified categories highlighted using \fcolorbox{blue}{gray!10}{~}.}
\label{tab:dataset}
%
% \begin{tabular}{@{}llccccc@{}}
\begin{tabular}{@{}llcc>{\columncolor{gray!20}} cc c@{}}
\toprule
\textbf{DeepFake Type} & \textbf{Attribution Class} & \textbf{Venue} & \textbf{Real Data Source} & \textbf{Protocol-1} & \textbf{Protocol-2} & \textbf{Protocol-3} \\
\midrule

\multirow{1}{*}{Real Face} 
  & Real & - & FF++ \& Celeb-DF & Known &  Known & Known   \\

 \midrule

% Face-swapping (10 rows)
\multirow{10}{*}{Face Swapping} 
  & S1.Deepfakes~\shortcite{Deepfakes}     & None 2017   & FF++           & Known & Known & Known \\
  & S2.FaceSwap~\shortcite{faceswap}      & None 2018   & ForgeryNet~\shortcite{forgerynet}     & Novel  & Novel & Novel \\
  & S3.DeepFaceLab~\shortcite{Deepfacelab}   & ArXiv 2020  & FF++           & Known  & Known & Known \\
  & S4.FaceShifter~\shortcite{faceshifter}   & CVPR 2020   & ForgeryNet~\shortcite{forgerynet}     & Novel  & Novel & Novel  \\
  & S5.FSGAN~\shortcite{fsgan}         & ICCV 2019   & ForgeryNet~\shortcite{forgerynet}     & Novel  & Novel & \fcolorbox{blue}{gray!10}{Known} \\

  & \ding{115} S6.BlendFace~\shortcite{blendface}   & ICCV 2023 & FF++ \& Celeb-DF  & Known  & Known & Known \\
  & \ding{115} S7.UniFace~\shortcite{uniface}     & ECCV 2022 & FF++ \& Celeb-DF  & Novel  & Novel & Novel  \\
  & \ding{115} S8.MobileSwap~\shortcite{mobileswap}  & AAAI 2022 & FF++ \& Celeb-DF  & Known  & Known & Known \\
  & \ding{115} S9.e4s~\shortcite{e4s}         & CVPR 2023 & FF++ \& Celeb-DF  & Novel  & Novel & \fcolorbox{blue}{gray!10}{Known}  \\
  & \ding{115} S10.FaceDancer~\shortcite{facedancer}  & WACV 2023 & FF++ \& Celeb-DF  & Known  & Known & Known \\

\midrule

\multirow{12}{*}{Face Reenactment} 
  & R1.Face2Face~\shortcite{face2face}         & CVPR 2016       & FF++           & Known  & Known & Known \\
  & R2.NeuralTextures~\shortcite{Neuraltextures}    & SIGGRAPH 2019   & FF++           & Novel  & Novel & \fcolorbox{blue}{gray!10}{Known}  \\
  & R3.FOMM~\shortcite{fomm}              & NeurIPS 2019    & ForgeryNet~\shortcite{forgerynet}     & Known  & Known & Known \\
  & R4.ATVG-Net~\shortcite{ATVGNet}          & CVPR 2019       & ForgeryNet~\shortcite{forgerynet}     & Novel  & Novel & Novel  \\
  & R5.Talking-Head-Video & ICASSP 2022     & ForgeryNet~\shortcite{forgerynet}     & Novel  & Novel & \fcolorbox{blue}{gray!10}{Known}  \\

  & \ding{115} R6.TPSMM~\shortcite{tpsmm}           & CVPR 2022     & FF++ \& Celeb-DF  & Novel  & Novel & \fcolorbox{blue}{gray!10}{Known}  \\
  & \ding{115} R7.LIA~\shortcite{lia}             & ICLR 2022     & FF++ \& Celeb-DF  & Known  & Known & Known \\
  & \ding{115} R8.DaGAN~\shortcite{dagan}           & CVPR 2022     & FF++ \& Celeb-DF  & Novel  & Novel & Novel  \\
  & \ding{115} R9.SadTalker~\shortcite{sadtalker}       & CVPR 2023     & FF++ \& Celeb-DF  & Known  & Known & Known \\
  & \ding{115} R10.MCNet~\shortcite{mcnet}           & ICCV 2023     & FF++ \& Celeb-DF  & Novel  & Novel & Novel  \\
  & \ding{115} R11.HyperReenact~\shortcite{hyperreenact}    & ICCV 2023     & FF++ \& Celeb-DF  & Known  & Known & Known \\
  & \ding{115} R12.OneShot~\shortcite{oneshot}         & CVPR 2021     & FF++ \& Celeb-DF  & Novel  & Novel & Novel  \\

\midrule

% Face Editing (6 rows)
\multirow{6}{*}{Face Editing} 
  & E1.MaskGAN~\shortcite{maskgan}   & CVPR 2020   & ForgeryNet~\shortcite{forgerynet}  & Known  & Known & Known \\
  & E2.StarGAN2~\shortcite{starganv2}  & CVPR 2020   & ForgeryNet~\shortcite{forgerynet}  & Novel  & Novel & Novel  \\
  & E3.SC-FEGAN~\shortcite{scfegan}  & ICCV 2019   & ForgeryNet~\shortcite{forgerynet}  & Novel  & Novel & \fcolorbox{blue}{gray!10}{Known}  \\
  & E4.FaceAPP~\shortcite{faceapp}   & None 2017   & DFFD~\shortcite{dffd}        & Known  & Known & Known \\
  & E5.StarGAN~\shortcite{stargan}   & CVPR 2018   & DFFD~\shortcite{dffd}        & Novel  & Novel & Novel  \\

  & \ding{115} E6.e4e~\shortcite{e4e}     & SIGGRAPH 2021 & FF++ \& Celeb-DF & Novel  & Novel & \fcolorbox{blue}{gray!10}{Known}  \\

\midrule

% Entire Face Synthesis (7 rows)
  & G1.StyleGAN2~\shortcite{styleganv2}    & CVPR 2020     & ForgeryNet~\shortcite{forgerynet}   & Novel  & Novel & Novel  \\
  & G2.StyleGAN~\shortcite{stylegan}     & CVPR 2019     & DFFD~\shortcite{dffd}         & Known  & \fcolorbox{blue}{gray!10}{Novel} & Known \\
  & G3.PGGAN~\shortcite{pggan}        & ICLR 2018     & DFFD~\shortcite{dffd}         & Novel  & Novel & \fcolorbox{blue}{gray!10}{Known}  \\
  & G4.CycleGAN~\shortcite{cyclegan}     & ICCV 2017     & ForgeryNIR~\shortcite{forgerynir}   & Known  & \fcolorbox{blue}{gray!10}{Novel} & Known \\
  & G5.StyleGAN2~\shortcite{styleganv2}    & CVPR 2020     & ForgeryNIR~\shortcite{forgerynir}   & Novel  & Novel & \fcolorbox{blue}{gray!10}{Known}  \\

  Entire Face Synthesis \& & \ding{115} G6.StyleGAN-XL~\shortcite{styleganxl} & SIGGRAPH 2022 & FF++ \& Celeb-DF & Novel  & Novel & Novel  \\
  Diffusion-Generated Faces & \ding{115} G7.SD-2.1~\shortcite{sd2.1}     & CVPR 2022   & FF++ \& Celeb-DF & Known  & \fcolorbox{blue}{gray!10}{Novel} & Known \\
  & \ding{115} G8.RDDM~\shortcite{rddm}           & CVPR 2024     & FF++ \& Celeb-DF & Novel & Novel   & Novel  \\
  & \ding{115} G9.PixArt-$\alpha$~\shortcite{pixart} & ICLR 2024     & FF++ \& Celeb-DF & Known  & \fcolorbox{blue}{gray!10}{Novel} & Known \\
  & \ding{115} G10.DiT-XL/2~\shortcite{ditxl}       & ICCV 2023     & FF++ \& Celeb-DF & Novel & Novel   & \fcolorbox{blue}{gray!10}{Known}  \\
  & \ding{115} G11.SiT-XL/2~\shortcite{sit}       & ECCV 2024     & FF++ \& Celeb-DF & Known  & \fcolorbox{blue}{gray!10}{Novel} & Known \\
  & \ding{115} G12StyleGAN3~\shortcite{stylegan3}  & NeurIPS 2021 & FF++ \& Celeb-DF & Known  & \fcolorbox{blue}{gray!10}{Novel} & Known \\

\bottomrule
\end{tabular}

\end{center}
\end{table*}

\begin{table}[t]
\centering
\caption{Data partitioning in the OW-DFA task, following the OW-DFA benchmark~\cite{cpl}. We divide samples into labeled and unlabeled subsets across known and unknown forgery types.}
\label{tab:owdfa-split}
\small
\begin{tabular}{lcc}
\toprule
\textbf{Forgery Type} & \textbf{\# Labeled} & \textbf{\# Unlabeled} \\
\midrule
Known forgeries (non-\textit{real}) & 1,500 & 500 \\
Known forgeries (\textit{real})     & 15,000 & 5,000 \\
Unknown forgeries                   & -- & 1,500 \\
\bottomrule
\end{tabular}
\end{table}

\subsubsection{Construction Details of the New Attribution Category}

\par
\noindent \textbf{1. MobileSwap}~\cite{mobileswap} leverages a lightweight identity-aware dynamic network with an identity injection module that predicts and modulates weights on-the-fly for subject-agnostic face swapping. It contains only 0.50M parameters and requires just 0.33 GFLOPs per frame, enabling real-time video swapping on mobile devices. Thanks to knowledge distillation training and adaptive loss re-weighting, MobileSwap achieves synthesis quality comparable to teacher models and other state-of-the-art methods, while maintaining minimal computational cost. We generate fake face images using the official code available at: \url{https://github.com/Seanseattle/MobileFaceSwap}.
\begin{figure}[h]
  \centering
  \includegraphics[width=0.85\linewidth]{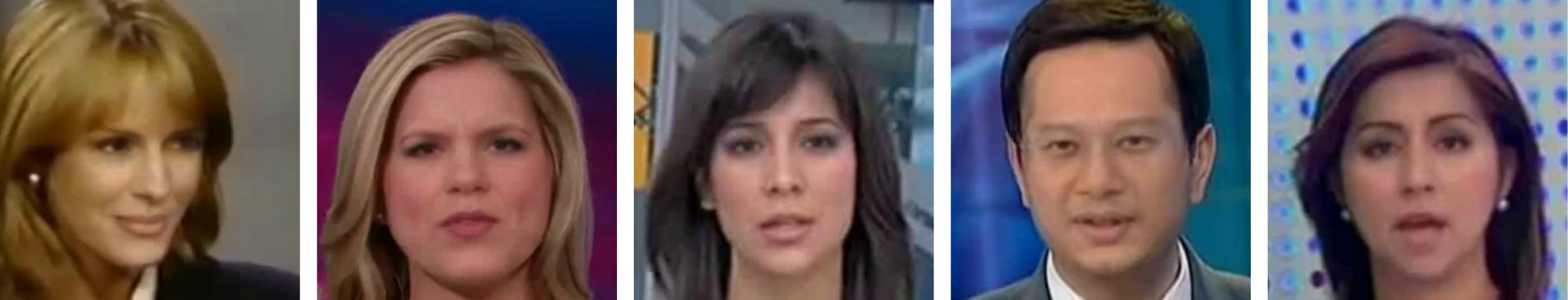}
  %\vspace{-.07in}
\caption{Example images generated by MobileSwap.}
% \vspace{-.07in}
  \label{fig:examples_mobileswap}
\end{figure}

\par
\noindent \textbf{2. UniFace}~\cite{uniface} introduces an end-to-end unified framework that simultaneously performs face reenactment and swapping by unsupervisedly disentangling identity and attribute representations.
It employs an Attribute Transfer module powered by learned Feature Displacement Fields for fine-grained attribute migration, and an Identity Transfer module for adaptively fusing identity features.
This joint design improves identity consistency in reenactment and attribute preservation in swapping. We generate fake face images using the official code available at: \url{https://github.com/xc-csc101/UniFace}.

\begin{figure}[h]
  \centering
  \includegraphics[width=0.85\linewidth]{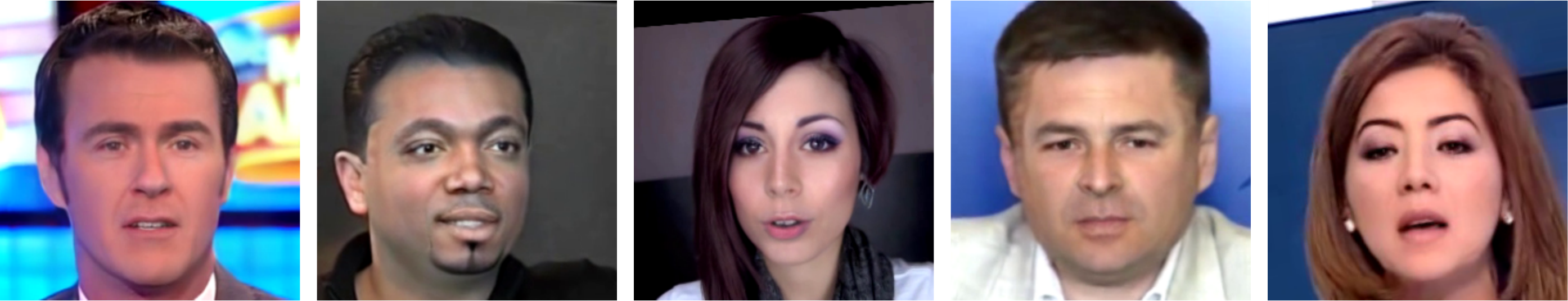}
  %\vspace{-.07in}
\caption{Example images generated by UniFace.}
  \label{fig:examples_uniface}
\end{figure}

\par
\noindent \textbf{3. BlendFace}~\cite{blendface} trains an identity encoder on blended faces whose attributes have been replaced, thereby reducing biases such as hairstyle and head shape, and alleviating identity–attribute entanglement.
The disentangled identity features are then fed into generators and also used as an identity loss to guide face-swapping networks.
Experiments demonstrate that BlendFace improves identity–attribute disentanglement in face swapping while maintaining quantitative performance comparable to previous methods. The fake face images are generated using the official code available at: \url{https://github.com/mapooon/BlendFace}.

\begin{figure}[h]
  \centering
  \includegraphics[width=0.85\linewidth]{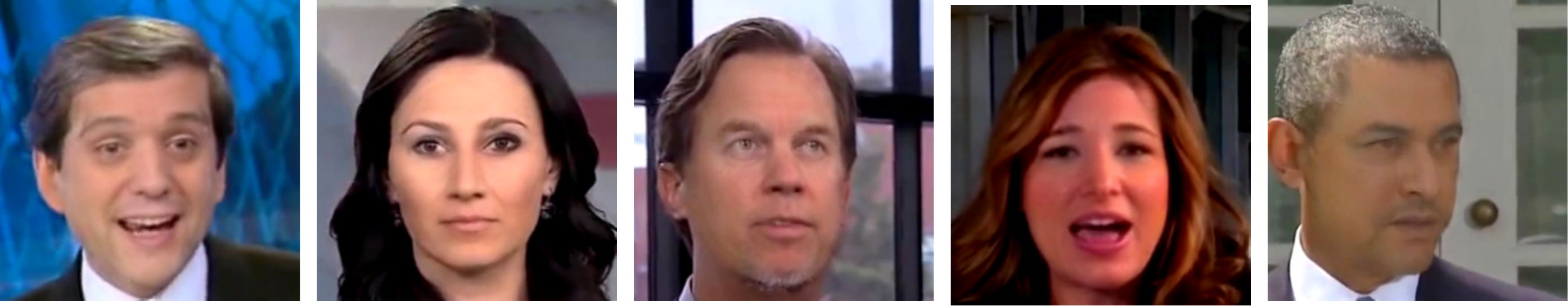}
  %\vspace{-.07in}
\caption{Example images generated by BlendFace.}
  \label{fig:examples_blendface}
\end{figure}

\par
\noindent \textbf{4. e4s}~\cite{e4s} employs Regional GAN Inversion (RGI) to explicitly disentangle the shape and texture of facial components, enabling fine-grained face swapping. It utilizes a multi-scale, mask-guided encoder and injection module, such that swapping in StyleGAN’s latent space reduces to exchanging regional style codes and masks. This design supports global, local, and user-controlled partial swaps, while inherently handling occlusions. e4s preserves geometric and texture details even in high-resolution images. We generate fake face images using the official code available at: \url{https://github.com/e4s2022/e4s}.

\begin{figure}[h]
  \centering
  \includegraphics[width=0.85\linewidth]{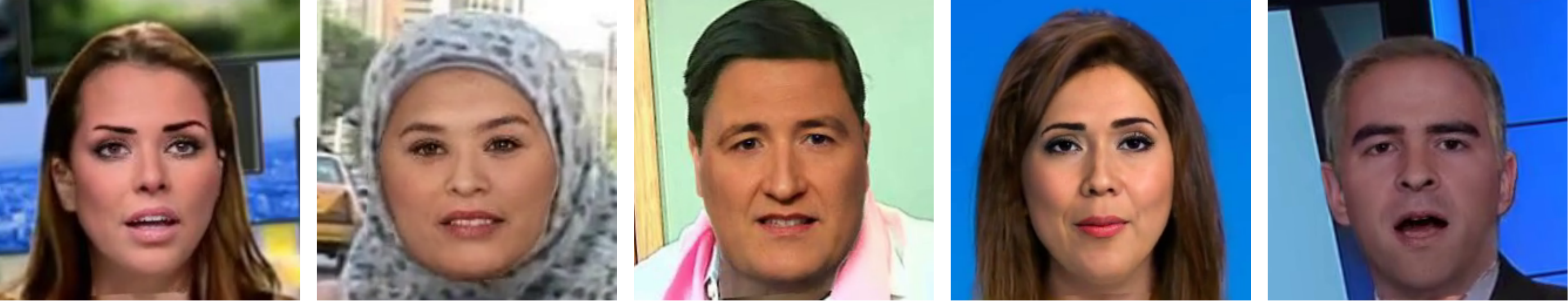}
  %\vspace{-.07in}
\caption{Example images generated by e4s.}
  \label{fig:examples_e4s}
\end{figure}

\par
\noindent \textbf{5. FaceDancer}~\cite{facedancer} is a single-stage framework that introduces Adaptive Feature Fusion Attention to blend identity-conditioned and attribute features without requiring any extra facial segmentation, enabling subject-agnostic face swapping and identity transfer. 
FaceDancer employs Interpreted Feature Similarity Regularization to leverage intermediate features from the identity encoder, achieving high-fidelity identity transfer while preserving head pose, expression, lighting, and occlusions in the target face. 
FaceDancer delivers superior performance in terms of identity consistency and pose preservation. We generate fake face images using the official code available at: \url{https://github.com/felixrosberg/FaceDancer}.

\begin{figure}[h]
  \centering
  \includegraphics[width=0.85\linewidth]{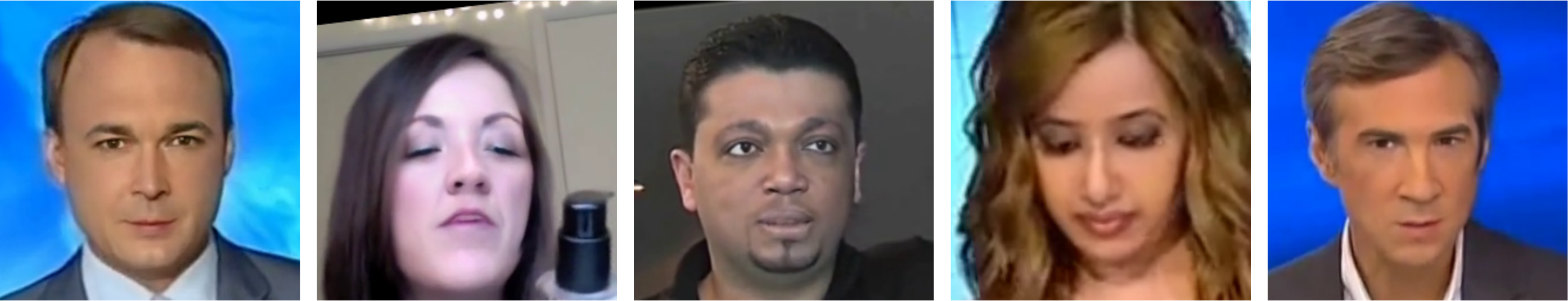}
  %\vspace{-.07in}
\caption{Example images generated by FaceDancer.}
  \label{fig:examples_facedancer}
\end{figure}

\par
\noindent \textbf{6. OneShot}~\cite{oneshot} synthesizes high-fidelity talking-head videos from a single source image and a driving clip by unsupervisedly disentangling identity and motion through a novel keypoint representation.
OneShot’s compact keypoint stream delivers H.264-level visual quality for video conferencing while consuming only one-tenth of the bandwidth, enabling real-time head-pose manipulation.
OneShot excels in both bandwidth efficiency and interactive realism. We generate fake face images using the official code available at: \url{https://github.com/zhanglonghao1992/One-Shot_Free-View_Neural_Talking_Head_Synthesis}.

\begin{figure}[h]
  \centering
  \includegraphics[width=0.85\linewidth]{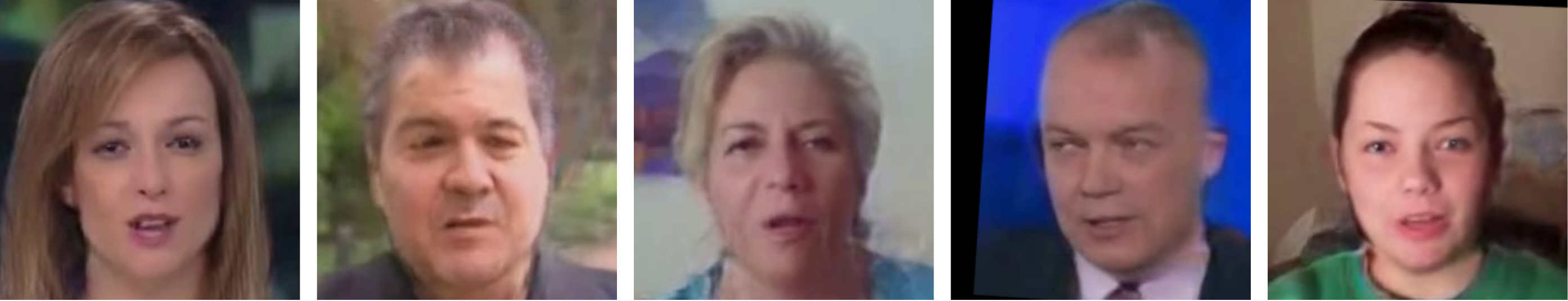}
  %\vspace{-.07in}
\caption{Example images generated by OneShot.}
  \label{fig:examples_oneshot}
\end{figure}

\par
\noindent \textbf{7. TPSMM}~\cite{tpsmm} employs thin-plate-spline motion estimation to produce flexible optical flow that warps source features into the driving image domain, enabling unsupervised motion transfer across large pose gaps.
TPSMM leverages multi-resolution occlusion masks for effective feature fusion and introduces auxiliary losses that enforce clear module specialization, resulting in realistic completion of missing regions.
It can animate diverse objects—including talking faces, human bodies, and pixel art—and outperforms state-of-the-art methods on most benchmarks, particularly in pose-related metrics.
We generate fake face images using the official code available at: \url{https://github.com/yoyo-nb/Thin-Plate-Spline-Motion-Model}.

\begin{figure}[h]
  \centering
  \includegraphics[width=0.85\linewidth]{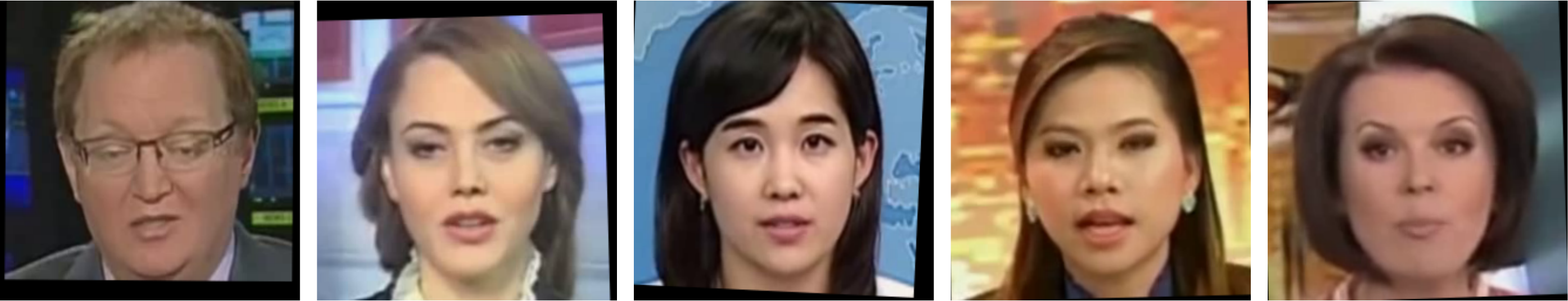}
  %\vspace{-.07in}
\caption{Example images generated by TPSMM.}
  \label{fig:examples_tpsmm}
\end{figure}

\par
\noindent \textbf{8. LIA}~\cite{lia} is a self-supervised autoencoder that animates still images by linearly navigating the latent space, without relying on any explicit structural representation.
LIA learns a set of orthogonal motion directions whose linear combinations encode motion as latent-code displacements, allowing it to handle large appearance variations between source images and driving videos.
LIA demonstrates strong performance in terms of generated video quality.
We generate fake face images using the official code available at: \url{https://github.com/wyhsirius/LIA}.

\begin{figure}[h]
  \centering
  \includegraphics[width=0.85\linewidth]{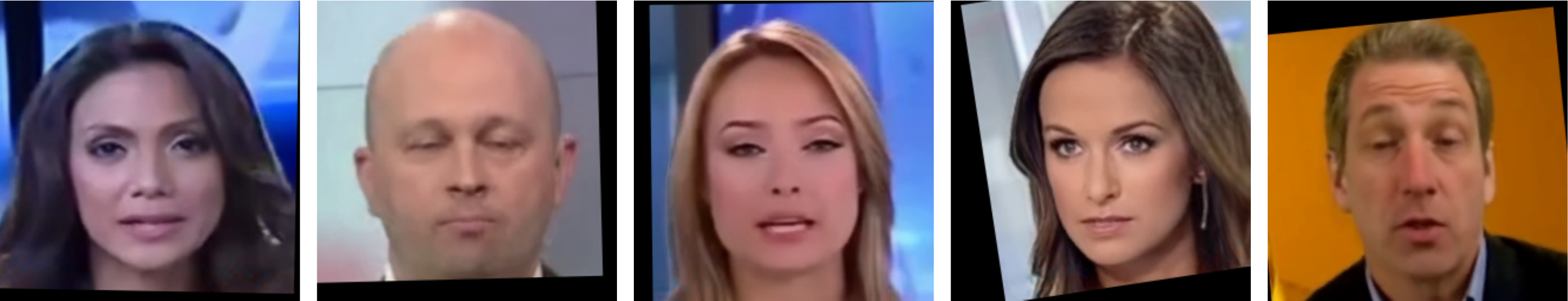}
  %\vspace{-.07in}
\caption{Example images generated by LIA.}
  \label{fig:examples_lia}
\end{figure}

\par
\noindent \textbf{9. DaGAN}~\cite{dagan} self-supervisedly learns dense facial depth from videos without requiring expensive 3D annotations.
DaGAN leverages the recovered depth to estimate sparse keypoints and construct 3D-aware cross-modal attention, which generates accurate motion fields for warping source image features.
It produces highly realistic talking-head videos and significantly outperforms previous methods on unseen faces.
We generate fake face images using the official code available at: \url{https://github.com/harlanhong/CVPR2022-DaGAN}.

\begin{figure}[h]
  \centering
  \includegraphics[width=0.85\linewidth]{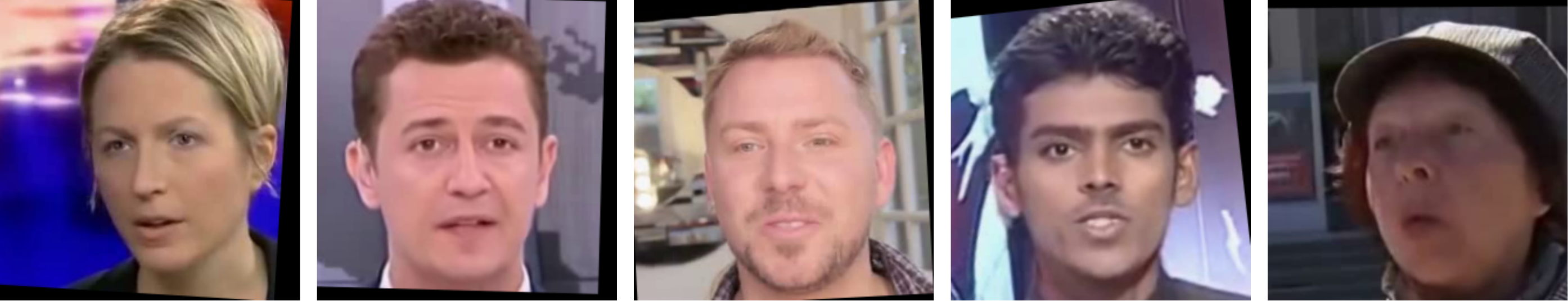}
  %\vspace{-.07in}
\caption{Example images generated by DaGAN.}
  \label{fig:examples_dagan}
\end{figure}

\par
\noindent \textbf{10. SadTalker}~\cite{sadtalker} converts speech audio into 3DMM motion coefficients—specifically, head pose and expression—and feeds them into a novel 3D-aware renderer to synthesize realistic talking-head videos.
SadTalker employs ExpNet to distill accurate expression coefficients from audio and uses a conditional PoseVAE to generate diverse head motions, effectively decoupling expression from pose.
It produces more natural motion and higher video quality than previous approaches.
We generate fake face images using the official code available at: \url{https://github.com/OpenTalker/SadTalker}.

\begin{figure}[h!]
  \centering
  \includegraphics[width=0.85\linewidth]{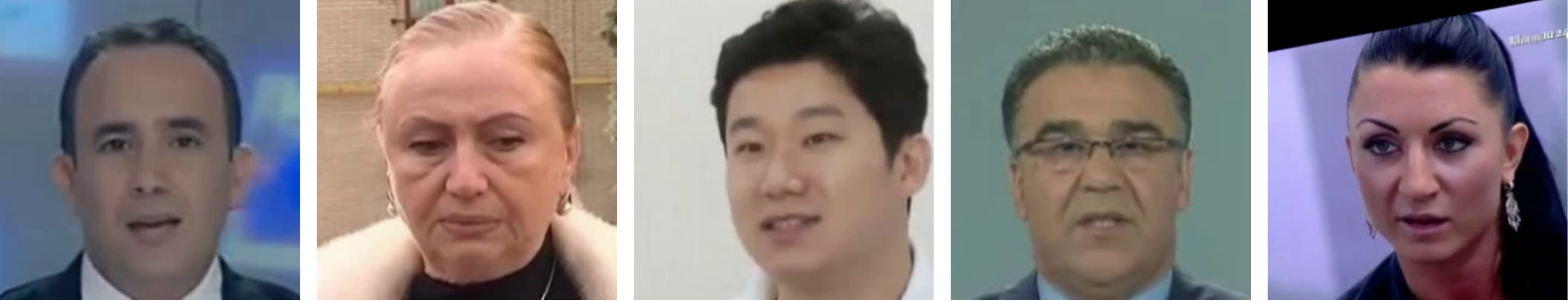}
  %\vspace{-.07in}
\caption{Example images generated by SadTalker.}
  \label{fig:examples_sadtalker}
\end{figure}

\par
\noindent \textbf{11. MCNet}~\cite{mcnet} learns a unified spatial facial meta-memory bank that provides rich structural and appearance priors to compensate for warped source features, enabling high-fidelity talking-head synthesis.
MCNet employs an implicit identity-conditioned query—derived from discrete keypoints—to retrieve the most relevant information from the memory bank, effectively filling in ambiguous or occluded regions caused by complex motions.
It produces fewer artifacts under dramatic pose and expression changes.
We generate fake face images using the official code available at: \url{https://github.com/harlanhong/ICCV2023-MCNET}.

\begin{figure}[h!]
  \centering
  \includegraphics[width=0.85\linewidth]{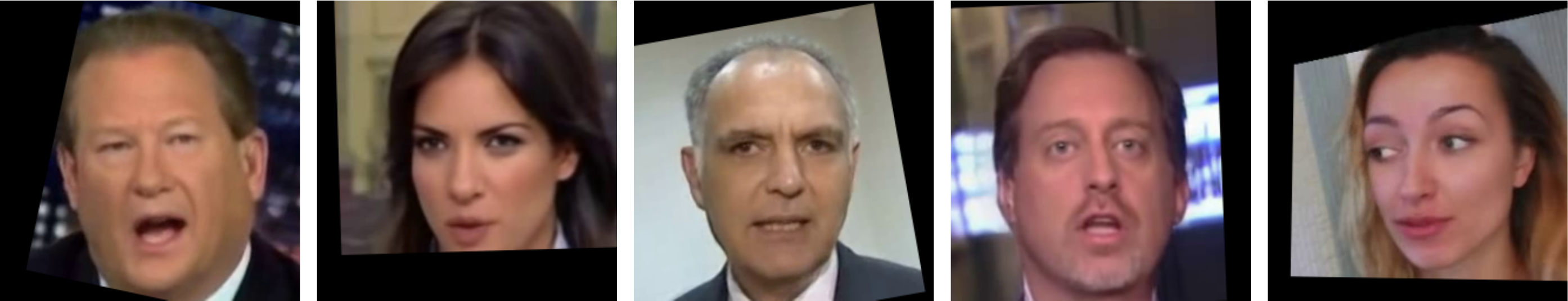}
  %\vspace{-.07in}
\caption{Example images generated by MCNet.}
  \label{fig:examples_mcnet}
\end{figure}

\par
\noindent \textbf{12. HyperReenact}~\cite{hyperreenact} leverages StyleGAN2 latent inversion and a hypernetwork that simultaneously refines identity traits and retargets facial pose, eliminating artifacts from external editing.
HyperReenact operates in a one-shot setting with a single source frame, requires no subject-specific fine-tuning, and maintains photorealism even under extreme head-pose changes during cross-subject reenactment.
It surpasses state-of-the-art methods on VoxCeleb1 and VoxCeleb2, producing virtually artifact-free images with superior quantitative and qualitative performance.
We generate fake face images using the official code available at: \url{https://github.com/StelaBou/HyperReenact}.

\begin{figure}[h!]
  \centering
  \includegraphics[width=0.85\linewidth]{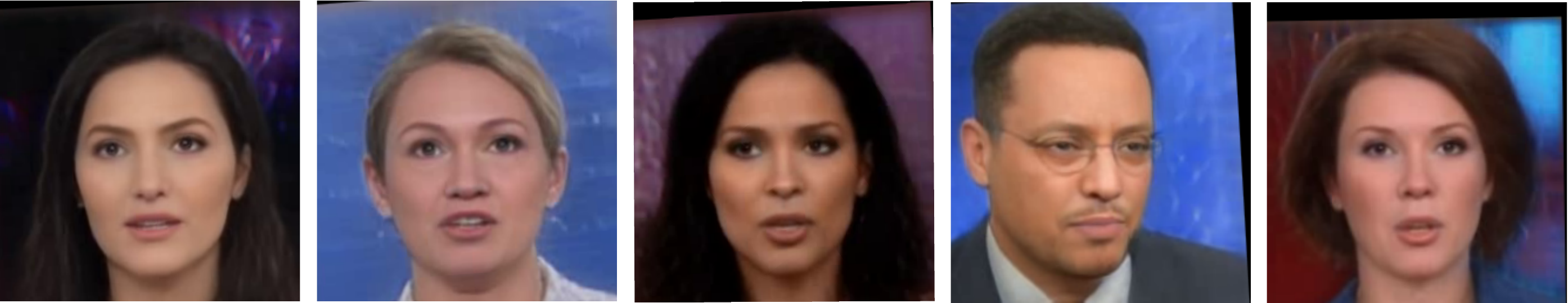}
  %\vspace{-.07in}
\caption{Example images generated by HyperReenact.}
  \label{fig:examples_hyperreenact}
\end{figure}

\par
\noindent \textbf{13. e4e}~\cite{e4e} systematically investigates StyleGAN’s latent space and identifies the trade-offs between distortion–editability and distortion–perception.
e4e applies two encoder design principles to invert real images into edit-friendly regions while maintaining a low reconstruction error.
It achieves superior real-image editing in challenging domains with only a minor drop in reconstruction accuracy.
We generate fake face images using the official code available at: \url{https://github.com/omertov/encoder4editing}.

\begin{figure}[h!]
  \centering
  \includegraphics[width=0.85\linewidth]{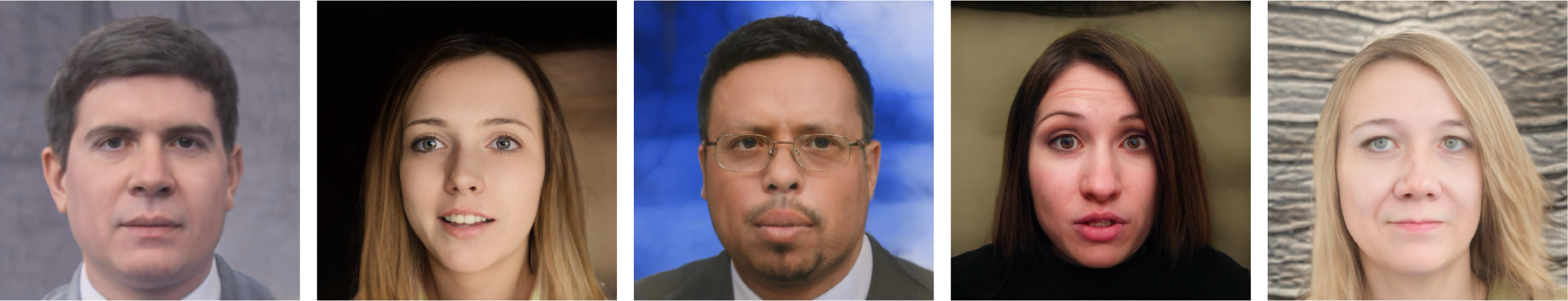}
  %\vspace{-.07in}
\caption{Example images generated by e4e.}
  \label{fig:examples_e4e}
\end{figure}

\par
\noindent \textbf{14. StyleGAN3}~\cite{stylegan3} treats every signal in the network as continuous and introduces small, generic architectural tweaks that eliminate aliasing, ensuring that details adhere to object surfaces rather than absolute pixel coordinates.
StyleGAN3 matches StyleGAN2’s FID while achieving full translation and rotation equivariance down to sub-pixel scales through radically different internal representations.
It therefore lays the groundwork for generative models that are far better suited for video and animation, offering physically consistent and easily animatable outputs.
We generate fake face images using the official code available at: \url{https://github.com/NVlabs/stylegan3}.

\begin{figure}[h!]
  \centering
  \includegraphics[width=0.85\linewidth]{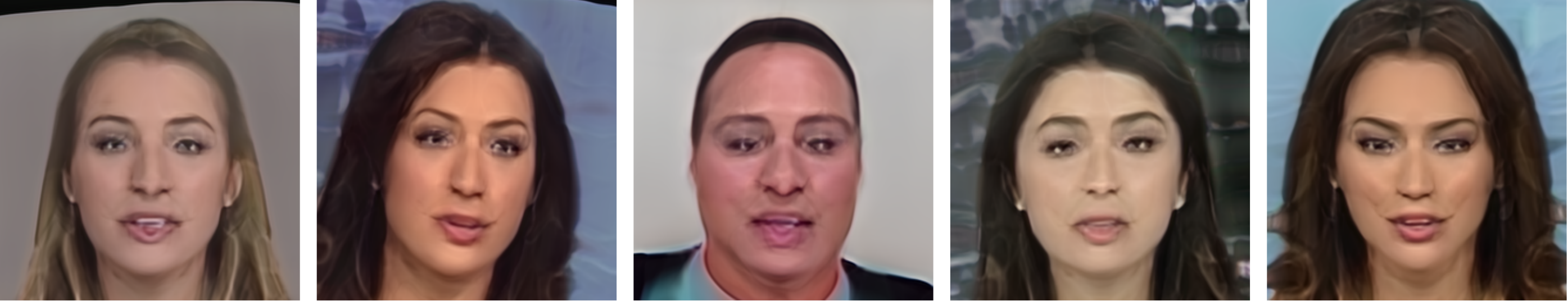}
  %\vspace{-.07in}
\caption{Example images generated by StyleGAN3.}
  \label{fig:examples_stylegan3}
\end{figure}

\par
\noindent \textbf{15. StyleGAN-XL}~\cite{styleganxl} leverages the Projected GAN paradigm and a progressive growing schedule to successfully train the latest StyleGAN3 on large, unstructured datasets such as ImageNet.
StyleGAN-XL sets a new state of the art in large-scale image synthesis, becoming the first model to generate 1024\textsuperscript{2}-resolution images at this dataset scale.
It also enables inversion and editing on a wide range of images, going beyond the narrow confines of portraits or specific object classes.
We generate fake face images using the official code available at: \url{https://github.com/autonomousvision/stylegan-xl}.

\begin{figure}[h!]
  \centering
  \includegraphics[width=0.85\linewidth]{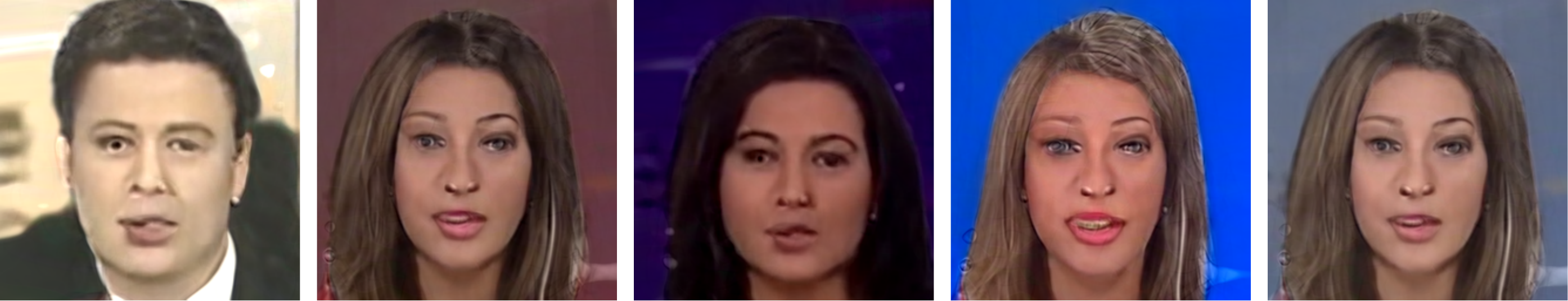}
  %\vspace{-.07in}
\caption{Example images generated by StyleGAN-XL.}
  \label{fig:examples_styleganxl}
\end{figure}

\par
\noindent \textbf{16. Stable-Diffusion-2.1}~\cite{sd2.1} trains diffusion models in the latent space of powerful pretrained autoencoders, greatly reducing computational demand while preserving fine details.
Stable-Diffusion-2.1 incorporates cross-attention layers to handle diverse conditioning inputs such as text and bounding boxes, enabling convolutional high-resolution synthesis.
It sets new state-of-the-art results in image inpainting and class-conditional generation, delivers competitive performance on unconditional, text-to-image, and super-resolution tasks, and offers far more efficient inference than pixel-level diffusion models.
We generate fake face images using the official code available at: \url{https://github.com/Stability-AI/stablediffusion}.

\begin{figure}[h!]
  \centering
  \includegraphics[width=0.85\linewidth]{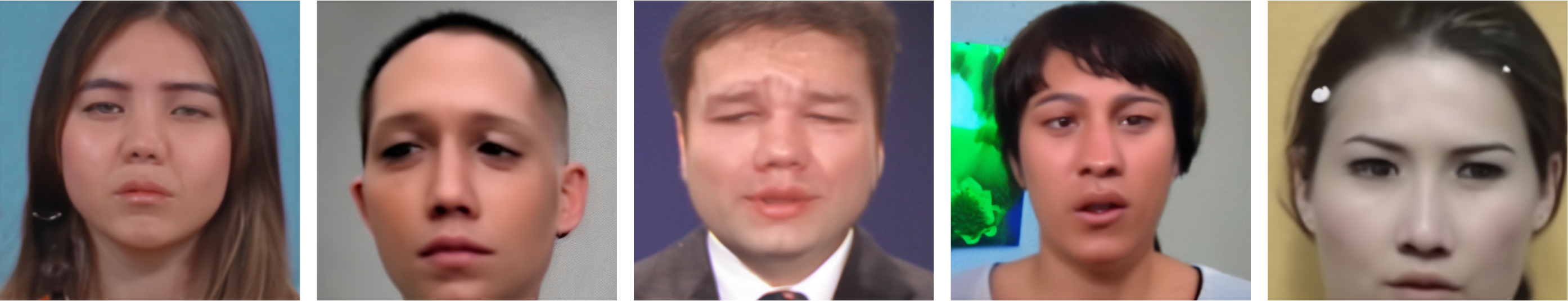}
  %\vspace{-.07in}
\caption{Example images generated by Stable-Diffusion-2.1.}
  \label{fig:examples_sd21}
\end{figure}

\par
\noindent \textbf{17. DiT-XL/2}~\cite{ditxl} replaces the conventional U-Net backbone with a Transformer that performs diffusion on latent image patches, forming a new class of latent diffusion models.
DiT-XL/2 scales favorably: increased forward-pass complexity (GFLOPs)—achieved via deeper or wider transformers or more input tokens—consistently yields lower FID scores.
We generate fake face images using the official code available at: \url{https://github.com/facebookresearch/DiT}.

\begin{figure}[h!]
  \centering
  \includegraphics[width=0.85\linewidth]{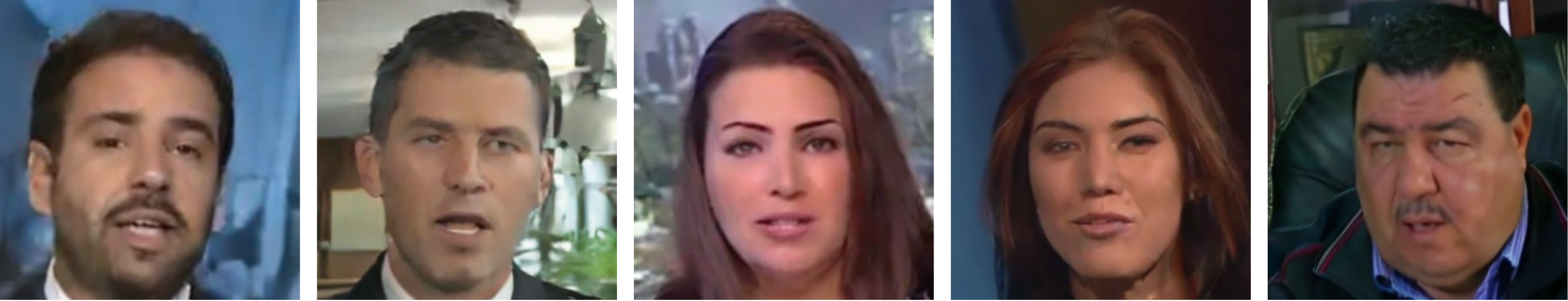}
  %\vspace{-.07in}
\caption{Example images generated by DiT-XL/2.}
  \label{fig:examples_dit}
\end{figure}

\par
\noindent \textbf{18. RDDM}~\cite{rddm} introduces a dual-diffusion framework—residual diffusion and noise diffusion—that decouples the traditional single denoising process into directional and random components.
RDDM’s residual diffusion maps the target image to the degraded input, providing deterministic guidance for restoration, while noise diffusion preserves diversity, unifying image generation and restoration within a single model.
RDDM demonstrates that its sampling aligns with DDPM and DDIM via coefficient transformation and proposes a partially path-independent reverse process for improved interpretability. We generate fake face images using the official code available at: \url{https://github.com/nachifur/RDDM}.

\begin{figure}[h!]
  \centering
  \includegraphics[width=0.85\linewidth]{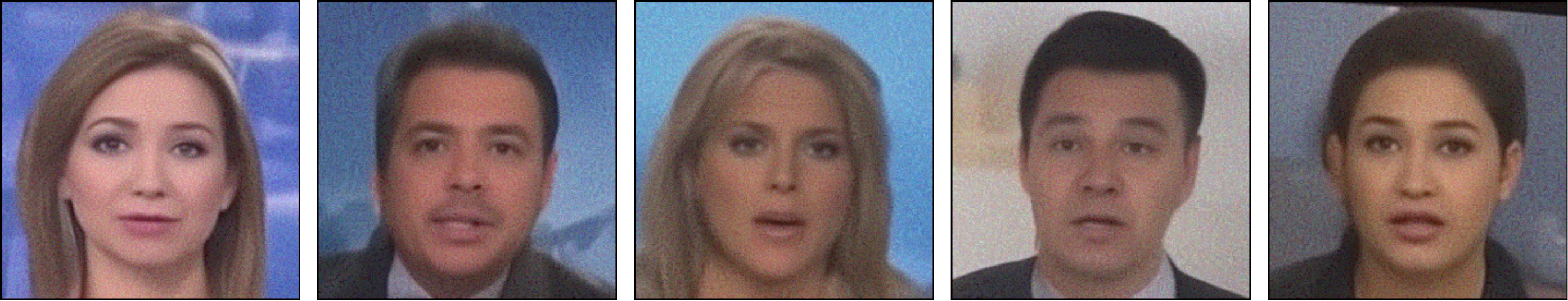}
  %\vspace{-.07in}
\caption{Example images generated by RDDM.}
  \label{fig:examples_rddm}
\end{figure}

\par
\noindent \textbf{19. PixArt-$\alpha$}~\cite{pixart} is a Transformer-based text-to-image diffusion model that generates 1024 px images with quality rivaling that of Imagen, SDXL, and even Midjourney.
It achieves this through three decomposed training stages—pixel dependency learning, text-image alignment, and aesthetic enhancement—an efficient T2I Transformer with cross-attention and a streamlined class-conditioning branch, and highly informative data enriched by pseudo-captions from a large vision-language model.
These designs reduce training time to approximately 10.8\% of Stable Diffusion v1.5 and just 1\% of RAPHAEL, cutting costs by around 90\%. We generate fake face images using the official code available at: \url{https://github.com/PixArt-alpha/PixArt-alpha}.

\begin{figure}[h!]
  \centering
  \includegraphics[width=0.85\linewidth]{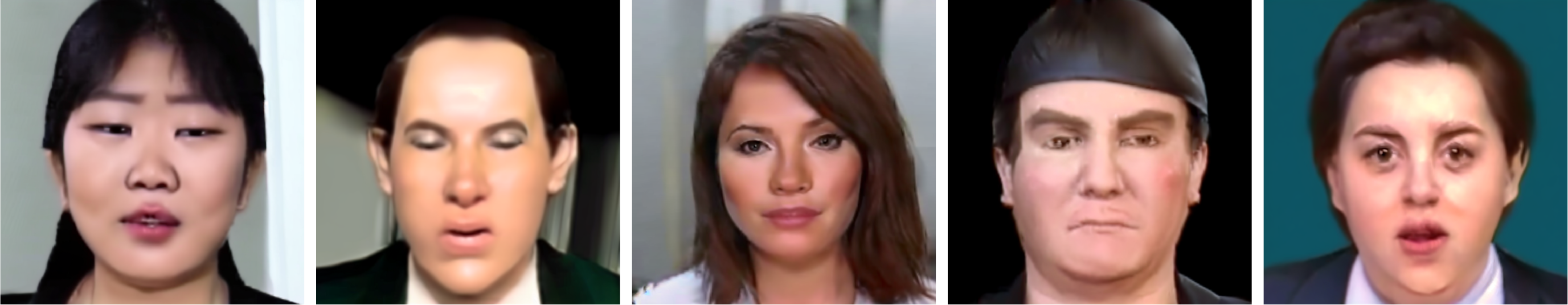}
  %\vspace{-.07in}
\caption{Example images generated by PixArt-$\alpha$.}
  \label{fig:examples_pixart}
\end{figure}

\par
\noindent \textbf{20. SiT}~\cite{sit} builds on Diffusion Transformers but adopts a more flexible interpolant framework that links two distributions in either discrete or continuous time.
The framework modularizes key design choices—objective, interpolant, diffusion coefficient, and deterministic vs. stochastic sampling—enabling targeted improvements in transport-based generative modeling.
Using exactly the same architecture, parameter count, and GFLOPs as DiT, SiT consistently outperforms it on conditional ImageNet at 512×512 and 256×256 resolutions.
By tuning diffusion coefficients after training, the best SiT variants achieve state-of-the-art FID-50K scores of 2.06 and 2.62, respectively. We generate fake face images using the official code available at: \url{https://github.com/willisma/SiT}.

\begin{figure}[h!]
  \centering
  \includegraphics[width=0.85\linewidth]{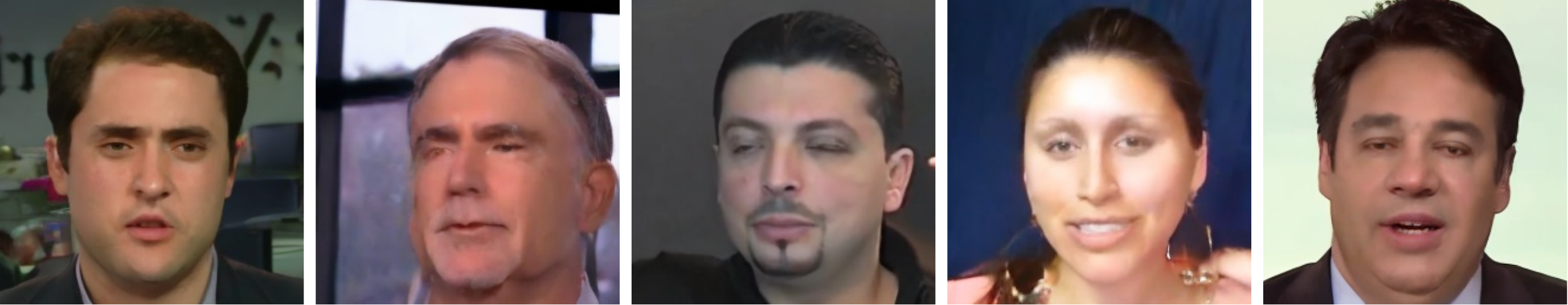}
  %\vspace{-.07in}
\caption{Example images generated by SiT.}
  \label{fig:examples_sit}
\end{figure}

\subsection{Evaluation Protocol}
\label{sec:supp_B2}
\noindent \textbf{Metrics Used.} We adopt clustering validity metrics to objectively evaluate the performance of clustering algorithms. Commonly used metrics include clustering accuracy (ACC), normalized mutual information (NMI), and adjusted Rand index (ARI). Given an input $\mathbf{X} = \{x_1, x_2, \cdots, x_n\}$ with $K$ ground-truth classes, the true clusters are denoted as $\hat{\mathcal{C}} = \{\hat{C}_1, \hat{C}_2, \cdots, \hat{C}_K\}$, and the corresponding ground-truth labels are $\hat{Y} = \{\hat{y}_1, \hat{y}_2, \cdots, \hat{y}_n\}$. After clustering, $K'$ result clusters are produced: $C = \{C_1, C_2, \cdots, C_{K'}\}$, and the predicted sample-level labels are $Y = \{y_1, y_2, \cdots, y_n\}$.

\par
\noindent \textbf{Clustering accuracy (ACC)} measures the proportion of samples whose predicted labels match the ground-truth labels. It lies in the range $[0, 1]$, with higher values indicating better clustering performance:

\begin{equation}
    \text{ACC} = \frac{1}{n} \sum_{i=1}^{n} \Delta(y_i, \text{map}(\hat{y}_i)) \quad 
\Delta(y, \hat{y}_i) = 
\begin{cases}
1, & y = \hat{y}_i \\
0, & y \ne \hat{y}_i
\end{cases} \tag{C.1}
\end{equation}

Here, $\text{map}$ denotes a mapping function that aligns predicted cluster labels with the ground-truth labels, which is optimized using the Hungarian algorithm~\cite{Hungarian}.

\par
\noindent \textbf{Normalized mutual information (NMI)} evaluates the amount of information shared between predicted clusters and ground-truth clusters. It lies in the range $[0, 1]$, where higher values indicate better clustering performance:

\begin{equation}
    \text{NMI} = 
\frac{
\sum_{i=1}^{K} \sum_{j=1}^{K'} |C_i \cap \hat{C}_j| \log \frac{n |C_i \cap \hat{C}_j|}{|C_i| |\hat{C}_j|}
}{
\sqrt{
\left( \sum_{i=1}^{K} |C_i| \log \frac{|C_i|}{n} \right)
\cdot
\left( \sum_{j=1}^{K'} |\hat{C}_j| \log \frac{|\hat{C}_j|}{n} \right)
}
}
\tag{C.2}
\end{equation}

\par
\noindent \textbf{Adjusted Rand index (ARI)} measures the similarity between the clustering results and the ground-truth labels. It lies in the range $[-1, 1]$, with higher values indicating better clustering performance:

\begin{equation}
    \text{ARI} = \frac{2(ad - bc)}{(a + b)(b + d) + (a + c)(c + d)}
    \tag{C.3}
\end{equation}

\begin{align*}
    a &= \{(x_i, x_j) \mid y_i = y_j, \hat{y}_i = \hat{y}_j, i < j \} \\
    b &= \{(x_i, x_j) \mid y_i = y_j, \hat{y}_i \ne \hat{y}_j, i < j \} \\
    c &= \{(x_i, x_j) \mid y_i \ne y_j, \hat{y}_i = \hat{y}_j, i < j \} \\
    d &= \{(x_i, x_j) \mid y_i \ne y_j, \hat{y}_i \ne \hat{y}_j, i < j \}
\end{align*}

\noindent \textbf{Metric Details.}  
Our evaluation metrics follow the evaluation protocol proposed in CPL~\cite{cpl}. Specifically, in the OW-DFA task, since labeled data is available for known categories, the model’s predictions can be directly aligned with the ground-truth labels. Therefore, we compute the Known ACC using standard classification accuracy. For novel categories and the entire test set, we use clustering-based metrics, including clustering accuracy (ACC), normalized mutual information (NMI), and adjusted Rand index (ARI). After applying the Hungarian algorithm~\cite{Hungarian} to find the optimal label assignment for the entire set of test predictions, we then compute All ACC, All NMI, and All ARI. We subsequently split the predictions into known and novel subsets and compute Known ACC, Novel ACC, Novel NMI, and Novel ARI separately.

\section{Implementation Details}
\label{sec:supp_C}
As stated in the ``Implementation Details'' and ``Compared Methods'' parts of Sec.~5 in the main text, the details of the training process (pseudo code) and compared methods are presented below.

\begin{algorithm}[h]
\caption{Pseudocode for CAL}
\label{alg:training}
\KwIn{Training data: labeled set $\mathcal{D}_L$, unlabeled set $\mathcal{D}_U$ (including known and novel forgeries); image encoder $\mathcal{E}(\cdot)$; prototype layer $\mathbf{C}$; convolutional network $\mathcal{G}(\cdot)$; hyperparameters \{training epochs $E$, warm-up epochs $e_0$, number of known categories $K_L$, etc.\}}

\textbf{/* Initialization */} \\
\uIf{$K_U$ is unknown}{
    Set total prototype number $K \leftarrow 10 \times K_L$\;
}
\Else{
    Set total prototype number $K \leftarrow K_L + K_U$\;
}
\textbf{/* CAL Framework */} \\
\For{each epoch $e = 1$ \KwTo $E$}{
    \For{each batch $\mathcal{B}$}{
        \textbf{/* FFE Module */} \\
        $\mathbf{f}_i \leftarrow DCT(\mathbf{x}_i)$ \hfill\tcp{Frequency domain}
        $\mathbf{M}_i \leftarrow \mathcal{G}(\mathbf{f}_i)$ \hfill\tcp{Frequency mask}
        $\mathbf{A}_i \leftarrow IDCT(\mathbf{M}_i \odot \mathbf{f}_i)$ \hfill\tcp{Spatial domain}
        $\mathbf{h}_i \leftarrow Pooling(\mathbf{A}_i \odot \mathcal{E}(\mathbf{x}_i); 1\times1)$ \hfill\tcp{Frequency-guided feature}

        \textbf{/* CCR Module */} \\
        Compute $\mathcal{L}_{\text{ccr}}$ using Eq.~(4) \hfill\tcp{CCR}
        \textbf{/* ACR Module */} \\
        \If{$e > e_0$}{
            Compute $\mathcal{L}_{\text{acr}}$ using Eq.~(6)\hfill\tcp{ACR}
        }
        Compute total loss $\mathcal{L}$ using Eq.~(7)\;

        Perform back-propagation and optimize $\mathcal{E}(\cdot)$, $\mathbf{C}$, and $\mathcal{G}(\cdot)$\;

        \textbf{/* DDP Module */} \\
        \If{$K_U$ is unknown}{
            Merge and update $\mathbf{C}$ using Algorithm~\ref{alg:dpp}
        }
    }
}
\Return Trained model $\text{CAL}(\cdot)$\;
\end{algorithm}

\subsection{CAL Training Details}
\label{sec:supp_C1}
Our training setup largely follows that of CPL~\cite{cpl}. The pseudocode for our CAL method is shown in~\cref{alg:training}. In scenarios where the number of novel forgery types is unknown, we employ the proposed DPP module to initially assign an overestimated number of clusters to the novel classes. During training, prototypes are dynamically merged to estimate the actual number of categories. The pseudocode for this procedure is provided in~\cref{alg:dpp}.

\subsection{Compared Methods Details}
\label{sec:supp_C2}
\par
\noindent \textbf{OwMatch}~\cite{owmatch} tackles open-world semi-supervised learning by combining conditional self-labeling with an open-world hierarchical thresholding scheme, preventing unseen classes from being misclassified as known ones.
It integrates self-labeling from self-supervised learning and consistency regularization from semi-supervised learning to adaptively assign labels or retain an “unknown” state for unlabeled samples.
Rigorous statistical analysis shows that OwMatch provides an unbiased estimator of class distributions over unlabeled data, ensuring reliable label assignment. 
The implementation is based on the official code available at \url{https://github.com/niusj03/OwMatch}. 
Considering image resolution, the hyperparameters are selected according to the Tiny ImageNet setting as described in the original paper. 
The training hyperparameters follow CPL~\cite{cpl}: we use the Adam optimizer with a learning rate of 2e-4, a batch size of 128, and train for 50 epochs.

\par
\noindent \textbf{LPS}~\cite{lps} applies an adaptive synchronizing marginal loss that assigns class-specific negative margins to mitigate bias toward seen classes.
It introduces pseudo-label contrastive clustering, which groups unlabeled samples with the same predicted label to capture the semantics of novel categories.
The implementation is based on the official code available at \url{https://github.com/yebo0216best/LPS-main}. 
Considering image resolution, the hyperparameters are selected according to the ImageNet setting described in the original paper.
The training hyperparameters follow CPL~\cite{cpl}, using the Adam optimizer with a learning rate of 2e-4, a batch size of 128, and 50 training epochs.

\par
\noindent \textbf{SimGCD}~\cite{simgcd} investigates why parametric classifiers underperform in Generalized Category Discovery and identifies two key biases caused by noisy pseudo-labels: the over-prediction of seen classes and the imbalance between seen and novel class distributions.
The study demonstrates that, with high-quality supervision, improved parametric classifiers can match the performance of the previously preferred semi-supervised k-means non-parametric approach.
SimGCD introduces entropy regularization during training to mitigate the bias toward seen categories and to balance the overall class distribution, significantly improving prediction reliability.
The implementation is based on the official code available at \url{https://github.com/CVMI-Lab/SimGCD}.
The training hyperparameters follow those used in CPL~\cite{cpl}, utilizing the Adam optimizer with a learning rate of 2e-4, a batch size of 128, and 50 training epochs.

\par
\noindent \textbf{ProtoGCD}~\cite{protogcd} introduces a unified and unbiased prototype-learning framework that jointly models both known and novel classes through shared prototypes and a single learning objective.
It employs a dual-level adaptive pseudo-labeling scheme to mitigate confirmation bias and incorporates two regularization terms that jointly guide representation learning for Generalized Category Discovery (GCD).
To enhance practicality, ProtoGCD also proposes a criterion for estimating the number of unseen classes and extends its framework to outlier detection, achieving task-level unification.
The implementation is based on the official code available at \url{https://github.com/mashijie1028/ProtoGCD}.
The training hyperparameters follow those used in CPL~\cite{cpl}, employing the Adam optimizer with a learning rate of 2e-4, a batch size of 128, and 50 training epochs.

\par
\noindent \textbf{LegoGCD}~\cite{legogcd} addresses the issue of catastrophic forgetting in Generalized Category Discovery by seamlessly integrating with existing methods to enhance novel-class discrimination while retaining knowledge of known classes.
It introduces Local Entropy Regularization (LER) to refine the distribution of potential known-class samples in the unlabeled data, thereby reinforcing the memory of previously learned patterns.
A Dual-View Kullback–Leibler constraint (DKL) is employed to enforce consistent prediction distributions across two augmented views of the same image, reducing prediction mismatches and yielding more reliable known-class candidates.
The implementation is based on the official code available at \url{https://github.com/Cliffia123/LegoGCD}.
The training hyperparameters follow those used in CPL~\cite{cpl}, using the Adam optimizer with a learning rate of 2e-4, a batch size of 128, and 50 training epochs.

\begin{algorithm}[t]
\caption{Pseudocode for Dynamic Prototype Pruning (DPP)}
\label{alg:dpp}
\KwIn{Prototypes $\mathbf{C}\!\in\!\mathbb{R}^{K\times d}$, Usage count $\{u_j\}_{j=1}^K$, Number of known prototypes $K_L$, Coverage threshold $\mathcal{I}$.}

\textbf{/* Stage-I: High-confidence Prototype Identification */}\\
$\mathbf{C}_{known} \leftarrow \mathbf{C}_{1:K_L}$,\;
$\mathbf{C}_{novel} \leftarrow \mathbf{C}_{K_L+1:K}$\hfill\tcp{Prototypes}
$\mathcal{U} \leftarrow u_{K_L+1:K}$

Sort $\mathcal{U}$ in descending order\;
Compute cumulative coverage $r_k \leftarrow \mathrm{cumsum}(u_k)/\sum{u_k}$ for $u_k \in \mathcal{U}$\;
$k^{\star} \leftarrow \min\{k\,|\,r_k\ge \mathcal{I}\}$

$\mathbf{C}_{high} \leftarrow$ top-$k^{\star}$ usage prototypes of $\mathbf{C}_{novel}$\; \hfill\tcp{High-confidence prototypes}

$\mathbf{C}_{cand} = \mathbf{C}_{novel} \setminus \mathbf{C}_{high}$. \hfill\tcp{Candidate}

\textbf{/* Stage-II: Low-confidence Prototype Filtering */}\\

$\overline{u} \leftarrow \mathrm{mean}(u_k)$ for $\mathbf{c}_k\in\mathbf{C}_{cand}$\hfill\tcp{Mean usage}
$\Delta r_k = r_k-r_{k-1}$\;
$\overline{\Delta r} \leftarrow \mathrm{mean}(\Delta r_k)$ for $\mathbf{c}_k\in\mathbf{C}_{cand}$\hfill\tcp{Mean first-order diff}

$\mathbf{C}_{low} \leftarrow 
  \{\,\mathbf{c}_k\in\mathbf{C}_{cand}
     \mid u_k\le \overline{u} \land \Delta r_k\le \overline{\Delta r}\,\}$\hfill\tcp{Low-confidence prototypes}

\textbf{/* Stage-III: Similarity-driven Prototype Merge */}\\
Normalize prototypes $\mathbf{C}_{high}$ and $\mathbf{C}_{low}$\;
$\mathbf{S} \leftarrow \mathbf{C}_{low}\mathbf{C}_{high}^{\top}$\hfill\tcp{Cosine similarities}
$\pi_i \leftarrow \arg\max_j S_{ij}\quad\forall \mathbf{c}_i \in \mathbf{C}_{low}$ \hfill\tcp{Nearest anchor index}

% \textbf{/* 5. Merge prototypes */}\\
\For{each anchor $j=1,\dots,k^{\star}$}{
    $\mathcal{A}_j \leftarrow \{\,i \mid \pi_i = j\,\}$\;
    $\mathbf{C}_{high}^{(j)} \leftarrow \frac{1}{|\mathcal{A}_j|+1}\Bigl(\mathbf{C}_{high}^{(j)} + \sum_{i\in\mathcal{A}_j}\mathbf{C}_{low}^{(i)}\Bigr)$ \hfill\tcp{Merge prototypes}
}

\Return updated prototypes $\mathbf{C}'$
\end{algorithm}

\par
\noindent \textbf{PALGCD}~\cite{palgcd} tackles Generalized Category Discovery (GCD) by introducing a Prior-constrained Association Learning framework that leverages both parametric and non-parametric mechanisms to discover novel categories while preserving known-class representations.
Unlike previous methods that rely solely on self-distillation and parametric classification, PALGCD fully exploits cross-instance similarities by integrating labeled category priors directly into the association process, guiding unlabeled data toward semantically consistent groupings.
The resulting semantic associations are utilized through non-parametric prototypical contrastive learning to enhance representation quality.
In parallel, a hybrid classifier jointly trained on both parametric logits and prototype assignments ensures complementary decision boundaries across known and novel classes.
The implementation is publicly available at \url{https://github.com/Terminator8758/PAL-GCD}.
Building on the parametric classifier of SimGCD, we integrate the Prior-constrained Association Learning proposed in PALGCD.

\par
\noindent \textbf{CPL}~\cite{cpl} addresses open-world deepfake attribution on the proposed OW-DFA benchmark by introducing a Global-Local Voting module that aligns forged-face features across differently manipulated regions.
A confidence-based soft pseudo-labeling scheme mitigates noise caused by similar forgery methods within the unlabeled set.
The framework is further extended into a multi-stage pipeline that leverages pre-training and iterative learning to progressively enhance traceability of unknown attacks.
The implementation is based on the official code available at \url{https://github.com/TencentYoutuResearch/OpenWorld-DeepFakeAttribution}.

\par
\noindent \textbf{CDAL}\cite{cdal} builds upon CPL\cite{cpl} by introducing a Counterfactually Decoupled Attention Learning module to enhance open-world deepfake attribution.
Instead of relying on heuristic region partitioning, CDAL explicitly models the causal relationship between attention maps and source model attribution, and leverages counterfactual analysis to decouple discriminative traces from confounding biases.
This causal effect is used as a supervisory signal to guide attention learning, thereby encouraging the model to capture essential generation patterns that generalize better to unseen forgeries.
The method incurs minimal computational overhead and significantly improves performance under open-world settings.
The implementation is available at \url{https://github.com/yzheng97/CDAL}.

\section{Additional Experimental Results and Analysis}
\label{sec:supp_D}
As stated in the ``Results on OW-DFA'' (Sec.~5), and ``Hyper-parameter Selection'' and ``Evaluation of Class Number Estimation'' (Sec.~5) in the main text, we present the additional experimental results as follows. We also provide an evaluation under a harder setting (with less labeled data) in \cref{sec:supp_D4}, an analysis of training and inference efficiency in \cref{sec:supp_D5}, and an analysis of Real/Fake Detection in \cref{sec:supp_D6}.

\begin{table*}[th]
\small
\setlength{\tabcolsep}{1.0pt}
\renewcommand{\arraystretch}{1}
\caption{Results on OW-DFA benchmark~\cite{cpl}. The best results are marked in \textbf{bold}.}
% \vspace{-.15in}
\label{tab:comparison_cpldataset}
\begin{center}
\begin{tabular}{lccccccccccccccc}
\toprule
\multirow{3}{*}{\textbf{Method}} & 
\multirow{3}{*}{\textbf{Venue}} & 
\multicolumn{7}{c}{\textbf{Protocol-1}} & 
\multicolumn{7}{c}{\textbf{Protocol-2}} \\
\cmidrule(lr){3-9} \cmidrule(lr){10-16}
& & \multicolumn{3}{c}{\textbf{All}} & \multicolumn{3}{c}{\textbf{Novel}} & \textbf{Known} 
  & \multicolumn{3}{c}{\textbf{All}} & \multicolumn{3}{c}{\textbf{Novel}} & \textbf{Known} \\
\cmidrule(lr){3-5} \cmidrule(lr){6-8} \cmidrule(lr){9-9} \cmidrule(lr){10-12} \cmidrule(lr){13-15} \cmidrule(lr){16-16}
& & \textbf{ACC} & \textbf{NMI} & \textbf{ARI} & \textbf{ACC} & \textbf{NMI} & \textbf{ARI} & \textbf{ACC} 
  & \textbf{ACC} & \textbf{NMI} & \textbf{ARI} & \textbf{ACC} & \textbf{NMI} & \textbf{ARI} & \textbf{ACC} \\
\midrule
Upper Bound     & -         & 96.7 & 93.9 & 93.6 & 95.4 & 91.6 & 92.1 & 98.2   & 96.8 & 93.8 & 95.0 & 94.2 & 91.9 & 93.1 & 98.6 \\
\midrule
DNA-Det~\cite{dnadet}       & AAAI22   & 35.0 & 55.6 & 24.9 & 34.8 & 44.2 & 19.4 & 74.5   & 54.4 & 50.1 & 31.5 & 28.4 & 26.0 & 8.2 & 89.1 \\
OW-GAN~\cite{openworldgan} & ICCV21   & 57.6 & 57.6 & 47.5 & 38.9 & 45.9 & 41.5 & \textbf{99.6}   & 69.3 & 58.6 & 61.1 & 46.7 & 53.7 & 45.8 & \textbf{99.6} \\
RankStats~\cite{rs}      & ICLR20   & 72.5 & 73.6 & 66.5 & 49.9 & 56.1 & 39.8 & 98.6   & 74.4 & 72.2 & 81.7 & 45.3 & 52.4 & 30.2 & 96.8 \\
ORCA~\cite{orca}         & ICLR22   & 80.8 & 79.2 & 74.1 & 66.3 & 63.0 & 53.3 & 97.2   & 79.0 & 78.0 & 83.8 & 53.8 & 60.0 & 38.9 & 95.0 \\
OpenLDN~\cite{openldn}      & ECCV22   & 63.9 & 71.4 & 62.5 & 45.8 & 51.0 & 38.1 & 97.4   & 71.2 & 73.3 & 82.5 & 42.2 & 50.7 & 28.9 & 96.4 \\
NACH~\cite{nach}         & NeurIPS22& 82.6 & 82.0 & 76.4 & 70.1 & 67.1 & 56.6 & 96.9   & 79.5 & 77.9 & 84.5 & 53.9 & 58.5 & 38.7 & 96.2 \\
CPL~\cite{cpl}             & ICCV23   & 83.7 & 82.3 & 77.6 & 71.9 & 68.2 & 59.4 & 97.5   & 81.1 & 80.2 & 85.0 & 59.9 & 63.9 & 43.8 & 95.6 \\
% \midrule
% Ours (w/o $K_U$)& -         & 76.3 & 75.3 & 66.5 & 60.2 & 59.1 & 49.4 & 97.1 & 80.1 & 81.6 & 84.9 & 60.9 & 69.7 & 53.2 & 94.1 \\
Ours & - & \textbf{88.4} & \textbf{85.9} & \textbf{82.3} & \textbf{81.2} & \textbf{77.2} & \textbf{71.6} & 96.5 & \textbf{83.1} & \textbf{83.8} & \textbf{89.0} & \textbf{64.8} & \textbf{72.8} & \textbf{53.1} & 97.5 \\
\bottomrule
\end{tabular}
\end{center}
\end{table*}

\begin{table*}[h]
\small
\setlength{\tabcolsep}{1pt}
\renewcommand{\arraystretch}{1}
\caption{Results on the OW-DFA-40 benchmark under the harder setting with 50\% (top) and 25\% (bottom) labeled data. Best results are in \textbf{bold}.}
% \vspace{-.15in}
\label{tab:comparison_hardersetting}
\begin{center}

\begin{tabular}{lccccccccccccccccccccc}
\toprule
\multicolumn{1}{c}{\multirow{3}{*}{\textbf{Method}}} &
\multicolumn{7}{c}{\textbf{Protocol-1}} & 
\multicolumn{7}{c}{\textbf{Protocol-2}} & 
\multicolumn{7}{c}{\textbf{Protocol-3}} \\

\cmidrule(lr){2-8} \cmidrule(lr){9-15} \cmidrule(lr){16-22}
& \multicolumn{3}{c}{\textbf{All}} & \multicolumn{3}{c}{\textbf{New}} & \textbf{Known} 
  & \multicolumn{3}{c}{\textbf{All}} & \multicolumn{3}{c}{\textbf{New}} & \textbf{Known} 
  & \multicolumn{3}{c}{\textbf{All}} & \multicolumn{3}{c}{\textbf{New}} & \textbf{Known} \\
  
\cmidrule(lr){2-4} \cmidrule(lr){5-7} \cmidrule(lr){8-8} \cmidrule(lr){9-11} \cmidrule(lr){12-14} \cmidrule(lr){15-15} \cmidrule(lr){16-18} \cmidrule(lr){19-21} \cmidrule(lr){22-22}

& \textbf{ACC} & \textbf{NMI} & \textbf{ARI} & \textbf{ACC} & \textbf{NMI} & \textbf{ARI} & \textbf{ACC} 
  & \textbf{ACC} & \textbf{NMI} & \textbf{ARI} & \textbf{ACC} & \textbf{NMI} & \textbf{ARI} & \textbf{ACC} 
  & \textbf{ACC} & \textbf{NMI} & \textbf{ARI} & \textbf{ACC} & \textbf{NMI} & \textbf{ARI} & \textbf{ACC}\\
  
\midrule
% ----------- Harder setting – 50% Labeled --------------
SimGCD     
& 72.3 & 80.6 & 53.8 & 64.3 & 74.4 & 56.6 & 81.6 
& 67.1 & 75.0 & 58.9 & 54.0 & 70.2 & 49.8 & 80.5 
& 75.1 & 82.2 & 57.9 & 69.6 & 74.5 & 62.7 & 80.7 \\

CPL         
& 75.6 & 82.2 & 71.2 & 56.9 & 69.5 & 49.2 & 89.2 
& 73.9 & 80.5 & 78.1 & 56.3 & 72.3 & 50.7 & 91.1 
& 78.7 & 85.5 & 57.3 & 72.2 & 77.7 & 65.1 & 81.5 \\

Ours        
& \textbf{86.5} & \textbf{89.2} & \textbf{88.9} & \textbf{73.1} & \textbf{81.5} & \textbf{68.9} & \textbf{97.3} 
& \textbf{78.7} & \textbf{85.3} & \textbf{86.6} & \textbf{60.5} & \textbf{78.2} & \textbf{58.7} & \textbf{96.1} 
& \textbf{87.0} & \textbf{90.0} & \textbf{89.1} & \textbf{77.1} & \textbf{82.0} & \textbf{74.1} & \textbf{94.4} \\
\midrule
% ----------- Harder setting – 25% Labeled --------------
SimGCD     
& 64.5 & 72.5 & 46.3 & 54.1 & 64.2 & 46.2 & 76.4 
& 60.4 & 70.7 & 47.0 & 49.0 & 66.1 & 42.6 & 74.2 
& 74.6 & 80.9 & 63.7 & 57.3 & 69.2 & 52.6 & 83.8 \\

CPL         
& 70.2 & 77.8 & 53.7 & 58.1 & 69.7 & 50.7 & 80.0 
& 66.1 & 74.0 & 65.7 & 49.8 & 66.5 & 42.9 & 81.8 
& 73.2 & 82.3 & 49.9 & 63.3 & 72.6 & 56.3 & 76.6 \\

Ours        
& \textbf{78.0} & \textbf{84.7} & \textbf{76.3} & \textbf{63.7} & \textbf{77.8} & \textbf{58.4} & \textbf{91.5} 
& \textbf{69.5} & \textbf{79.7} & \textbf{72.1} & \textbf{50.7} & \textbf{73.3} & \textbf{49.2} & \textbf{88.3} 
& \textbf{83.6} & \textbf{87.6} & \textbf{79.6} & \textbf{77.6} & \textbf{82.9} & \textbf{74.0} & \textbf{90.3} \\

\bottomrule
\end{tabular}%
\end{center}
% \vspace{-.25in}
\end{table*}

\subsection{Results on the OW-DFA Benchmark}
\label{sec:supp_D1}
To comprehensively evaluate our proposed method, we conduct comparative experiments on the OW-DFA~\cite{cpl} benchmark, as shown in \Cref{tab:comparison_cpldataset}.  
Baseline results are taken from CPL~\cite{cpl}, while our results are obtained by following the official dataset construction script available at \url{https://github.com/TencentYoutuResearch/OpenWorld-DeepFakeAttribution}. Our method achieves state-of-the-art performance across nearly all metrics under both evaluation protocols.  
Compared to CPL, it improves All ACC by 4.7\% and Novel ACC by 9.3\% under Protocol-1.  
Under Protocol-2, our method still achieves consistent gains, outperforming CPL by 2.0\% in All ACC and 4.9\% in Novel ACC.  
These results demonstrate stronger generalization to novel forgery types while maintaining high accuracy on known classes.  
Even without prior knowledge of the number of novel forgery categories, our method remains effective and robust across different evaluation settings.

\subsection{Evaluation on Hyperparameter Selection}
\label{sec:supp_D2}
\label{sec:hyper}
\begin{figure}[]
  \centering
  \includegraphics[width=\linewidth]{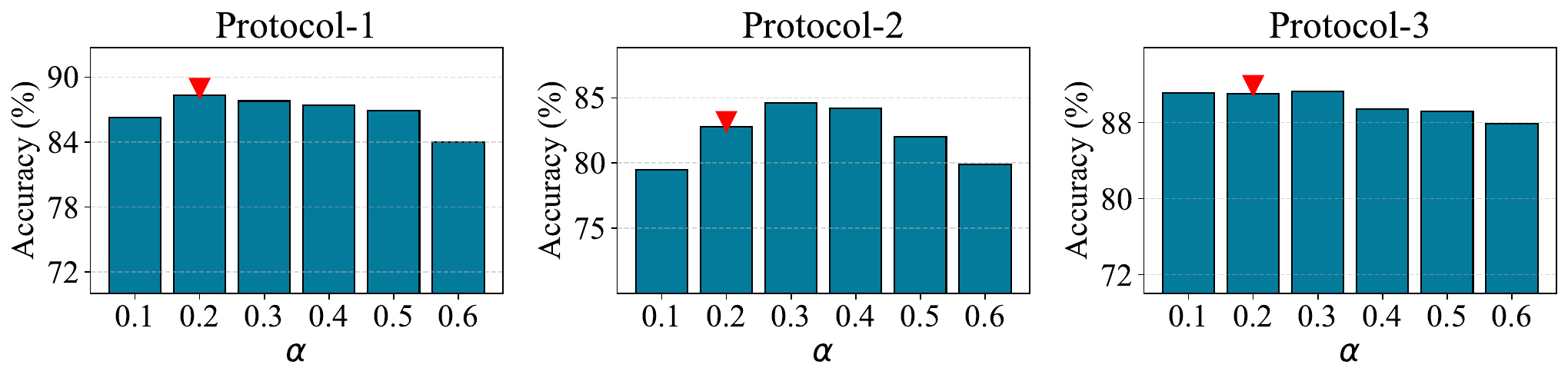}
  \includegraphics[width=\linewidth]{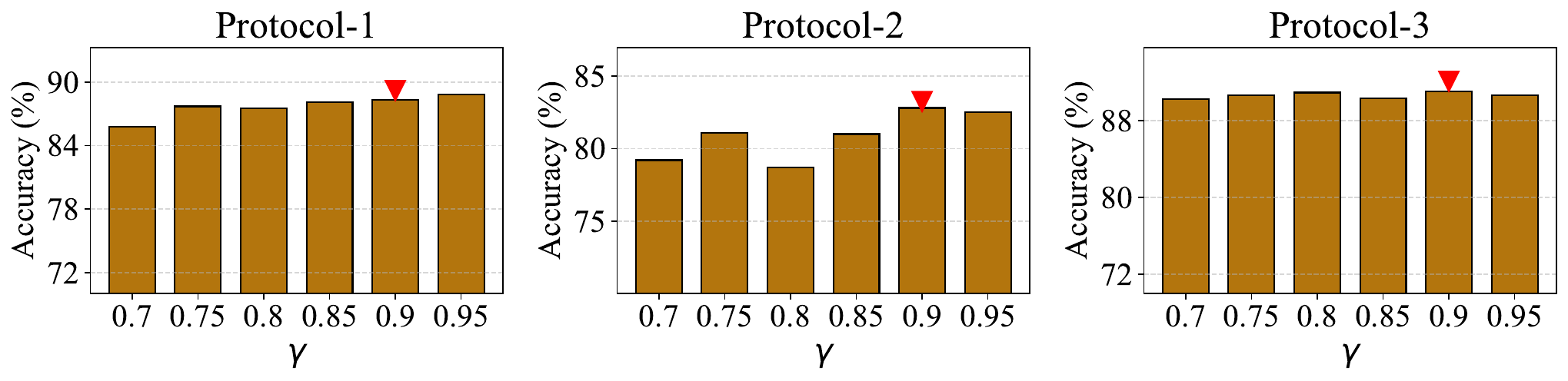}
  \caption{Effect of hyperparameters $\alpha$ and $\gamma$ under different evaluation protocols.\label{fig:hyper}}
\end{figure}

We adopt fixed hyperparameter values ($\alpha = 0.2$, $\gamma = 0.9$) throughout all experiments, as described in the main text. Although these values are not always the best across all test protocols, they are selected to avoid over tuning. In this appendix, we provide a detailed analysis of how $\alpha$ and $\gamma$ affect model performance under Protocols 1–3. As shown in \cref{fig:hyper}, both hyperparameters demonstrate stable performance across a range of values. Specifically, $\alpha$ is relatively robust in the range of 0.2 to 0.5, and $\gamma$ is robust between 0.85 and 0.95.

\subsection{Robustness of Class Number Estimation}
\label{sec:supp_D3}
\begin{figure}[]
\centering
\includegraphics[width=\linewidth]{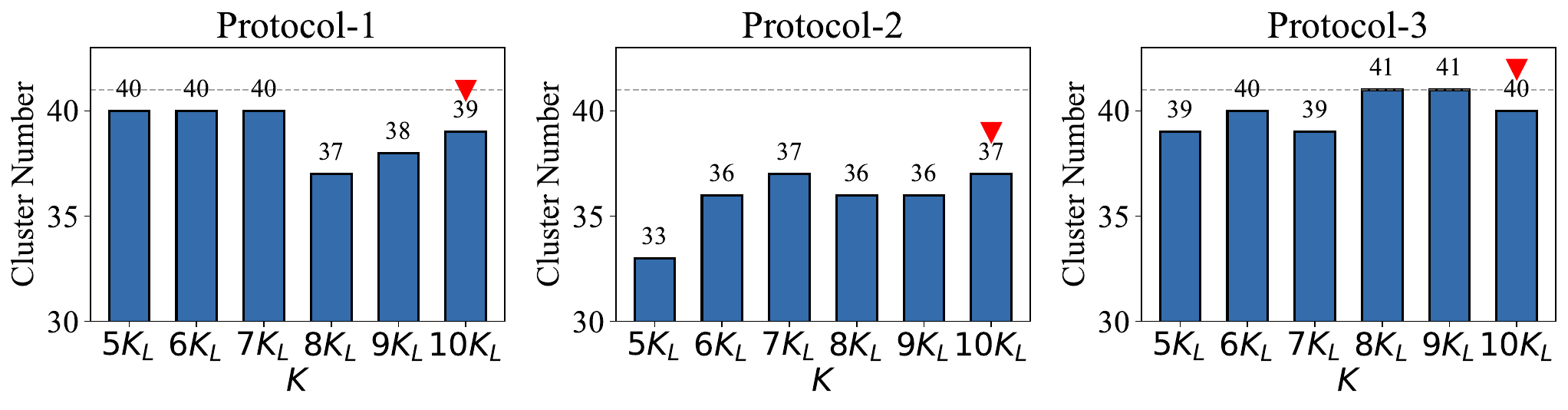}
\includegraphics[width=\linewidth]{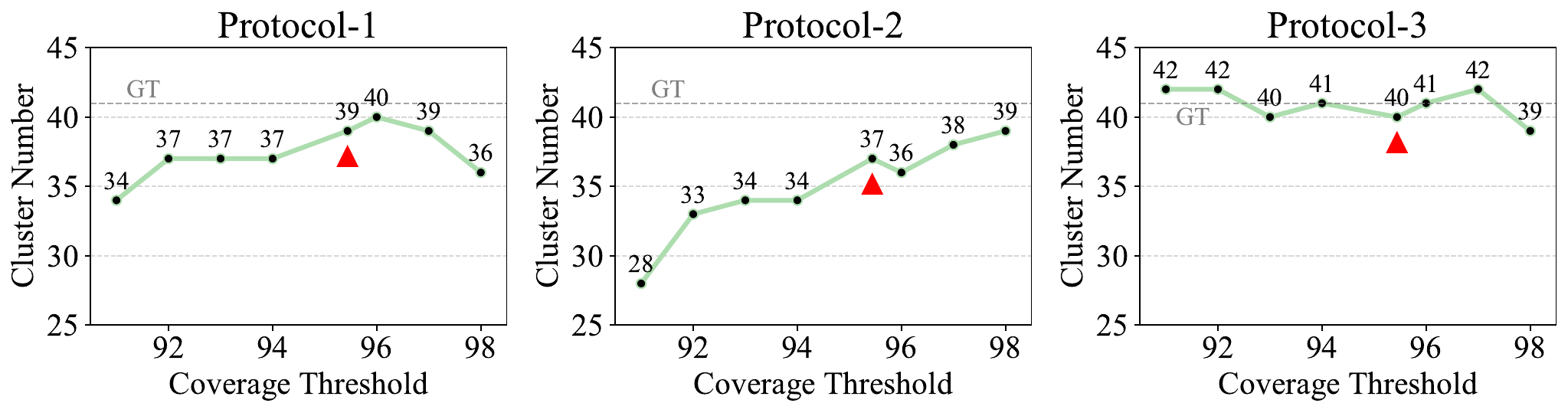}
\caption{Robustness of class number estimation with respect to different initializations of $K_L$ and coverage threshold $\mathcal{I}$. \label{fig:estk}}
\end{figure}

In the main text, we compared our class number estimation method with the GCD baseline and visualized the estimated class number over training epochs. Here, we further analyze the robustness of our estimation strategy under varying initialization settings. We focus on two key initialization parameters: the initial cluster number $K$ and the coverage threshold $\mathcal{I}$. In our default setup, $K$ is initialized as $10 \times K_L$ and $\mathcal{I}$ is set to 95.44\%. The results across different test protocols are summarized in \cref{fig:estk}. For $K$, we vary the initialization of $K$ from $5K_L$ to $10K_L$ and evaluate the final estimated class number. As shown in the top row of \cref{fig:estk}, our estimation remains consistently close to the ground truth (GT = 41) across all three protocols. In particular, Protocol-1 achieves stable estimates between 37 and 40; Protocol-2 shows minor fluctuations between 33 and 37; and Protocol-3 remains tightly around the GT, ranging from 39 to 41. This confirms that our method is robust to the choice of $K_L$. For the coverage threshold $\mathcal{I}$, we vary it from 91 to 98 and observe its influence on the estimated number of clusters. As shown in the bottom row of \cref{fig:estk}, higher values of $\mathcal{I}$ generally yield more accurate estimates. In Protocol-1, the estimation improves from 34 to 40 as $\mathcal{I}$ increases; Protocol-2 and Protocol-3 both exhibit stable behavior within the range of 94\%–98\%, with final values close to the GT. These results suggest that our estimation method is both effective and robust to initialization.

\subsection{Evaluation under Harder Settings}
\label{sec:supp_D4}
To assess the robustness of our approach under limited supervision, we follow~\cite{cpl} and reduce the proportion of labeled data from known classes to 50\% and 25\%. As shown in \cref{tab:comparison_hardersetting}, our method consistently outperforms CPL and SimGCD in both cases.
Under the 50\%-labeled setting, averaged over all three protocols, our method surpasses CPL by \textbf{8.0\%} in All ACC and \textbf{8.4\%} in Novel ACC, and outperforms SimGCD by \textbf{12.6\%} and \textbf{7.6\%}, respectively.
When the labeled proportion is further reduced to 25\%, our model still achieves strong results, exceeding CPL by \textbf{7.2\%} in All ACC and \textbf{6.9\%} in Novel ACC, and outperforming SimGCD by \textbf{10.5\%} in both All ACC and Novel ACC. \textit{These results clearly demonstrate that our method exhibits superior robustness under reduced supervision, effectively leveraging limited labeled data to achieve reliable attribution in open-world settings.}

\begin{table}[h]
\small
\setlength{\tabcolsep}{2pt}
\renewcommand{\arraystretch}{1}
\caption{Comparison of average training time (per epoch) and inference time (in seconds) across three protocols.}
\label{tab:efficiency}
\begin{center}
\begin{tabular}{lcccccc}
\toprule
\multirow{2}{*}{\textbf{Method}} &
\multicolumn{2}{c}{\textbf{Protocol-1}} &
\multicolumn{2}{c}{\textbf{Protocol-2}} &
\multicolumn{2}{c}{\textbf{Protocol-3}} \\
\cmidrule(lr){2-3} \cmidrule(lr){4-5} \cmidrule(lr){6-7}
& \textbf{Train} & \textbf{Infer} 
& \textbf{Train} & \textbf{Infer} 
& \textbf{Train} & \textbf{Infer} \\
\midrule
SimGCD         & 1022.2 & 206.3 & 944.2  & 192.7 & 1068.3 & 189.6 \\
CPL            & 993.4  & 185.4 & 951.8  & 197.1 & 1067.1 & 194.8 \\
Ours           & 971.0  & 183.6 & 894.5  & 174.4 & 1054.0 & 191.3 \\
Ours (w/o $K_U$)  & 987.7  & 185.9 & 911.2  & 175.1 & 1072.4 & 195.7 \\
\bottomrule
\end{tabular}
\end{center}
\end{table}

\subsection{Training and Inference Efficiency}
\label{sec:supp_D5}
To analyze the training and inference efficiency of our method compared to existing state-of-the-art approaches, we conduct experiments across three protocols, as summarized in \cref{tab:efficiency}. Overall, our method achieves consistently lower training time per epoch and inference time across all protocols. Specifically, the average training time of our method is 973.2 seconds, which is lower than that of SimGCD (1011.6s) and CPL (1004.1s). Similarly, our average inference time is 183.1 seconds, compared to 196.2s for SimGCD and 192.4s for CPL. In addition, we evaluate the overhead of our DPP strategy under the setting where the number of novel classes is unknown. Compared to our variant without $K_U$, the full method increases training time by only 1.9\% and inference time by 0.9\% on average. \textit{This shows that our DPP strategy introduces negligible computational cost while enabling effective open-world adaptation.}

\subsection{Real/Fake Detection}
\label{sec:supp_D6}
To further analyze the benefit of the deepfake attribution task in enhancing real/fake detection, we conduct additional experiments under our OW-DFA-40 benchmark. Specifically, we evaluate four approaches: (1) deepfake binary classification, which predicts whether an image is real or fake; (2) deepfake multi-class classification, which identifies the generation method among known forgeries and treats real faces as a separate class; (3) the CPL framework, which incorporates unlabeled data via open-world attribution; and (4) our proposed method. Approaches (1) and (2) are trained using only the labeled subset, while (3) and (4) additionally leverage unlabeled data during training. All methods are evaluated using classification accuracy across the three protocols of OW-DFA-40. As shown in Table~\ref{tab:binary_comparison}, deepfake attribution methods (CPL and Ours) consistently outperform the binary and multi-class classification baselines, highlighting the advantage of leveraging unlabeled data and modeling fine-grained attribution. Notably, our method achieves the highest accuracy across all three protocols. Compared to CPL, it improves detection accuracy by an average of 1.07\%, demonstrating that \textit{better attribution can lead to more reliable real/fake classification.}

\begin{table}[h]
\small
\setlength{\tabcolsep}{4pt}
\renewcommand{\arraystretch}{1.1}
\caption{Results for real/fake detection.}
\label{tab:binary_comparison}
\begin{center}
\begin{tabular}{lccc}
\toprule
\textbf{Method} & \textbf{Protocol-1} & \textbf{Protocol-2} & \textbf{Protocol-3} \\
\midrule
Binary & 96.12 & 96.69 & 96.03 \\
Multi  & 96.68 & 97.60 & 96.91 \\
CPL    & 97.66 & 98.06 & 97.40 \\
Ours   & \textbf{98.83} & \textbf{98.95} & \textbf{98.96} \\
\bottomrule
\end{tabular}
\end{center}
\end{table}

\section{Broader Impact and Limitations Discussion}
\label{sec:supp_E}

\subsection{Broader Impact}
\label{sec:supp_E1}
This paper focuses on the Open-World Deepfake Attribution (OW-DFA) task, which aims to trace the origin of forged content amid the rapid evolution of generative models. This has significant implications for enhancing the credibility of digital media. The positive impacts include: \textbf{\textit{(i)}} providing more robust forensic tools for news agencies, social media platforms, and law enforcement to combat deepfake-driven fraud, defamation, and opinion manipulation; \textbf{\textit{(ii)}} encouraging generative model developers to take greater responsibility for security through systematic analysis of model fingerprints (i.e., traceable features left by generative models), thereby fostering a healthier and more regulation-compliant ecosystem; and \textbf{\textit{(iii)}} establishing standardized benchmarks and shared datasets for the academic community to promote interdisciplinary research in model interpretability. 
However, this task also poses \textbf{\textit{potential risks}}. For example, due to bias in the training data, the attribution algorithm may exhibit higher false positive rates for videos involving certain languages or demographic groups, potentially leading to unintended algorithmic discrimination against marginalized communities.

\subsection{Limitations and Future Work}
\label{sec:supp_E2}
This study has several limitations that warrant further exploration:

\noindent\textbf{\textit{(1) First}}, although our method outperforms existing state-of-the-art approaches on both known and novel categories in the OW-DFA task, the attribution accuracy for novel categories remains significantly lower than that for known categories. Incorrect attribution by the algorithm could lead to inappropriate accountability or even legal disputes, necessitating human review and uncertainty quantification mechanisms in real-world deployment.

\noindent \textbf{\textit{(2) Second}}, while facial deepfakes represent some of the most harmful visual forgeries, manipulated images are no longer limited to faces. Our work does not explore deepfake attribution in other visual contexts such as object-level or scene-level manipulations, nor does it address multimodal forgeries involving synthetic audio, text, or cross-modal content. Extending the attribution framework to such cases remains an important direction for future research.

\noindent \textbf{\textit{(3) Third}}, the OW-DFA-40 benchmark relies solely on real face images from FaceForensics++ and Celeb-DF. While this design ensures consistency with prior studies, it introduces domain bias: both datasets primarily feature Western-style, studio-quality videos and lack coverage of diverse demographics, real-world camera artifacts, and low-resolution social media content. This may lead to overly optimistic performance estimates in practice. Incorporating more varied real-image sources would enhance the benchmark's realism and robustness.

\noindent \textbf{\textit{(4) Fourth}}, our current framework focuses on static attribution under a fixed training distribution and does not explicitly consider long-term knowledge accumulation or privacy-aware representation learning. Recent studies on privacy-preserving position embeddings~\cite{ren2023masked} and semantic sharing Transformers for image restoration~\cite{ren2024sharing} suggest that controlling spatial information exposure and selectively propagating key semantics can improve robustness and efficiency in vision models. Incorporating such ideas into OW-DFA may help reduce overfitting to spurious spatial cues and enhance generalization to unseen forgery types. Moreover, extending OW-DFA toward a continual learning setting, where new manipulation methods emerge over time, remains an important and challenging direction for future research. In addition, integrating semantic segmentation~\cite{Zhao_2023_CVPR,Zhao_2023_ICCV} into OW-DFA could enable region-aware attribution by disentangling method-specific artifacts across semantic regions, thereby reducing reliance on spurious spatial cues and improving generalization to unseen forgery types.

We leave these challenges to future work as promising directions to further strengthen the generalizability, robustness, and real-world applicability of OW-DFA.

\end{document}